\documentclass[10pt,twocolumn,letterpaper]{article}

\usepackage[pagenumbers]{cvpr} 

\newcommand\blfootnote[1]{%
  \begingroup
  \renewcommand\thefootnote{}\footnote{#1}%
  \addtocounter{footnote}{-1}%
  \endgroup
}

\definecolor{cvprblue}{rgb}{0.21,0.49,0.74}
\definecolor{pastelblue}{RGB}{174, 198, 207}

\usepackage[pagebackref,breaklinks,colorlinks,allcolors=cvprblue]{hyperref}

\DeclareMathOperator*{\argmax}{argmax}
\DeclareMathOperator*{\argmin}{argmin}

\def\paperID{12571} 
\def\confName{CVPR}
\def\confYear{2025}

\title{Boost Your Human Image Generation Model via Direct Preference Optimization}

\author{
    Sanghyeon Na$^{*}$ \quad Yonggyu Kim$^{*}$ \quad Hyunjoon Lee$^{\dag}$ \\
    Kakao \\
    \texttt{\{orca.ai, arthur.a, malfo.y\}@kakaocorp.com} 
}

\begin{document}

\twocolumn[{%
\renewcommand\twocolumn[1][]{#1}
\maketitle
\includegraphics[width=\linewidth]{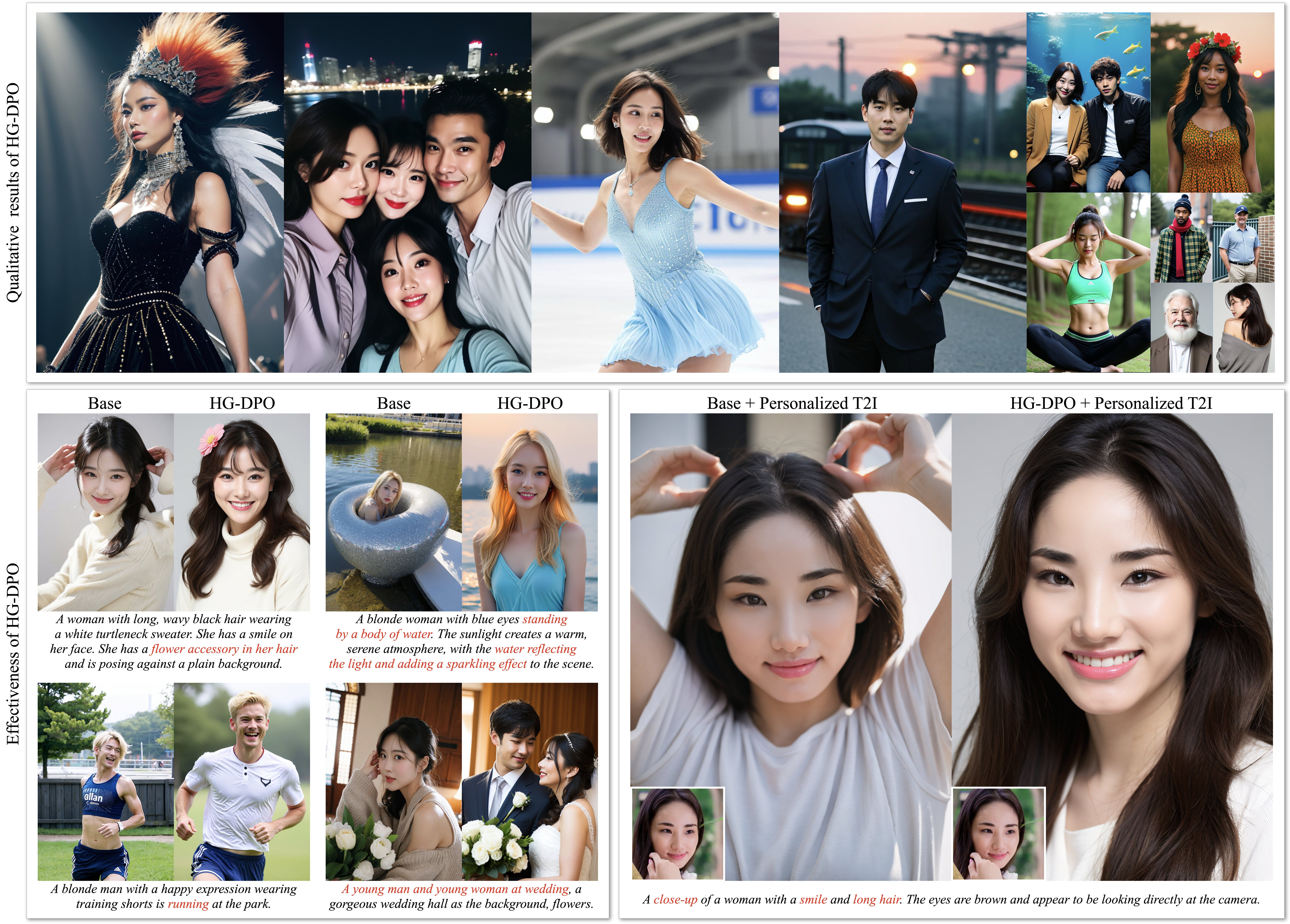}
\vspace{-2em}
\captionof{figure}{
    \textbf{Top:} HG-DPO generates high-quality human images that encompass a wide range of actions, appearances, group sizes, and backgrounds. \textbf{Bottom left:} This is because HG-DPO improves the base model to generate images with more realistic anatomical features and poses, while also better aligning with the prompt (red text in the prompt). \textbf{Bottom right:} The benefits of HG-DPO transfer to personalized text-to-image tasks without additional training, generating high-quality images with the identity of concept image.
    \vspace{2.1em}
}
\label{fig:teaser}
}]

\begin{abstract}
Human image generation is a key focus in image synthesis due to its broad applications, but even slight inaccuracies in anatomy, pose, or details can compromise realism. To address these challenges, we explore Direct Preference Optimization (DPO), which trains models to generate preferred (winning) images while diverging from non-preferred (losing) ones. However, conventional DPO methods use generated images as winning images, limiting realism. To overcome this limitation, we propose an enhanced DPO approach that incorporates high-quality real images as winning images, encouraging outputs to resemble real images rather than generated ones. However, implementing this concept is not a trivial task. Therefore, our approach, \textbf{HG-DPO} (\textbf{H}uman image \textbf{G}eneration through \textbf{DPO}), employs a novel curriculum learning framework that gradually improves the output of the model toward greater realism, making training more feasible. Furthermore, HG-DPO effectively adapts to personalized text-to-image tasks, generating high-quality and identity-specific images, which highlights the practical value of our approach. 
\end{abstract}    
\section{Introduction}
\blfootnote{$^*$ Equal contribution. }
\blfootnote{$^{\dag}$ Corresponding author.}
\label{sec:intro}
Human image generation is a key focus in generative modeling due to its broad applications in entertainment and social media. Despite advances in text-to-image generation~\cite{saharia2022photorealistic,gu2022vector,ramesh2022hierarchical,nichol2021glide,rombach2022high,podell2023sdxl} using diffusion models~\cite{ho2020denoising,song2020score,dhariwal2021diffusion}, it remains difficult to generate realistic human images because even slight inaccuracies in anatomy, pose, or fine details can create artifacts that reduce realism. Conventional approaches, which rely on supervised fine-tuning of diffusion models using high-quality images, often struggle to achieve the desired realism. Our objective is to build upon this fine-tuned model as a base, enhancing it to produce realistic human images as shown in Figure~\ref{fig:teaser}.

In response, we explore Direct Preference Optimization (DPO)~\cite{rafailov2024direct,wallace2023diffusion}, which trains models on pairs of preferred (winning) and non-preferred (losing) images, guiding outputs toward preferred characteristics while avoiding non-preferred ones. This approach suits complex tasks like human image generation by leveraging the contrasts between winning and losing pairs. However, Diffusion-DPO~\cite{wallace2023diffusion} and its variants~\cite{yang2023using,yuan2024self,ethayarajh2024kto,gu2024diffusion,liang2024step,gambashidze2024aligning,hong2024margin,croitoru2024curriculum} struggle to achieve high realism in human image generation because they use \textit{generated} images as winning images, limiting output quality to the suboptimal level of generated images. To overcome this, we propose an enhanced DPO method with a novel preference structure that uses \textit{real} images as winning images, while treating generated images as losing images.

Notably, this preference structure integrates the training mechanism of Generative Adversarial Networks (GANs)~\cite{goodfellow2014generative}, which has proven highly effective in guiding human image generation toward greater realism~\cite{na2022mfim,karras2019style,karras2020analyzing,karras2021alias}, into diffusion models. In GANs, a discriminator assesses how closely a generated image resembles real images and penalizes deviations, thereby guiding the outputs to be more similar to real images. Similarly, HG-DPO penalizes outputs that resemble generated (losing) images, while encouraging outputs more similar to real (winning) images. Both methods guide outputs to resemble real images rather than generated ones, enhancing realism. However, unlike GANs, HG-DPO implements it through DPO framework by defining a preference structure between real and generated images.

Based on this concept, our initial experiments used a naive approach, applying DPO with real images as winning and generated images as losing. However, this approach fell short, likely due to the significant domain gap between real and generated images. For example, real images typically have more realistic compositions, poses, and intricate details. This gap can make a single-stage training difficult.

To bridge this gap, we integrate curriculum learning~\cite{bengio2009curriculum} into the DPO framework, gradually training the model from easy to hard tasks. As shown in Figure~\ref{fig:model_figure}, HG-DPO training consists of three stages: easy, normal, and hard. To create tasks of varying difficulty at each stage, each stage utilizes a dataset constructed in a different manner. As a result, in the easy stage, the model learns basic human preferences, focusing on undistorted anatomy and poses, and better image-text alignment. The normal stage enhances visual quality by capturing more realistic compositions and poses. The hard stage refines fine details to match real images, enabling high-fidelity outputs. This gradual progression allows the model to produce images that closely resemble real ones. Notably, unlike existing DPO datasets~\cite{wu2023human,kirstain2024pick}, our datasets are constructed without costly human feedback.

Furthermore, HG-DPO can improve personalized text-to-image (PT2I)~\cite{ruiz2023dreambooth,shi2023instantbooth,xiao2023fastcomposer,ma2023subject,chen2023anydoor,chen2023disenbooth}, by generating higher-quality images tailored to specific identities, as shown in Figure~\ref{fig:teaser}, without requiring additional training. This versatility highlights its practical value for creative and social media applications.

In summary, our contributions are as follows:

(\lowercase\expandafter{\romannumeral1}) We propose a novel DPO approach, HG-DPO, to generate high-quality human images. Unlike existing DPO methods, our approach uses real images as winning images. This can be viewed as introducing the training mechanism of GANs into diffusion models.

(\lowercase\expandafter{\romannumeral2}) However, implementing our approach is challenging. To address this, we present a three-stage curriculum learning pipeline that enables the model to generate realistic images through gradual improvement.

(\lowercase\expandafter{\romannumeral3}) Unlike existing DPO datasets, our proposed methods for constructing DPO datasets does not require costly human feedback, making it more efficient.

(\lowercase\expandafter{\romannumeral4}) We demonstrate that HG-DPO effectively adapts to PT2I tasks without additional training, highlighting the practical value of our work.
\section{Related Work}
\label{sec:related_work}
\paragraph{Aligning diffusion models with human preferences.}
Direct Preference Optimization (DPO) offers an improvement over Reinforcement Learning from Human Feedback (RLHF)~\cite{ouyang2022training}, which is widely used to align large language models (LLMs) to human preferences~\cite{liu2023aligning, korbak2023pretraining, wu2024fine, song2024preference, liu2023statistical, zhao2023slic, cheng2023adversarial, liu2024lipo, dai2023safe, chen2024self}, by directly using human preference dataset without requiring a reward model. Building on the success of DPO in the field of LLMs, Diffusion-DPO~\cite{wallace2023diffusion} extends DPO to diffusion models.Since then, several methods~\cite{yang2023using,yuan2024self,ethayarajh2024kto,gu2024diffusion,liang2024step,gambashidze2024aligning,hong2024margin,croitoru2024curriculum} have demonstrated improved performance over Diffusion-DPO. However, these approaches primarily focus on enhancing training techniques while utilizing the Pick-a-Pic dataset~\cite{kirstain2024pick}, which contains only generated images. On the other hand, DPOK~\cite{fan2023dpok}, D3PO~\cite{yang2024using}, and AlignProp~\cite{prabhudesai2023aligning} propose online learning methods based on policy optimization, DPO, and reward model gradients, respectively. However, these approaches also do not incorporate real images into the training process. In contrast, we enable the use of real images in the DPO dataset by integrating curriculum learning into DPO. Among existing methods, Curriculum-DPO~\cite{croitoru2024curriculum} is the most closely related to ours, as it also incorporates curriculum learning into DPO. However, its performance gains are limited because it relies exclusively on generated images.

\paragraph{Curriculum Learning.}
Curriculum learning~\cite{bengio2009curriculum} trains models by first exposing them to simpler data or tasks, then progressively introducing more complex ones. It has proven effective in fields such as computer vision~\cite{tang2012self,zhang2021flexmatch,shi2016weakly,soviany2021curriculum,kumar2011learning,qin2020balanced,buyuktacs2021curriculum,soviany2020image}, natural language processing~\cite{kocmi2017curriculum,zhan2021meta,zhou2020uncertainty,zhao2020reinforced,liu2020norm,sachan2016easy,bao2020plato,liu2018curriculum}, and reinforcement learning~\cite{florensa2017reverse,murali2018cassl,fang2019curriculum,manela2022curriculum,luo2020accelerating,milano2021automated,narvekar2016source}. We employ curriculum learning to ease the transition of model's outputs from the generative to real domain, a shift challenging to achieve through single-stage training.

\section{HG-DPO}
\label{sec:hg_dpo}
\begin{figure}
    \centering
    \includegraphics[width=0.95\linewidth]{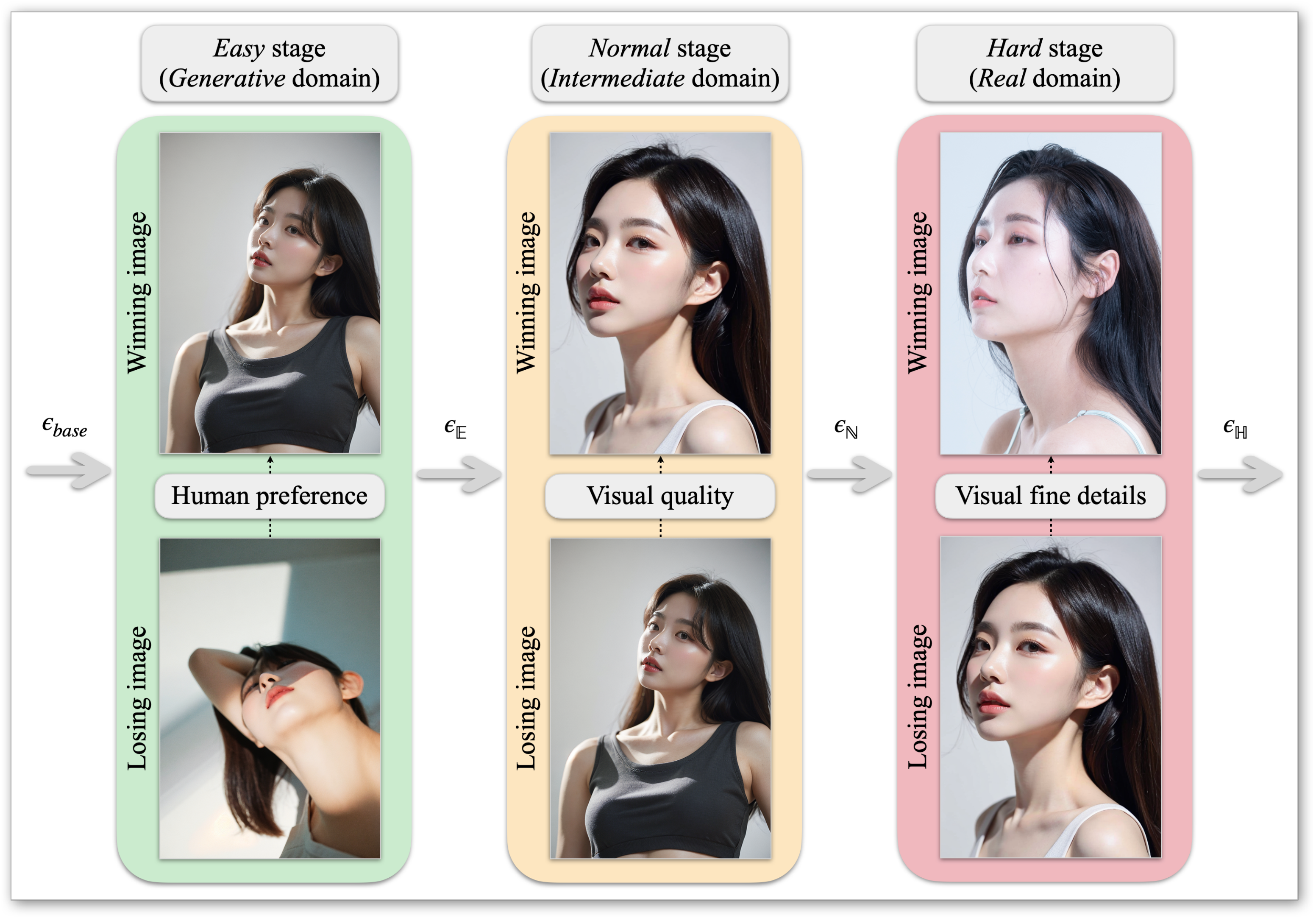}
    \vspace{-2.0mm}
    \caption{
    \textbf{Three-stage training of HG-DPO.} It progressively enhances the model's human image generation capabilities. 
    }
    \vspace{-3.0mm}
    \label{fig:model_figure}
\end{figure}

Given a human-specific paired image-text dataset $\mathcal{D}_\mathrm{real}=(\mathcal{P},\mathcal{X}_\mathrm{real})$, where $\mathcal{P}$ is a set of prompts and $\mathcal{X}_\mathrm{real}$ is a set of real human images, our objective is to design a model that can generate images with a level of realism similar to that of $\mathcal{X}_\mathrm{real}$. We begin by establishing a base model $\epsilon_{base}$ by supervised fine-tuning a backbone model, which is a latent diffusion model~\cite{rombach2022high} with $\mathcal{D}_\mathrm{real}$. Despite being fine-tuned on high-quality real images, $\epsilon_{base}$ often generates low-quality human images as shown in Figure~\ref{fig:qualitative_result_progress}.

Therefore, we refine $\epsilon_{base}$ to improve its human image generation using HG-DPO, which consists of three progressively challenging DPO stages: easy, normal, and hard (Figure~\ref{fig:model_figure}). In each stage, the model faces increasingly difficult objectives. The difficulty of each stage is determined by the domain of winning images, the target images that the model aims to generate. Specifically, the easy stage uses generated images as winning images. Since there is no domain gap - the model is trained on the same type of images it generates - this stage represents an easier task. In contrast, the hard stage uses real images as winning images. The model must learn from a real domain, which differs significantly from its generative domain, making it a more difficult task. For losing images, we use the winning images from the previous stage, except in the easy stage, which has no prior stage. Through each stage, the model is progressively refined to generate images that more closely resemble real images. Here, we employ LoRA~\cite{hu2021lora} for training.

\subsection{Easy Stage}
\label{subsec:easy_stage}
\begin{figure}
    \centering
    \includegraphics[width=\linewidth]{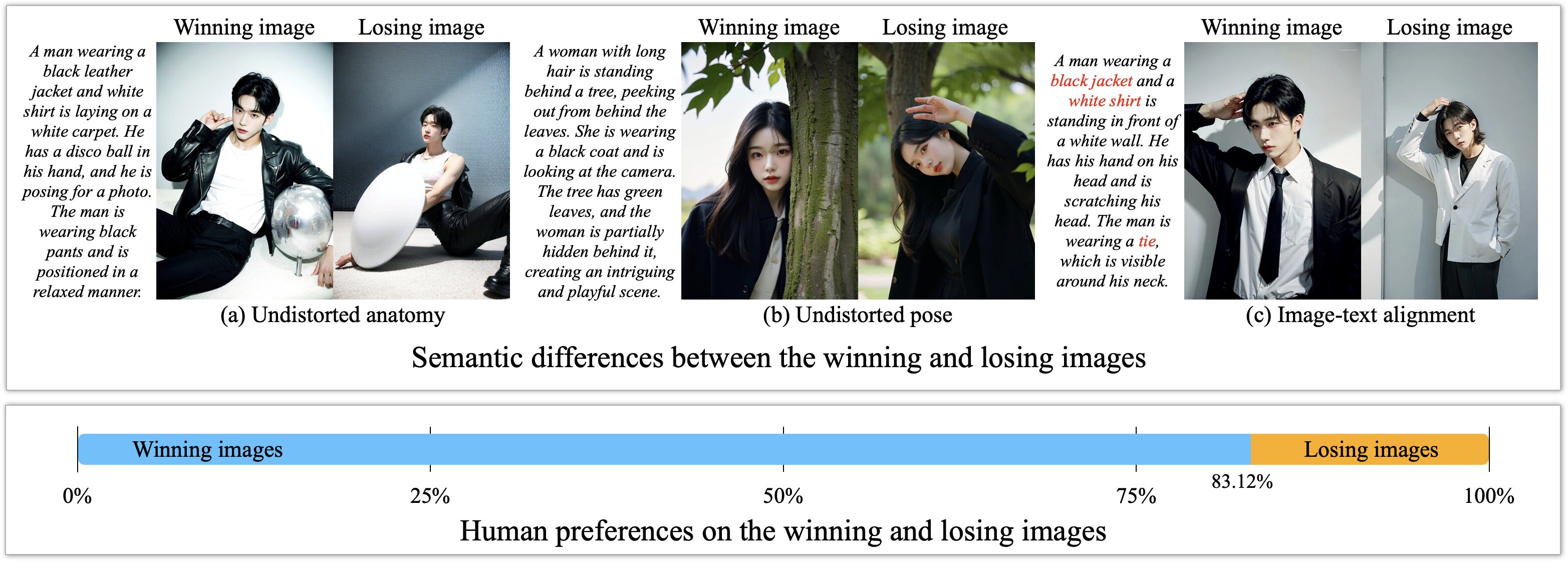}
    \vspace{-6.0mm}
    \caption{\textbf{DPO Dataset for the easy stage.} In the upper figure, $\mathcal{D}_\mathbb{E}$, constructed with AI rather than human feedback, shows winning images with superior features over losing images. A user study in the lower figure confirms this outcome.}
    \vspace{-4.0mm}
    \label{fig:qualitative_result_dataset_easy_stage}
\end{figure}

In the easy stage, we refine $\epsilon_{base}$ into $\epsilon_\mathbb{E}$ to generate images more likely preferred by humans. To achieve this, we create pairs of winning and losing images, where the winning images exhibit better anatomy, pose, and prompt alignment than the losing images as shown in the green box in Figure~\ref{fig:model_figure} and Figure~\ref{fig:qualitative_result_dataset_easy_stage}. 

\paragraph{Image pool generation.}
The first step is to generate images using a prompt set $\mathcal{P} = \{p^i\}_{i=1}^{D}$ where $p^i$ is a prompt and $D$ is the size of $\mathcal{D}_\mathrm{real}$. Instead of generating exactly two images for winning and losing, we create an \textit{image pool} with $N$ distinct images per prompt. For a prompt $p^i$, the image pool $\mathcal{X}_\mathrm{gen}^i$ of size $N$ is defined as:
\begin{align}
    \label{eq:easy_stage_image_pool}
    \mathcal{X}&_\mathrm{gen}^i= \{ x_\mathrm{gen}^{i_j} \}_{j=1}^N\text{ where }x_\mathrm{gen}^{i_j}=\mathcal{G}(\epsilon_{base},p^i,r^{i_j}). 
\end{align}
Here, $\mathcal{G}$ is a text-to-image sampler with a random seed $r^{i_j}$ used to generate $x_\mathrm{gen}^{i_j}$. To generate $N$ different images, we employ $N$ different random seeds $\{r^{i_j}\}_{j=1}^N$.

\paragraph{Selection of winning and losing images.}  
We then score the images using a human preference estimator $ f $~\cite{kirstain2024pick}:  
\begin{align}
    \label{eq:easy_stage_ai_feedback}
    \mathcal{S}_\mathrm{gen}^i=\{s_\mathrm{gen}^{i_j}\}_{j=1}^N\text{ where }s_\mathrm{gen}^{i_j}=f(x_\mathrm{gen}^{i_j},p^i),
\end{align}
where $s_\mathrm{gen}^{i_j}$ is a preference score of $x_\mathrm{gen}^{i_j}$ considering $p^i$. Unlike existing DPO datasets~\cite{kirstain2024pick,wu2023human} that rely on costly human feedback, we use a more efficient AI-based method.  
We select the images with the highest and lowest preference scores from the image pool $ \mathcal{X}_\mathrm{gen}^i $ as the winning and losing images, $x_\mathbb{E}^{\mathbf{w},i}$ and $x_\mathbb{E}^{\mathbf{l},i}$, respectively. It ensures clear semantic superiority of the winning image, consistent with human preferences (Figure~\ref{fig:qualitative_result_dataset_easy_stage}). Formally, this is defined as:  
\begin{align}
    \label{eq:easy_stage_image_choosing}
    (&x_\mathbb{E}^{\mathbf{w},i},~x_\mathbb{E}^{\mathbf{l},i})=(\mathcal{X}_\mathrm{gen}^i[j^w_\mathbb{E}],~\mathcal{X}_\mathrm{gen}^i[j^l_\mathbb{E}]) \\
    &\text{where }(j^w_\mathbb{E},~j^l_\mathbb{E})=(\argmax_{j^w_\mathbb{E}\in\{1,...,N\}} S_\mathrm{gen}^i,~\argmin_{j^l_\mathbb{E}\in\{1,...,N\}} S_\mathrm{gen}^i). \nonumber 
\end{align}
By assigning winning and losing images for all prompts in $ \mathcal{P} $, we complete the dataset:  
$\mathcal{D}_\mathbb{E}=\{(p^i,x_\mathbb{E}^{\mathbf{w},i},x_\mathbb{E}^{\mathbf{l},i})\}_{i=1}^{D}$.

\paragraph{Statistics matching loss.}  
With $ \mathcal{D}_\mathbb{E} $, we can apply the objective function of Diffusion-DPO ($ \mathcal{L}_\textit{D-DPO} $) to update $ \epsilon_{base} $ to $ \epsilon_\mathbb{E} $. However, $ \epsilon_\mathbb{E} $ trained this way produces color shift artifacts, making images unnatural (see $ N>2 $ in Figure~\ref{fig:qualitative_result_easy_stage_ablation}). This issue arises from the statistics of latents sampled by $ \epsilon_\mathbb{E} $ diverging from those sampled by $ \epsilon_{base} $. To address this, we design a statistics matching loss to prevent this divergence. Let $ l^t $ be a noisy latent of a winning image generated by the forward diffusion process at timestep $t$ during training, and let $\epsilon_\mathbb{E}^\theta$ be the model being updated in the easy stage. From $ l^t $, we sample $ l_{\theta}^{t-1} $ and $ l_{base}^{t-1} $, which are latents sampled using $\epsilon_\mathbb{E}^\theta$ and $ \epsilon_{base} $, respectively. The statistics matching loss is defined as:  
\begin{align}
    \label{eq:stat_matching_loss}
    \mathcal{L}_\textit{stat}=\mathbb{E}_{\mathcal{D}_\mathbb{E},~t\sim\mathcal{U}(0,~T)}\bigg[~||~\mu(l_{\theta}^{t-1})-\mu(l_{base}^{t-1})~||_2^2~\bigg]
\end{align}
where $\mu$ calculates the channel-wise mean. $\mathcal{L}_{stat}$ only matches the mean because it sufficiently resolves the color shift artifacts. During the easy stage, $ \epsilon_{base} $ is updated using the combined objective: $\mathcal{L} = \mathcal{L}_{\textit{D-DPO}} + \lambda_\textit{stat}\mathcal{L}_{\textit{stat}}$. $ \mathcal{L}_{\textit{stat}} $ is applied only in the easy stage, since color shift artifacts are not observed in the normal and hard stages.

\begin{table*}
    \begin{center}
        \scalebox{0.97}{
        \small
            \begin{tabular}                
            {lc@{~~~~}c@{~~~~}c@{~~~~}c@{~~~~}c@{~~~~}c@{~~~~}c@{~~~~}c@{~~~~}c}
                \toprule
                 Model & P-Score ($\uparrow$) & HPS ($\uparrow$) & I-Reward ($\uparrow$) & AES ($\uparrow$) & CLIP ($\uparrow$) & FID ($\downarrow$) & CI-Q ($\uparrow$) & CI-S ($\uparrow$) & ATHEC ($\uparrow$) \\
                \midrule
                HPD v2~\cite{wu2023human} & 21.7855 & 0.2829 & -0.0002 & 6.1132 & 29.99 & 35.98 & \underline{0.9015} & 0.9592 & 19.22 \\
                Pick-a-Pic v2~\cite{kirstain2024pick} & 21.7433 & 0.2831 & 0.0268 & 6.1315 & 30.05 & 37.89 & 0.8801 & 0.9464 & 19.00 \\
                Diffusion-DPO~\cite{wallace2023diffusion} & 17.9314 & 0.2425 & -1.8873 & 5.1183 & 24.03 & 112.67 & 0.8198 & 0.9438 & \textbf{36.30} \\
                NCP-DPO~\cite{gambashidze2024aligning} & 18.7679 & 0.2560 & -1.6644 & 5.5309 & 24.37 & 96.72 & 0.7164 & 0.8702 & 18.81 \\
                MAPO~\cite{hong2024margin} & 20.4401 & 0.2707 & -0.3613 & 5.4477 & 28.35 & 59.36 & 0.7117 & 0.8343 & \underline{32.68} \\
                Curriculum-DPO~\cite{croitoru2024curriculum} & 22.4381 & \underline{0.2869} & \underline{0.6532} & \underline{6.1925} & \underline{31.50} & \underline{35.35} & 0.8886 & 0.9561 & 23.36 \\
                AlignProp~\cite{prabhudesai2023aligning} & \textbf{23.0202} & 0.2854 & 0.1989 & \textbf{6.2773} & 29.67 & 49.92 & 0.8599 & \underline{0.9661} & 17.05 \\
                \rowcolor{pastelblue!30}
                HG-DPO (Ours) & \underline{22.6043} & \textbf{0.2872} & \textbf{0.7568} & 6.1785 & \textbf{31.57} & \textbf{29.41} & \textbf{0.9343} & \textbf{0.9858} & 29.41 \\
                \bottomrule
            \end{tabular}
        }
    \end{center}
    \vspace{-5.5mm}
    \caption{\textbf{Quantitative comparison with the previous methods.} HG-DPO achieves superior performance over the existing methods.}
    \label{table:quantitative_comparison}
\end{table*}

\begin{figure*}
    \centering
    \includegraphics[width=0.97\linewidth]{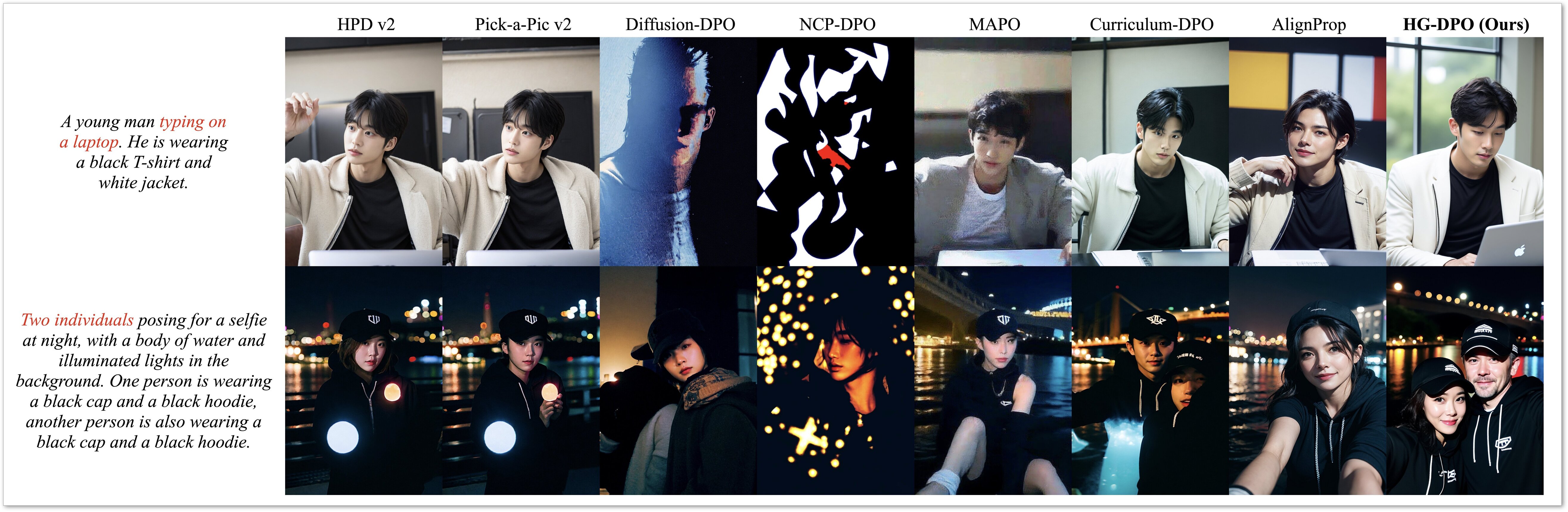}
    \vspace{-2.0mm}
    \caption{\textbf{Qualitative comparison with the previous methods.} HG-DPO generates high-quality human images with more realistic compositions and poses, providing superior text alignment compared to the prior methods.}
    \vspace{-2.0mm}
    \label{fig:qualitative_comparison}
\end{figure*}

\subsection{Normal Stage}
\label{subsec:normal_stage}
While $\epsilon_\mathbb{E}$ generates human images aligned with human preferences, they lack the realistic compositions and poses of real images. To address this, we refine $\epsilon_\mathbb{E}$ into $\epsilon_\mathbb{N}$ through the normal stage, improving visual quality (\eg, realistic composition and pose). Instead of immediately using real images as winning images after the easy stage, we introduce the intermediate domain for winning images as shown in the yellow box in Figure~\ref{fig:model_figure}. The normal stage facilitates training by providing an intermediate task before progressing from the easy to the hard stage.

\paragraph{Intermediate domain.}
The intermediate domain has mixed characteristics of generated and real domains. Specifically, we produce the intermediate images through \textit{Stochastic Differential Reconstruction (SDRecon)}, which is similar with SDEdit~\cite{meng2021sdedit}, where noise is added to a real image to generate a noisy image, which is then reconstructed back into a real image using $\epsilon_{base}$. Here, we can use $\epsilon_\mathbb{E}$, but by using $\epsilon_{base}$ instead, we eliminate the need to wait for $\epsilon_\mathbb{E}$ to be fully trained in order to construct the intermediate domain. This process retains certain features of the real image (\eg, composition and pose), while other features, like texture and fine details, resemble those of the generative domain. For example, in Figure~\ref{fig:model_figure}, the intermediate image (the winning image in the yellow box) maintains the pose of the real image (the winning image in the red box), but it lacks the fine details present in the real image (\eg, texture) resulting in a smoother and more synthetic appearance.

We perform SDRecon by varying diffusion timesteps to control noise magnitude, creating multi-level intermediate domains. It allows us to adaptively select the most appropriate intermediate domain during training. Formally, for each prompt $p^i \in \mathcal{P}$ and paired real image $x_\mathrm{real}^i \in \mathcal{X}_\mathrm{real}$, we use a set of $T$ timesteps, $\mathcal{T} = \{t_1, t_2, \ldots, t_T\}$, to generate a set of $T$ intermediate images:  
\begin{align}
    \label{eq:normal_stage_image_pool}
    \mathcal{X}_\mathrm{int}^i=\{ x_\mathrm{int}^{i_t} \}_{t\in\mathcal{T}}\text{ where }x_\mathrm{int}^{i_t}=\mathcal{R}(\epsilon_{base},p^i,x_\mathrm{real}^i,t)
\end{align}
where $\mathcal{R}$ is the proposed SDRecon operator. When $t = t_T$, strong noise produces an intermediate image closest to the generative domain, while $t = t_1$ results in an intermediate image closest to the real domain.

\paragraph{Selection of winning and losing images.} 
We score each image in $\mathcal{X}_{\mathrm{int}}^i$ to select the optimal winning image:
\begin{align}
    \label{eq:normal_stage_ai_feedback}
    \mathcal{S}_\mathrm{int}^i=\{ s_\mathrm{int}^{i_t} \}_{t\in\mathcal{T}}, 
    \text{ where }s_\mathrm{int}^{i_t}=f(x_\mathrm{int}^{i_t},p^i).
\end{align}
Then, we designate the image with the highest score in $\mathcal{X}_{\mathrm{int}}^i$ as the winning image $x_\mathbb{N}^{\mathbf{w},i}$, while the winning image from the easy stage is used as the losing image $x_\mathbb{N}^{\mathbf{l},i}$. Here, we consider only mid-level images from $\mathcal{X}_{\mathrm{int}}^i$ as candidates for winning images. Formally, this process is defined as
\begin{align}
    \label{eq:normal_stage_image_choosing}
    (x&_\mathbb{N}^{\mathbf{w},i},~x_\mathbb{N}^{\mathbf{l},i})=(\mathcal{X}_\mathrm{int}^i[j^w_\mathbb{N}],~\mathcal{X}_\mathrm{gen}^i[j^l_\mathbb{N}]) \\
    &\text{ where }(j^w_\mathbb{N},~j^l_\mathbb{N})=(\argmax_{j^w_\mathbb{N}\in\hat{\mathcal{T}}}S_\mathrm{int}^i,~\argmax_{j^l_\mathbb{N}\in\{1,...,N\}} S_\mathrm{gen}^i). \nonumber
\end{align}
Here, $\hat{\mathcal{T}}=\{t~|~t_r\leq t \leq t_g\}\text{ where }t_1 < t_r\text{ and }t_g < t_T$. Also, note that $x_\mathbb{N}^{\mathbf{l},i} = x_\mathbb{E}^{\mathbf{w},i}$.
By applying this process to all prompts in $ \mathcal{P} $, we obtain a set of triplets $\{ p^i, x_\mathbb{N}^{\mathbf{w}, i}, x_\mathbb{N}^{\mathbf{l}, i} \}_{i=1}^D$.

\paragraph{Filtering.} 
Instead of using all $D$ triplets, we filter them to retain only those containing images of sufficiently high quality, which are expected to benefit training. Specifically, a triplet is included in the normal stage dataset $\mathcal{D}_\mathbb{N}$ only when the score of $x_\mathbb{N}^{\mathbf{w},i}$ meets or exceeds that of $x_\mathbb{N}^{\mathbf{l},i}$:
\begin{align}
    \label{eq:normal_stage_filtering}
    \mathcal{D}_\mathbb{N}=\{(p^i&,x_\mathbb{N}^{\mathbf{w},i},x_\mathbb{N}^{\mathbf{l},i})\}_{i\in\mathcal{K}} \nonumber \\
    &\text{ where }\mathcal{K}=\{k|\mathcal{S}_\mathrm{int}^k[j^w_\mathbb{N}]\geq\mathcal{S}_\mathrm{gen}^k[j^l_\mathbb{N}] \}.
\end{align}
During the normal stage, we obtain $\epsilon_\mathbb{N}$ by updating $\epsilon_\mathbb{E}$ using $\mathcal{L}_\textit{D-DPO}$ with the dataset $\mathcal{D}_\mathbb{N}$.

\subsection{Hard Stage}
\label{subsec:hard_stage}
The objective of the hard stage is to enhance $ \epsilon_\mathbb{N} $ enabling it to generate images that closely resemble real images, including visual fine details. To achieve this goal, as shown in the red box of Figure~\ref{fig:model_figure}, the hard stage is designed to leverage \textit{real} images as winning images. However, we use images from the intermediate domain $t_1$, which are nearly indistinguishable from real images, as this approach has yielded better results in our experiments. Formally, the dataset for the hard stage, denoted as $\mathcal{D}_\mathbb{H}=\{(p^i,x_{\mathbb{H}}^{\mathbf{w},i},x_{\mathbb{H}}^{\mathbf{l},i})\}_{i\in\mathcal{K}}$ where $\mathcal{K}$ is the same as $\mathcal{K}$ in Eq.~\eqref{eq:normal_stage_filtering}, is constructed with winning and losing images defined as
\begin{align}    
    \label{eq:hard_stage_image_choosing}
    (x_{\mathbb{H}}^{\mathbf{w},i},x_{\mathbb{H}}^{\mathbf{l},i})=(\mathcal{X}_\mathrm{int}^i[t_1], x_{\mathbb{N}}^{\mathbf{w},i}).
\end{align}
Using the dataset $\mathcal{D}_\mathbb{H}$, we train $\epsilon_\mathbb{N}$ via $\mathcal{L}_\textit{D-DPO}$ to obtain $\epsilon_\mathbb{H}$.

\subsection{Training the Text Encoder}
\label{subsec:training_the_text_encoder}
To improve image-text alignment, we train the text encoder separately from the U-Net, following TexForce~\cite{chen2023enhancing}. We use both $\epsilon_\mathbb{H}$ and the enhanced text encoder at inference. The text encoder is trained only up to the easy stage, as the goal is to improve image-text alignment rather than visual quality.
\section{Experimental Settings}
\label{sec:experimental_settings}
In this section, we describe our experimental settings. Additional details not covered in this section are available in the Appendices.

\begin{table*}
    \begin{center}
        \scalebox{0.932}{
        \small
            \begin{tabular}
                {lc@{~~~~~}c@{~~~~~}c@{~~~~~}c@{~~~~~}c@{~~~~~}c@{~~~~~}c@{~~~~~}c@{~~~~~}c}
                \toprule
                Model & P-Score ($\uparrow$) & HPS ($\uparrow$) & I-Reward ($\uparrow$) & AES ($\uparrow$) & CLIP ($\uparrow$) & FID ($\downarrow$) & CI-Q ($\uparrow$) & CI-S ($\uparrow$) & ATHEC ($\uparrow$) \\
                \midrule
                Base ($\epsilon_{base}$) & 21.7364 & 0.2819 & -0.0665 & 6.1061 & 29.72 & 37.34 & 0.9058 & 0.9573 & 18.73 \\
                \hdashline[3pt/1.5pt]
                Naive & 17.9314 & 0.2425 & -1.8873 & 5.1183 & 24.03 & 112.67 & 0.8198 & 0.9438 & \textbf{36.30} \\
                Easy ($\epsilon_\mathbb{E}$) & 22.5384 & \textbf{0.2878} & \underline{0.7146} & 6.1775 & \underline{31.56} & 36.00 & 0.9057 & 0.9547 & 19.58 \\
                Normal ($\epsilon_\mathbb{N}$) & \underline{22.5422} & 0.2865 & 0.6515 & 6.1637 & 31.45 & \textbf{26.05} & 0.9302 & 0.9778 & 25.47 \\
                Hard ($\epsilon_\mathbb{H}$) & 22.4698 & 0.2867 & 0.5791 & \textbf{6.1955} & 31.15 & \underline{28.66} & \textbf{0.9365} & \textbf{0.9859} & \underline{30.08} \\
                \rowcolor{pastelblue!30}
                Hard ($\epsilon_\mathbb{H}$) + TE (HG-DPO) & \textbf{22.6043} & \underline{0.2872} & \textbf{0.7568} & \underline{6.1785} & \textbf{31.57} & 29.41 & \underline{0.9343} & \underline{0.9858} & 29.41 \\
                \hdashline[3pt/1.5pt]
                E2E training & 21.3244 & 0.2773 & -0.0487 & 6.0892 & 29.50 & 57.10 & 0.7962 & 0.7862 & 9.46 \\
                Hard w/o easy & 19.8417 & 0.2687 & -0.8262 & 5.7305 & 27.36 & 72.34 & 0.8801 & 0.9631 & 19.72 \\
                Hard w/o normal & 22.2541 & 0.2849 & 0.3932 & 6.1990 & 30.30 & 32.24 & 0.9233 & 0.9577 & 21.55 \\
                Base ($\epsilon_{base}$) + SFT & 21.732 & 0.2808 & 0.0140 & 6.0948 & 30.67 & 34.05 & 0.8860 & 0.9531 & 22.09 \\
                Easy ($\epsilon_\mathbb{E}$) + SFT & 21.858 & 0.2818 & 0.1788 & 6.0569 & 31.20 & 35.23 & 0.8802 & 0.9498 & 23.10 \\
                Normal ($\epsilon_\mathbb{N}$) + SFT & 21.9077 & 0.2818 & 0.1840 & 6.0691 & 31.08 & 35.12 & 0.8882 & 0.9534 & 23.29 \\
                \bottomrule
            \end{tabular}
        }
    \end{center}
    \vspace{-5.0mm}
    \caption{\textbf{Quantitative analysis of the HG-DPO pipeline.} The row labeled \textit{Base ($\epsilon_{base}$)} shows the base model's performance, while \textit{Naive} to \textit{Hard ($\epsilon_\mathbb{H}$) + TE} illustrate model progress through curriculum stages, ending with the final model, \textbf{HG-DPO}. Subsequent rows examine the importance of each curriculum stage. \textit{E2E training} refers to a model trained end-to-end using the combined training datasets from all three stages, instead of the proposed three-stage training. \textit{Hard w/o easy} and \textit{Hard w/o normal} exclude the easy and normal stages from the training pipeline, respectively. The last three rows indicate models with supervised fine-tuning (\textit{SFT}) using winning images of the hard stage after each curriculum step.}
    \vspace{-4.0mm}
    \label{table:quantitative_ablation_curriculum}
\end{table*}

\paragraph{Datasets.} 
We constructed an internal dataset consisting of approximately 300k high-quality human images. From this dataset, 5k images were randomly selected for testing, while the remaining images were utilized for training. Captions for the training and test images were generated using LLaVA~\cite{liu2023llava} and Qwen2-VL~\cite{Qwen2VL}, respectively. For evaluation, we generate images using these 5k test prompts.

\paragraph{Metrics.} 
To assess prompt-aware human preferences, we use PickScore (P-Score)~\cite{kirstain2024pick}, HPS-v2 (HPS)~\cite{wu2023human}, and ImageReward (I-Reward)~\cite{xu2024imagereward}. For prompt-independent preferences, we use the AestheticScore (AES) estimator~\cite{laionaes}. For image-text alignment, we employ CLIP~\cite{radford2021learning}. We apply FID~\cite{heusel2017gans} to measure the distance between the generated and real distributions using 5k test images. We also use CLIP-IQA~\cite{wang2022exploring} to evaluate image quality (CI-Q) and sharpness (CI-S), and ATHEC~\cite{peng2021athec} for additional sharpness assessment. CI-S uses pretrained CLIP, while ATHEC relies on the standard deviation of the Laplacian of image pixels. To quantify the color shift artifacts, we convert RGB images to HSV and calculate the circular mean of hue. We measure color shift severity of the target model by comparing the mean hue difference between the target model and the base model. Finally, for identity similarity in PT2I tasks, we compute feature distances using ArcFace~\cite{deng2019arcface} and VGGFace~\cite{cao2018vggface2}. Higher values indicate better performance for all metrics except FID, hue distance, ArcFace, and VGGFace.

\begin{figure}
    \centering
    \includegraphics[width=\linewidth]{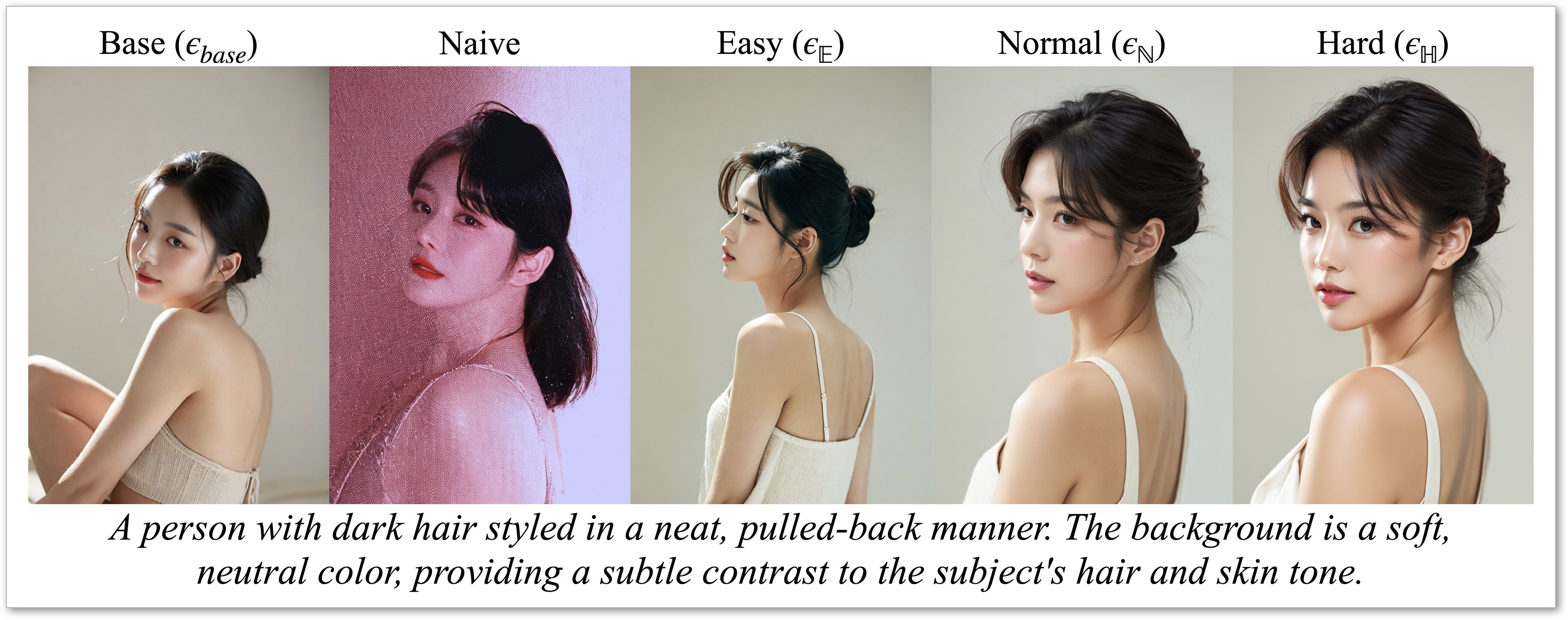}
    \vspace{-7.0mm}
    \caption{\textbf{Qualitative progress.} $\epsilon_{base}$ evolves as it progresses through each stage of the HG-DPO pipeline up to the hard stage.}
    \label{fig:qualitative_result_progress}
    \vspace{-4.0mm}
\end{figure}

\begin{figure}
    \centering
    \includegraphics[width=\linewidth]{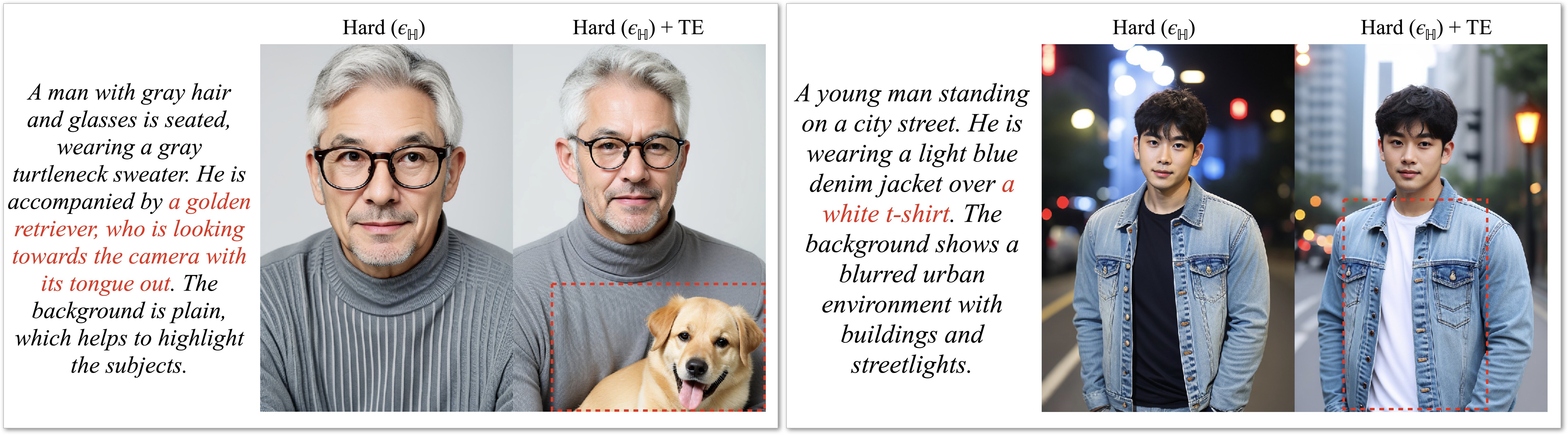}
    \vspace{-7.0mm}
    \caption{\textbf{Qualitative results by the enhanced text encoder.} With the improved text encoder, Hard ($\epsilon_\mathbb{H}$) + TE can achieve the enhanced image-text alignment, retaining the image quality of $\epsilon_\mathbb{H}$.}    
    \vspace{-3.0mm}
    \label{fig:qualitative_result_text_encoder}
\end{figure}

\begin{figure}
    \centering
    \includegraphics[width=\linewidth]{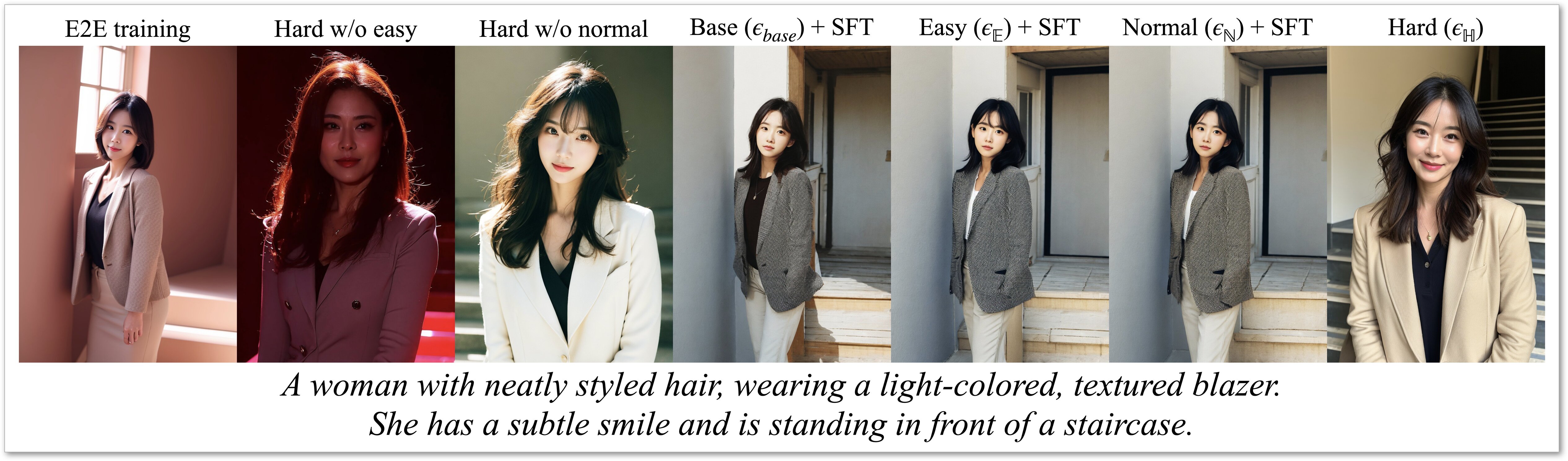}
    \vspace{-6.0mm}
    \caption{\textbf{Qualitative results illustrating the importance of each stage.} To generate high-quality images like the one labeled as Hard ($\epsilon_\mathbb{H}$), each stage of the HG-DPO pipeline is essential.}
    \vspace{-4.0mm}
    \label{fig:qualitative_result_necessity}
\end{figure}

\paragraph{Baselines.}
We compare HG-DPO with several existing methods. To show that publicly available datasets may not yield the best results, we include baselines trained on the HPD v2~\cite{wu2023human} and Pick-a-Pic v2~\cite{kirstain2024pick} datasets. To demonstrate that the naive approach described in Section~\ref{sec:intro} is suboptimal even with improved objectives, we introduce baselines trained with NCP-DPO~\cite{gambashidze2024aligning} and MAPO~\cite{hong2024margin}, which have outperformed Diffusion-DPO~\cite{wallace2023diffusion}. We compare HG-DPO with Curriculum-DPO~\cite{croitoru2024curriculum}, which applies curriculum learning within DPO, but uses only generated images. To clearly demonstrate the effectiveness of incorporating real images, Curriculum-DPO is trained on our effective image pool dataset (Eqs.~\eqref{eq:easy_stage_image_pool} and \eqref{eq:easy_stage_ai_feedback}). Finally, we introduce AlignProp~\cite{prabhudesai2023aligning} as an online learning baseline using PickScore~\cite{kirstain2024pick} as a reward model to highlight the advantage of our approach over the online learning approach.

\begin{table*}
    \begin{center}
        \scalebox{0.96}{
        \small
            \begin{tabular}
                {lc@{~~~~}c@{~~~~}c@{~~~~}c@{~~~~}c@{~~~~}c@{~~~~}c@{~~~~}c@{~~~~}c@{~~~~}c}
                \toprule
                Model & P-Score ($\uparrow$) & HPS ($\uparrow$) & I-Reward ($\uparrow$) & AES ($\uparrow$) & CLIP ($\uparrow$) & FID ($\downarrow$) & CI-Q ($\uparrow$) & CI-S ($\uparrow$) & ATHEC ($\uparrow$) & Hue ($\downarrow$)\\
                \midrule
                Base ($\epsilon_{base}$) & 21.7364 & 0.2819 & -0.0665 & 6.1061 & 29.72 & 37.34 & \textbf{0.9058} & \textbf{0.9573} & 18.73 & -\\
                \hdashline[3pt/1.5pt]
                $N = 2$ & 22.1939 & 0.2854 & 0.3610 & 6.1408 & 30.66 & \textbf{34.44} & 0.8887 & 0.9472 & 18.96 & \textbf{10.24}\\
                $N > 2$ & \textbf{22.5688} & \underline{0.2872} & \textbf{0.7830} & \textbf{6.2544} & \underline{31.50} & 37.29 & 0.8879 & 0.9471 & \textbf{27.20} & 98.54\\
                $N > 2$ + $\beta \uparrow$ & 22.2506 & 0.2864 & 0.5435 & 6.1129 & 31.30 & \underline{36.00} & 0.8416 & 0.9141 & 19.17 & \underline{23.77}\\
                \rowcolor{pastelblue!30}
                $N > 2$ + $\mathcal{L}_\textit{stat}$ ($\epsilon_\mathbb{E}$) & \underline{22.5384} & \textbf{0.2878} & \underline{0.7146} & \underline{6.1775} & \textbf{31.56} & \underline{36.00} & \underline{0.9057} & \underline{0.9547} & \underline{19.58} & 27.94\\
                \bottomrule
            \end{tabular}
        }
    \end{center}
    \vspace{-5.0mm}
    \caption{\textbf{Quantitative analysis of the easy stage.} For $\mathcal{D}_\mathbb{E}$, $N=2$ generates exactly two images per prompt, while $N>2$ builds an image pool as defined in Eq.~\eqref{eq:easy_stage_image_pool}. $N > 2 + \beta \uparrow$ and $N > 2 + \mathcal{L}_\textit{stat}$ add regularization to address the color shift artifacts in $N>2$. Specifically, $N>2 + \beta \uparrow$ applies a higher $\beta$, which is a strength of the original regularization in $\mathcal{L}_\textit{D-DPO}$, and $N > 2 + \mathcal{L}_\textit{stat}$ integrates $\mathcal{L}_\textit{stat}$ (Eq.~\eqref{eq:stat_matching_loss}). $N > 2 + \mathcal{L}_\textit{stat}$, which is highlighted in blue, is our proposed training configuration for the easy stage.}
    \vspace{-2.0mm}
    \label{table:quantitative_ablation_easy_stage}
\end{table*}

\section{Analysis on HG-DPO}
\label{sec:analysis_on_hg_dpo}
We quantitatively and qualitatively analyze HG-DPO to demonstrate its effectiveness. We retain the notations introduced in Section~\ref{sec:hg_dpo}. In Tables~\ref{table:quantitative_comparison},~\ref{table:quantitative_ablation_curriculum},~\ref{table:quantitative_ablation_easy_stage}, and \ref{table:quantitative_pt2i}, \textbf{bold} text and \underline{underlined} text indicate the best and second-best results, respectively. The row corresponding to our final model, HG-DPO, or the proposed training configuration is highlighted in blue in each table. Additional results not included in this section including a user study are provided in the Appendices.

\subsection{Comparison with the Previous Methods}
\label{subsec:comparison}
In this section, we compare HG-DPO with existing methods discussed in Section~\ref{sec:experimental_settings}. As shown in Figure~\ref{fig:qualitative_comparison}, HG-DPO produces more natural poses (first row) and better text alignment (second row) than models trained on the HPD v2~\cite{wu2023human} and Pick-a-Pic v2~\cite{kirstain2024pick} datasets. This highlights the limitations of relying solely on public datasets for human image generation, supported by the quantitative results in Table~\ref{table:quantitative_comparison}.

We also evaluate baselines trained with the naive single-stage approach, where real images are treated as winning images and generated ones as losing images, using various objective functions (Diffusion-DPO, NCP-DPO, and MAPO). However, Figure~\ref{fig:qualitative_comparison} shows that this naive approach fails to deliver satisfactory results, regardless of the objective function. In contrast, HG-DPO, leveraging curriculum learning, generates high-quality human images. Table~\ref{table:quantitative_comparison} further confirms the superiority of our curriculum-based approach over naive methods.

Additionally, we compare HG-DPO to Curriculum-DPO, which also employs curriculum learning but is exclusively trained on generated images. In Figure~\ref{fig:qualitative_comparison}, Curriculum-DPO fails to achieve realistic compositions, poses, and fine details compared to HG-DPO, demonstrating the benefits of including real images in training. This is consistent with the superior realism (FID and CI-Q) and sharpness (CI-S, and ATHEC) scores of HG-DPO compared to those of Curriculum-DPO in Table~\ref{table:quantitative_comparison}.

Finally, Figure~\ref{fig:qualitative_comparison} shows that AlignProp produces images with a distinct aesthetic style, deviating from the model's original style. This likely stems from optimizing only PickScore, which may cause catastrophic forgetting and mode collapse~\cite{li2024textcraftor,clark2023directly,wallace2023diffusion}. As a result, while AlignProp achieves higher P-Score and AES in Table~\ref{table:quantitative_comparison}, it scores lower on other metrics compared to HG-DPO.

\subsection{Progress through the HG-DPO Pipeline}
\label{subsec:ablation_study_progress}
We evaluate the impact of each stage in the HG-DPO pipeline. First, the naive single-stage approach, which treats real images as winning and images generated from their captions as losing, performs worse than $\epsilon_{base}$ (Figure~\ref{fig:qualitative_result_progress} and Table~\ref{table:quantitative_ablation_curriculum}). This underscores the challenge of using real images as winning images. Additionally, the model labeled as Naive in Table~\ref{table:quantitative_ablation_curriculum} corresponds to Diffusion-DPO in Table~\ref{table:quantitative_comparison}.

To address this, we first train $\epsilon_{base}$ to obtain $\epsilon_\mathbb{E}$ through the easy stage. In Table~\ref{table:quantitative_ablation_curriculum}, $\epsilon_\mathbb{E}$ shows significant improvements over $\epsilon_{base}$ in human preference metrics (P-Score, HPS, I-Reward, and AES) and image-text alignment (CLIP). Figure~\ref{fig:qualitative_result_progress} also shows that $\epsilon_\mathbb{E}$ produces better anatomical features than $\epsilon_{base}$. These gains stem from the superiority of the winning images in $\mathcal{D}_\mathbb{E}$ over the losing images (Figure~\ref{fig:qualitative_result_dataset_easy_stage}). However, FID and CI-Q, which measure image realism, show no significant improvement since the winning images are generated rather than real.

In the normal stage, $\epsilon_\mathbb{N}$ uses intermediate domains to produce more realistic images than $\epsilon_\mathbb{E}$. Figure~\ref{fig:qualitative_result_progress} shows that $\epsilon_\mathbb{N}$ generates more realistic composition, pose, and facial quality than $\epsilon_\mathbb{E}$, consistent with FID and CI-Q improvements in Table~\ref{table:quantitative_ablation_curriculum}. As realism improves, image sharpness also increases, reflected in CI-S and ATHEC scores.

\begin{figure}
    \centering
    \includegraphics[width=\linewidth]{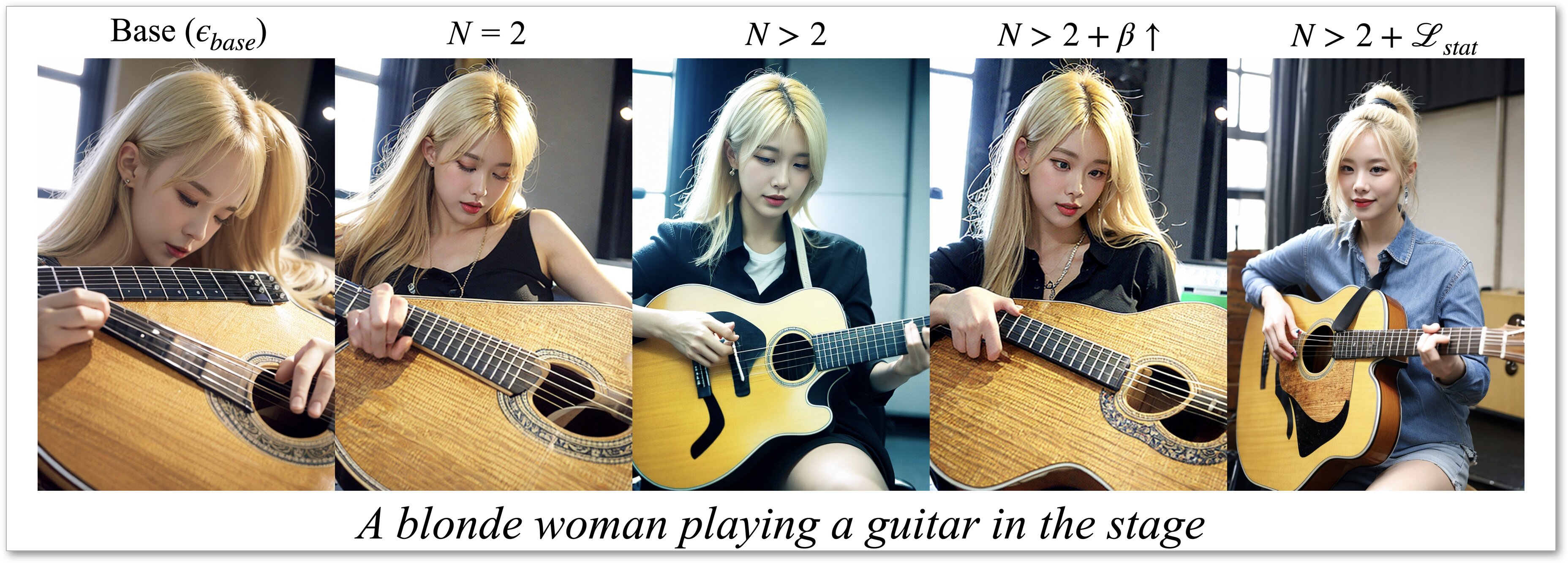}
    \vspace{-6.0mm}
    \caption{\textbf{Qualitative analysis of the easy stage.} In the easy stage, only the model trained with both $\mathcal{D}_\mathbb{E}$ and $\mathcal{L}_\textit{stat}$, namely $N>2 + \mathcal{L}_\textit{stat}$, produces images without distortions in pose and color.}
    \vspace{-4.0mm}
    \label{fig:qualitative_result_easy_stage_ablation}
\end{figure}

\begin{table*}
    \begin{center}
        \scalebox{0.89}{
        \small
            \begin{tabular}
                {lc@{~~~}c@{~~~}c@{~~~}c@{~~~}c@{~~~}c@{~~~}c@{~~~}c@{~~~}c@{~~~}c@{~~~}c}
                \toprule
                 Model & P-Score ($\uparrow$) & HPS ($\uparrow$) & I-Reward ($\uparrow$) & AES ($\uparrow$) & CLIP ($\uparrow$) & FID ($\downarrow$) & CI-Q ($\uparrow$) & CI-S ($\uparrow$) & ATHEC ($\uparrow$) & ArcFace ($\downarrow$) & VGGFace ($\downarrow$) \\
                \midrule
                IB~\cite{shi2023instantbooth} & 21.6847 & 0.2807 & -0.1045 & 6.0697 & 29.72 & 39.61 & 0.9034 & 0.9427 & 18.02 & 0.2662 & 72.11 \\
                \rowcolor{pastelblue!30}
                HG-DPO + IB & \textbf{22.5674} & \textbf{0.2855} & \textbf{0.6864} & \textbf{6.1321} & \textbf{31.24} & \textbf{29.30} & \textbf{0.9279} & \textbf{0.9806} & \textbf{26.79} & \textbf{0.2586} & \textbf{71.45} \\
                \bottomrule
            \end{tabular}
        }
    \end{center}
    \vspace{-5.0mm}
    \caption{\textbf{Quantitative results on PT2I.} HG-DPO brings its improvements to PT2I generation while preserving the identity injection capability of pre-trained PT2I model, InstantBooth (IB)~\cite{shi2023instantbooth}. The qualitative results are reported in the Appendices.}
    \vspace{-4.0mm}
    \label{table:quantitative_pt2i}
\end{table*}

Through the hard stage, $\epsilon_\mathbb{H}$ achieves even greater realism and sharpness than $\epsilon_\mathbb{N}$ (Figure~\ref{fig:qualitative_result_progress}), supported by higher CI-Q, CI-S, and ATHEC in Table~\ref{table:quantitative_ablation_curriculum}. However, $\epsilon_\mathbb{H}$ shows a slight FID degradation compared to $\epsilon_\mathbb{N}$, though it still surpasses $\epsilon_{base}$ and $\epsilon_\mathbb{E}$. Given that $\epsilon_\mathbb{H}$ achieves higher CI-Q, CI-S, and ATHEC scores than $\epsilon_\mathbb{N}$, this likely reflects reduced image diversity. However, our goal prioritizes image quality over diversity. Also, $\epsilon_\mathbb{H}$ shows lower image-text alignment than $\epsilon_\mathbb{E}$ and $\epsilon_\mathbb{N}$, though it remains better than $\epsilon_{base}$. This may result from the absence of the prompt-aware preference estimator (PickScore) in the hard stage, which was used in the easy and normal stages to select winning images (Eqs.~\eqref{eq:easy_stage_image_choosing} and \eqref{eq:normal_stage_image_choosing}). To address this, we first improve visual quality by training the U-Net through the hard stage, accepting some image-text alignment degradation, and then refine the text encoder to restore image-text alignment.

Our final model, \textbf{HG-DPO} (Hard ($\epsilon_\mathbb{H}$) + TE), integrates text encoder enhancements, improving image-text alignment over $\epsilon_\mathbb{H}$ (Table~\ref{table:quantitative_ablation_curriculum} and Figure~\ref{fig:qualitative_result_text_encoder}). This likely boosts prompt-aware human preference metrics (P-Score, HPS, and I-Reward) in Table~\ref{table:quantitative_ablation_curriculum}. HG-DPO also preserves the high realism and sharpness of $\epsilon_\mathbb{H}$, as shown by FID, CI-Q, CI-S, and ATHEC scores. Overall, HG-DPO significantly outperforms $\epsilon_{base}$ across all metrics, aligning with Figure~\ref{fig:teaser}.

\subsection{Necessity of the Each Proposed Stage}
\label{subsec:ablation_necessity}
We demonstrate that each stage in the proposed HG-DPO pipeline is essential for achieving optimal results.

Firstly, the model labeled as E2E training in Table~\ref{table:quantitative_ablation_curriculum} and Figure~\ref{fig:qualitative_result_necessity} is trained in an end-to-end manner by combining training datasets from all three stages, but this approach leads to suboptimal results, showing the effectiveness of our three-stage curriculum learning.

The model labeled as Hard w/o easy, which is trained only on the normal and hard stages, also produces a noticeably degraded results as shown in Table~\ref{table:quantitative_ablation_curriculum} and Figure~\ref{fig:qualitative_result_necessity}, underscoring the importance of the easy stage. In contrast, Hard w/o normal, trained only on the easy and hard stages, generates a more natural image than Hard w/o easy, as illustrated in Figure~\ref{fig:qualitative_result_necessity}. However, when it comes to fine detail, $\epsilon_\mathbb{H}$, which includes the normal stage, produces a more realistic image than Hard w/o normal. Table~\ref{table:quantitative_ablation_curriculum} supports this finding, where $\epsilon_\mathbb{H}$ achieves higher scores in FID, CI-Q, CI-S, and ATHEC, indicating the normal stage's crucial role to achieve optimal results. 

In addition, compared to the result for $\epsilon_\mathbb{H}$, the results for Base ($\epsilon_{base}$) + SFT, Easy ($\epsilon_\mathbb{E}$) + SFT, and Normal ($\epsilon_\mathbb{N}$) + SFT in Table~\ref{table:quantitative_ablation_curriculum} and Figure~\ref{fig:qualitative_result_necessity} reveal that supervised fine-tuning (SFT) with the hard-stage winning images is suboptimal when applied to models trained on earlier stages. Furthermore, these three models generate similar images, suggesting that SFT, regardless of the starting point ($\epsilon_{base}$, $\epsilon_\mathbb{E}$, or $\epsilon_\mathbb{N}$), may undesirably lead to forgetting knowledge previously learned through each stage.

\subsection{Additional Analysis on the Easy Stage}
\label{subsec:additional_analysis_on_easy_stage}
The significantly degraded results of Hard w/o Easy in Table~\ref{table:quantitative_ablation_curriculum} and Figure~\ref{fig:qualitative_result_necessity} highlight the importance of the easy stage. To explore this further, we conduct an additional analysis of the easy stage.

In Table~\ref{table:quantitative_ablation_easy_stage}, we observe that the model labeled as $N=2$ underperforms in human preference metrics and image-text alignment compared to configurations the model labeled as $N>2$, highlighting the effectiveness of employing the image pool. This observation is further supported by Figure~\ref{fig:qualitative_result_easy_stage_ablation}, where $N=2$ results in an image with a distorted pose, similar to that produced by $\epsilon_{base}$, whereas $N>2$ generates an image with an undistorted pose. 

However, Figure~\ref{fig:qualitative_result_easy_stage_ablation} shows that for $N > 2$, unnatural color shift artifacts occur, corroborated by the highest hue distance in Table~\ref{table:quantitative_ablation_easy_stage}. To mitigate these artifacts, increasing the weight of the regularization term from the original Diffusion-DPO objective significantly reduces the hue distance, as seen in the $N > 2 + \beta \uparrow$ results. However, this adjustment leads to a noticeable decline in human preference metrics and image-text alignment in Table~\ref{table:quantitative_ablation_easy_stage}. In contrast, our proposed statistics matching loss reduces the hue distance while maintaining strong performance in human preference metrics and image-text alignment, as evidenced by the results of $N > 2 + \mathcal{L}_\textit{stat}$ in Table~\ref{table:quantitative_ablation_easy_stage} and Figure~\ref{fig:qualitative_result_easy_stage_ablation}.

\subsection{Personalized T2I with HG-DPO}
\label{subsec:personalized_t2i}
To improve PT2I generation, we adapt pre-trained HG-DPO LoRA layers to the pre-trained InstantBooth (IB)~\cite{shi2023instantbooth}. As shown in Table~\ref{table:quantitative_pt2i}, HG-DPO + IB outperforms IB in human preferences, image-text alignment, image realism, and image sharpness while preserving identity injection, as indicated by similar ArcFace and VGGFace distances.

\section{Conclusion}
\label{sec:conclusion}
HG-DPO represents a significant advancement in human image generation by integrating real images and curriculum learning into the DPO framework. By gradually training the model from achieving basic anatomical accuracy to complex details of real images, HG-DPO narrows the realism gap between generated and real images. Furthermore, HG-DPO is adaptable to personalized T2I generation, consistently improving image quality. This adaptability makes it a valuable tool for creative applications and social media.
\newpage

{\LARGE \textbf{Appendices}}

\setcounter{section}{0}
\def\thesection{\Alph{section}}

\section{Additional Results of HG-DPO}
\label{sec:sm_additional_results_of_hg_dpo}

\begin{figure*}
    \centering
    \includegraphics[width=\linewidth]{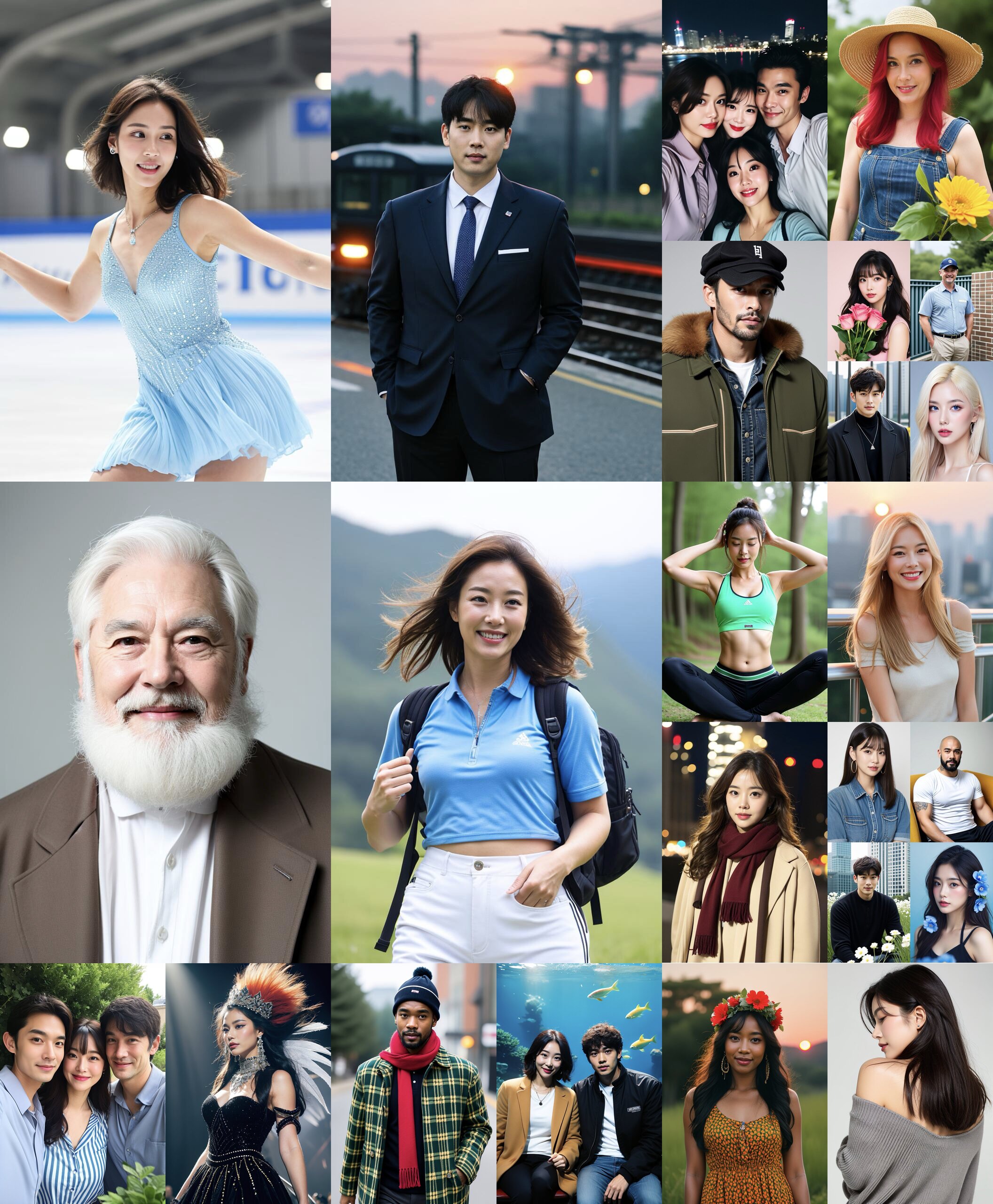}
    \caption{\textbf{Qualitative results of HG-DPO.} HG-DPO is capable of effectively generating high-quality human images that encompass a wide range of actions, appearances, group sizes, and backgrounds.}
    \label{fig:sm_teaser}
\end{figure*}

In this section, we present additional qualitative and quantitative results of HG-DPO to demonstrate the effectiveness of HG-DPO.

\subsection{Text-to-Image Generation}
\begin{figure}
    \centering
    \includegraphics[width=\linewidth]{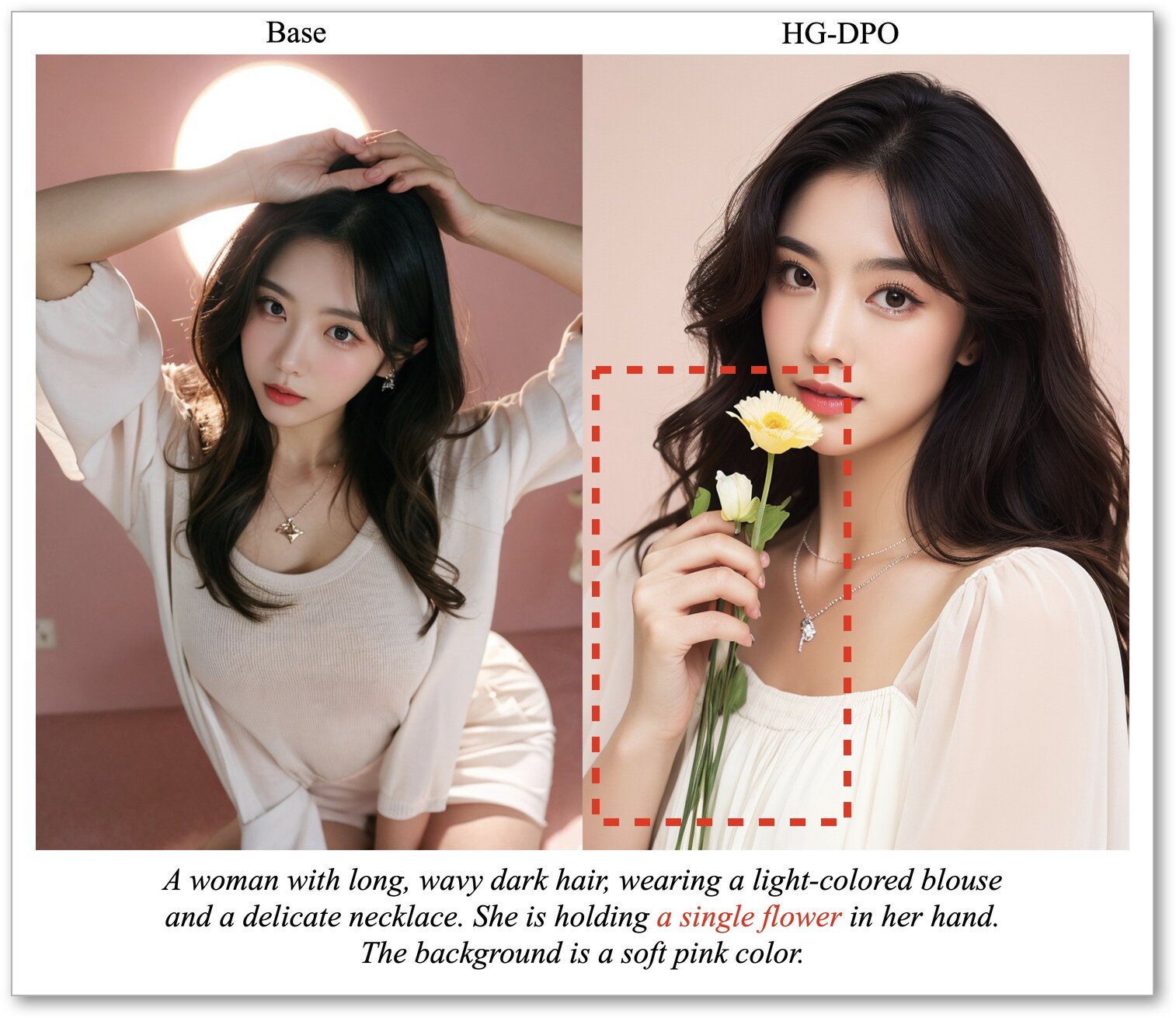}
    \vspace{-6.0mm}
    \caption{\textbf{Qualitative enhancements in text-to-image generation through HG-DPO.} HG-DPO improves the base model's capability to generate human images with more realistic poses and anatomies that align more accurately with the given prompt.}
    \label{fig:sm_base_vs_hg_dpo}
\end{figure}

As demonstrated in Figure~\ref{fig:sm_teaser}, HG-DPO successfully generates high-quality human images with diverse actions, appearances, group sizes, and backgrounds. This is made possible by HG-DPO's effective enhancement of the base model, as demonstrated by extensive experimental results in our manuscript and Figure~\ref{fig:sm_base_vs_hg_dpo}. 

As a result, in Table~\ref{table:sm_quantitative_comparison}, HG-DPO outperforms other existing methods. Table~\ref{table:sm_quantitative_comparison} is similar to Table~\ref{table:quantitative_comparison} in the manuscript but differs in two key aspects: it includes additional baselines, DPOK~\cite{fan2023dpok} and D3PO~\cite{yang2024using}, and uses 10 random seeds instead of a single one. To train DPOK and D3PO, we use our training prompt set $\mathcal{P}$ and PickScore~\cite{kirstain2024pick} as the reward model. While D3PO originally uses human feedback, we follow the authors’ setup by using the reward model instead. The results in Table~\ref{table:sm_win_rate}, which converts Table~\ref{table:sm_quantitative_comparison} to samplewise win rates, further highlight the effectiveness of HG-DPO.

Furthermore, HG-DPO significantly outperforms the base model and the previous approaches in the user study, as shown in Figure~\ref{fig:sm_user_study}. In the user study, we evaluated a selected subset of the baselines introduced in Section~\ref{sec:experimental_settings} of our manuscript against HG-DPO. Specifically, since the model trained with HPD~\cite{wu2023human} yields results similar to the model trained with Pick-a-Pic~\cite{kirstain2024pick} (see Figure~\ref{fig:qualitative_comparison} in our manuscript), we compared HG-DPO exclusively with the model trained using Pick-a-Pic~\cite{kirstain2024pick}, which is widely used in DPO-related studies. Furthermore, we excluded Diffusion-DPO~\cite{wallace2023diffusion}, NCP-DPO~\cite{gambashidze2024aligning}, and MAPO~\cite{hong2024margin} from the user study because these models often failed to generate images reliably and exhibited severe artifacts (see Figure~\ref{fig:qualitative_comparison} in our manuscript).

\begin{table*}
    \begin{center}
        \scalebox{1.0}{
        \small
            \begin{tabular}
                {lccccccccccc}
                \toprule
                 Model & P-Score ($\uparrow$) & HPS ($\uparrow$) & I-Reward ($\uparrow$) & AES ($\uparrow$) & CLIP ($\uparrow$) & FID ($\downarrow$) & CI-Q ($\uparrow$) & CI-S ($\uparrow$) & ATHEC ($\uparrow$) \\
                \midrule
                HPD v2 & 21.7211 & 0.2821 & -0.1353 & 6.0928 & 29.71 & 39.53 & 0.8856 & 0.9507 & 17.45 \\
                Pick-a-Pic v2 & 21.6778 & 0.2821 & -0.1352 & 6.0999 & 29.72 & 40.85 & 0.8614 & 0.9383 & 17.43 \\
                Diffusion-DPO & 18.0731 & 0.2408 & -1.9616 & 5.0637 & 23.49 & 160.11 & 0.6638 & 0.8715 & \textbf{40.31} \\
                NCP-DPO & 17.4631 & 0.2327 & -2.0222 & 4.7983 & 21.53 & 184.81 & 0.6342 & 0.8236 & 12.09 \\
                MAPO & 20.3971 & 0.2692 & -0.5150 & 5.4260 & 28.22 & 63.33 & 0.6459 & 0.7566 & \underline{30.71} \\
                Curriculum-DPO & 22.4298 & \underline{0.2868} & \underline{0.5823} & \underline{6.1874} & \underline{31.43} & \underline{37.02} & 0.8857 & 0.9528 & 21.63 \\
                AlignProp & \textbf{22.8933} & 0.2843 & 0.0693 & \textbf{6.2670} & 29.50 & 53.87 & 0.8534 & \underline{0.9609} & 15.67 \\
                DPOK & 21.6709 & 0.2809 & -0.2344 & 6.0998 & 29.25 & 41.52 & 0.8756 & 0.9332 & 15.68 \\
                D3PO & 21.6905 & 0.2810 & -0.1914 & 6.0764 & 29.59 & 41.26 & \underline{0.8902} & 0.9508 & 17.41 \\
                \rowcolor{pastelblue!30}
                HG-DPO (Ours) & \underline{22.5781} & \textbf{0.2871} & \textbf{0.7384} & 6.1758 & \textbf{31.53} & \textbf{30.91} & \textbf{0.9327} & \textbf{0.9852} & 28.28 \\
                \bottomrule
            \end{tabular}
        }
    \end{center}
    \vspace{-5.5mm}
    \caption{\textbf{Quantitative comparison with the previous methods.} HG-DPO achieves superior performance over the existing methods across nearly all evaluation metrics. \textbf{Bold} text and \underline{underlined} text indicate the best and second-best results, respectively. The row corresponding to our final model, HG-DPO, is highlighted in blue. For a more accurate comparison, we evaluate using 10 random seeds.}
    \label{table:sm_quantitative_comparison}
\end{table*}

\begin{table*}
    \begin{center}
        \scalebox{1.0}{
        \small
            \begin{tabular}
                {lcccccccccc}
                \toprule
                 Model & P-Score ($\uparrow$) & HPS ($\uparrow$) & I-Reward ($\uparrow$) & AES ($\uparrow$) & CLIP ($\uparrow$) & CI-Q ($\uparrow$) & CI-S ($\uparrow$) & ATHEC ($\uparrow$) \\
                \midrule
                vs HPD v2 & 85.13~\% & 76.44~\% & 82.25~\% & 62.17~\% & 73.15~\% & 82.31~\% & 86.64~\% & 93.75~\% \\
                vs Pick-a-Pic v2 & 86.03~\% & 76.14~\% & 82.08~\% & 61.06~\% & 72.64~\% & 89.63~\% & 90.87~\% & 93.79~\% \\
                vs Diffusion-DPO & 99.97~\% & 99.96~\% & 99.67~\% & 96.91~\% & 96.48~\% & 96.74~\% & 88.17~\% & 27.24~\% \\
                vs NCP-DPO & 99.97~\% & 99.85~\% & 99.78~\% & 95.04~\% & 98.95~\% & 99.61~\% & 96.94~\% & 97.02~\% \\
                vs MAPO & 97.92~\% & 98.35~\% & 88.51~\% & 96.26~\% & 84.78~\% & 98.89~\% & 98.10~\% & 42.07~\% \\
                vs Curriculum-DPO & 60.85~\% & 51.86~\% & 57.10~\% & 49.51~\% & 50.21~\% & 84.35~\% & 88.66~\% & 82.74~\% \\
                vs AlignProp & 33.82~\% & 62.12~\% & 75.80~\% & 37.73~\% & 74.02~\% & 95.35~\% & 85.45~\% & 97.98~\% \\
                vs DPOK & 86.19~\% & 80.71~\% & 84.06~\% & 61.80~\% & 77.67~\% & 85.67~\% & 91.82~\% & 95.91~\% \\
                vs D3PO & 85.90~\% & 81.70~\% & 83.69~\% & 64.16~\% & 74.82~\% & 81.09~\% & 87.46~\% & 92.90~\% \\
                \bottomrule
            \end{tabular}
        }
    \end{center}
    \vspace{-5.5mm}
    \caption{\textbf{Samplewise win rates~(\%) of HG-DPO against the previous methods.}  HG-DPO achieves superior performance over the existing methods across nearly all evaluation metrics. This table converts Table~\ref{table:sm_quantitative_comparison} into win rates, which means that these results are also calculated using 10 random seeds.}
    \label{table:sm_win_rate}
\end{table*}

\subsection{Personalized Text-to-Image Generation}
\begin{figure*}
    \centering
    \includegraphics[width=1.0\linewidth]{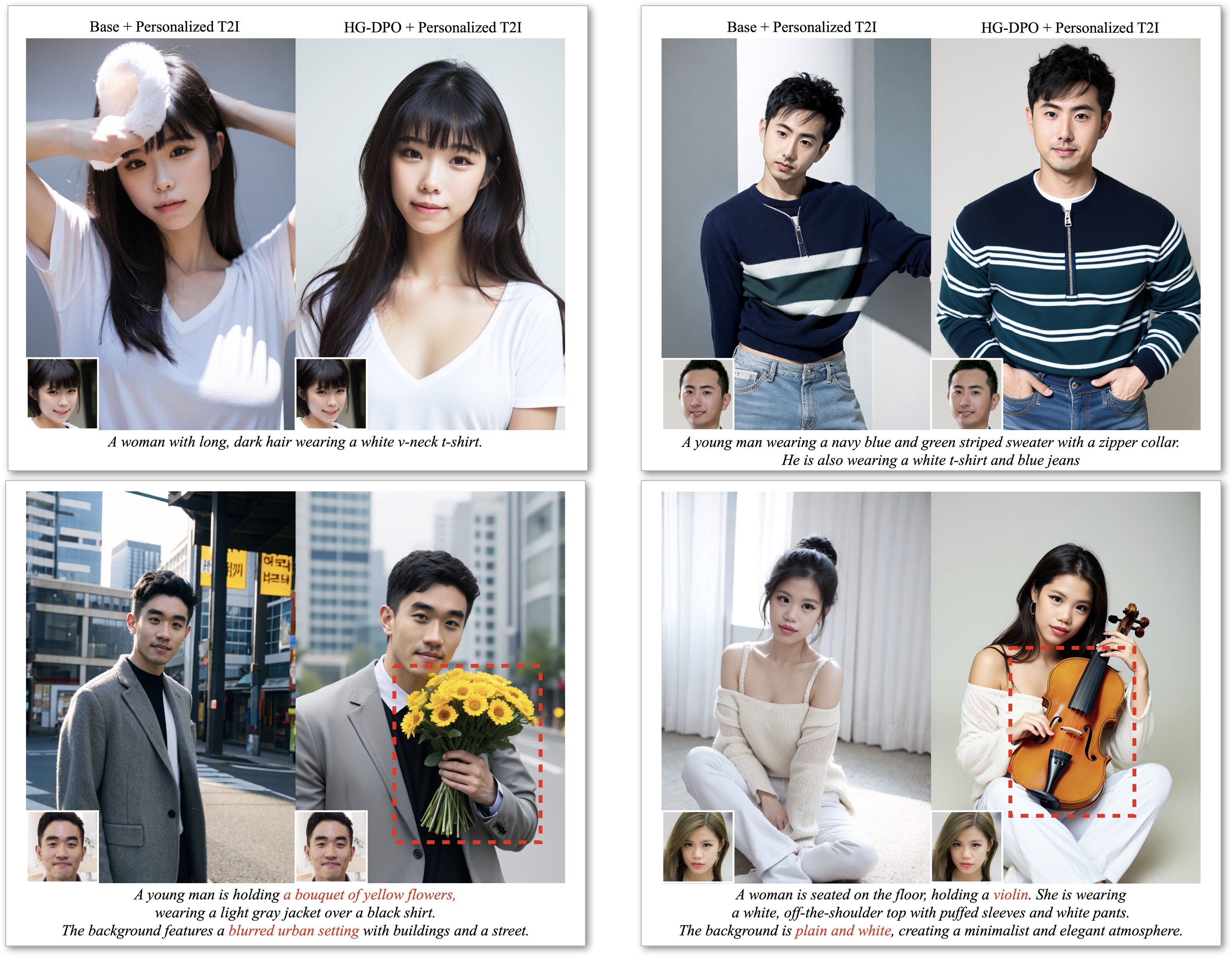}
    \vspace{-4.0mm}
    \caption{\textbf{Qualitative advancements achieved through in personalized text-to-image (PT2I) generation through HG-DPO.} HG-DPO improves the base model's capability to generate human images with more realistic poses and anatomies that align more accurately with the given prompt, and these improvements extend to PT2I generation. As a result, we can produce high-quality images that accurately reflect the identity of the concept image shown in the bottom left.}
    \label{fig:sm_pt2i}
\end{figure*}

HG-DPO significantly improves personalized text-to-image (PT2I) generation. As shown in Figure~\ref{fig:sm_pt2i}, this allows the generation of high-quality images that accurately reflect specific identities. Notably, these improvements are achieved without compromising the identity injection capability of the base PT2I model.

\section{Additional Analysis on the Easy Stage}
\label{sec:sm_additional_analysis_on_the_easy_stage}
In this section, we present additional experimental results and further analysis of the easy stage.

\subsection{Effectiveness of the Easy Stage}
\begin{figure*}
    \centering
    \includegraphics[width=1.0\linewidth]{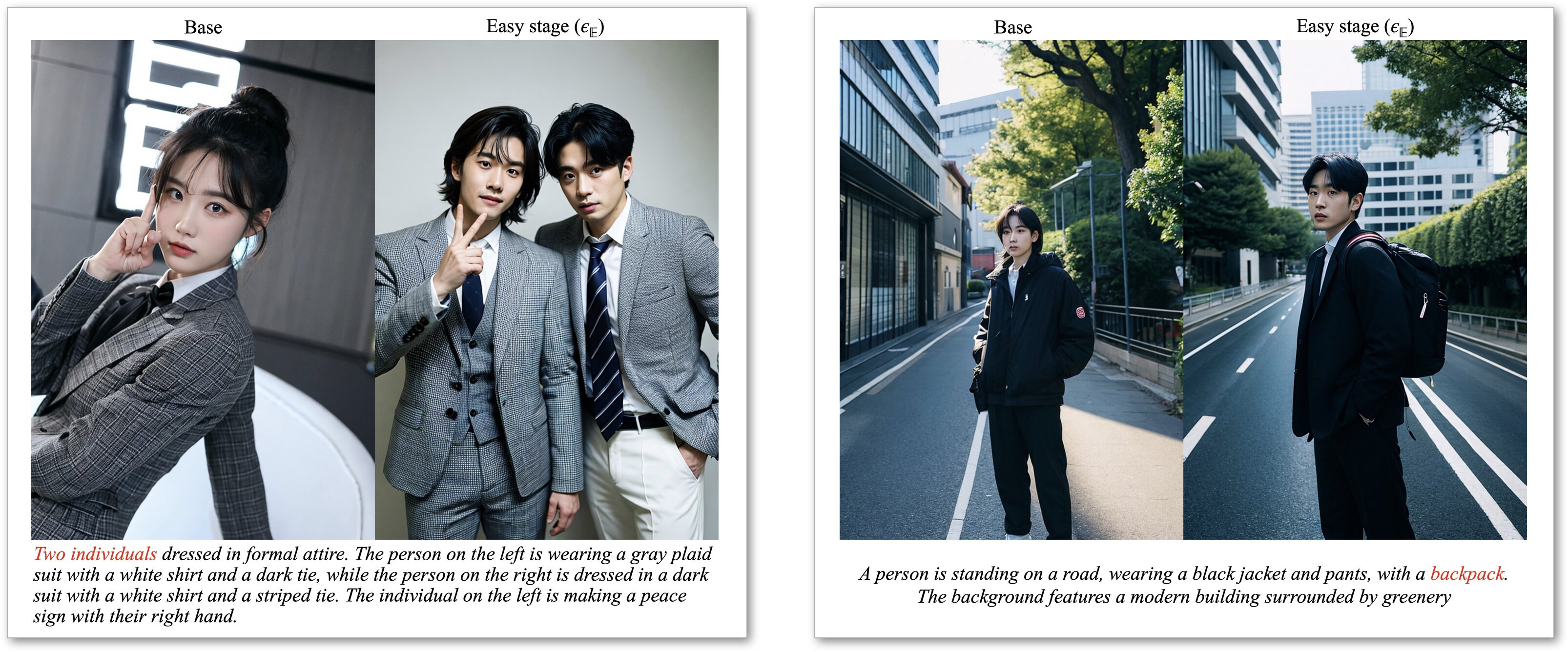}
    \vspace{-6.0mm}
        \caption{\textbf{Qualitative advancements achieved through the easy stage.} We enhance the base model through the easy stage to generate images that better align with human preferences. Specifically, the model is improved to produce images with undistorted poses and anatomies and stronger alignment with the given prompts.}
    \label{fig:sm_base_vs_easy}
\end{figure*}

In the easy stage, we refine the base model to generate images that align more closely with human preferences as shown in Figure~\ref{fig:sm_base_vs_easy}. Specifically, the model is improved to produce images with undistorted poses and anatomies and stronger alignment with the given prompts.

\subsection{Image Pool with AI Feedback}
\label{subsec:sm_image_pool_with_AI_feedback}
In our manuscript, we propose a method for selecting winning and losing images from the image pool using AI feedback (PickScore~\cite{kirstain2024pick}). This method assumes that a larger PickScore difference between the winning and losing images indicates greater semantic differences, which are crucial for enhancing the model through DPO and align better with actual human preferences. As shown in Figure~\ref{fig:sm_easy_dataset}, comparing the image with the highest PickScore to the image with the $l$-th highest PickScore reveals that the semantic differences between the two images (e.g., anatomy, pose, and text-image alignment) become more pronounced as $l$ increases. By choosing the images with the highest and the 20th highest PickScores as the winning and losing images, respectively, we accentuate the semantic differences between them, better reflecting human preferences.

\begin{figure}
    \centering
    \includegraphics[width=1.0\linewidth]{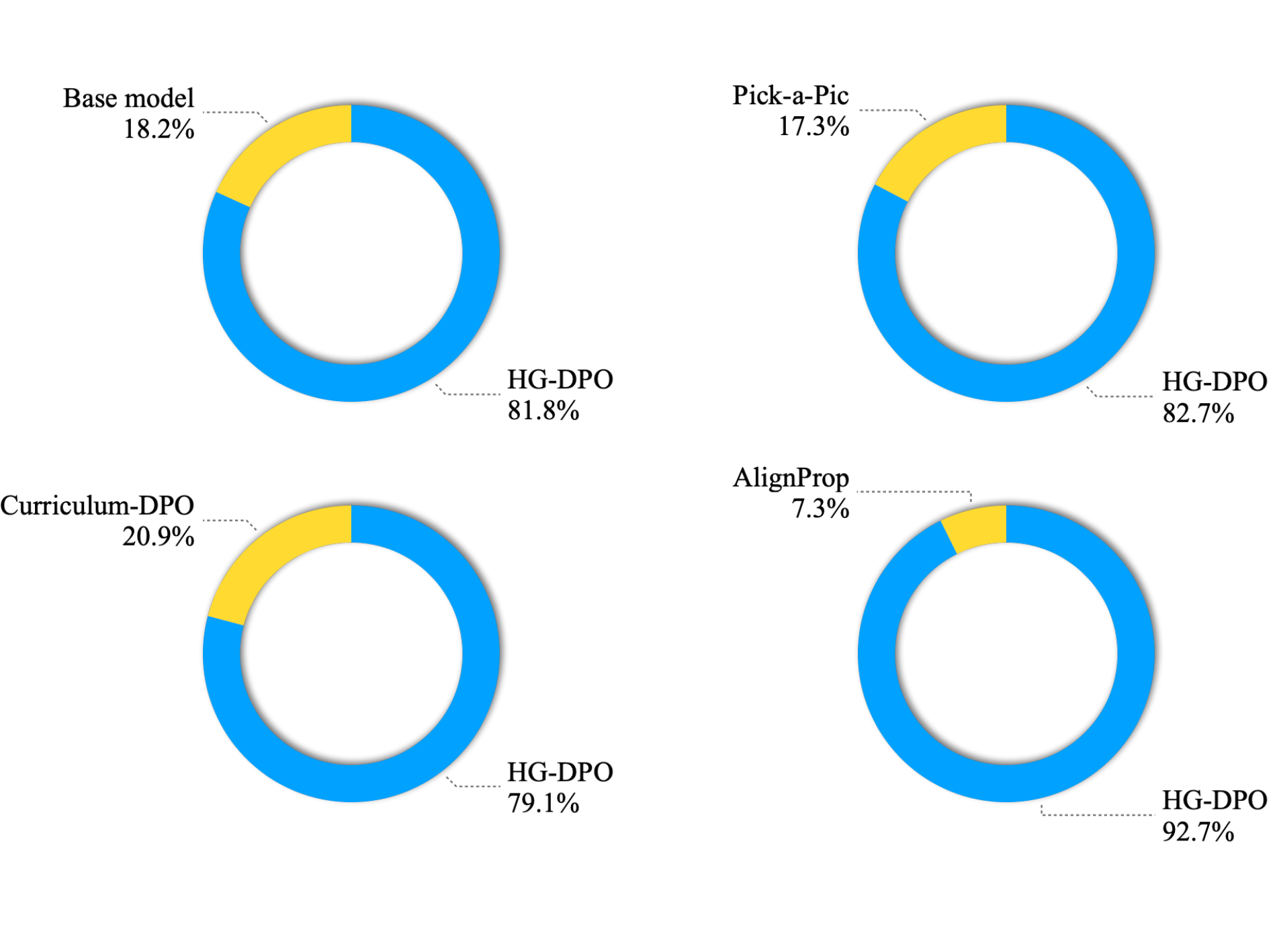}
    \vspace{-8.0mm}
    \caption{\textbf{User studies comparing HG-DPO and baselines.} HG-DPO demonstrates superior performance compared to the base model and previous approaches in human evaluation. Participants were tasked with choosing the image that exhibited higher realism and better alignment with the given prompt from the outputs of the two models. The detailed process for conducting the user study is described in Section~\ref{subsec:user_study}.}
    \label{fig:sm_user_study}
\end{figure}

\subsection{Statistics Matching Loss}
\label{subsec:sm_statistics_matching_loss}
In this section, we further analyze the statistics matching loss.

\subsubsection{Hypothesis test}
Here, we validate the hypothesis underlying the statistics matching loss, $\mathcal{L}_\textit{stat}$. Let us denote the model obtained by training $\epsilon_{base}$ through the easy stage without $\mathcal{L}_\textit{stat}$ as $\hat{\epsilon_\mathbb{E}}$. $\hat{\epsilon_\mathbb{E}}$ is a model that suffers from the color shift artifacts. As explained in our manuscript, we hypothesize that the cause of the color shift artifacts is the divergence between the latent statistics sampled by $\hat{\epsilon_\mathbb{E}}$ and those of $\epsilon_{base}$ during inference. $\mathcal{L}_\textit{stat}$ is designed to prevent such divergence based on this assumption.

To verify our hypothesis more directly, we design an inference-time statistics matching approach called \textit{latent adaptive normalization} (\textit{LAN}). If the gaps in the channel-wise statistics of the latents during inference cause the color shift artifacts, then eliminating those gaps should resolve those artifacts. 

Let $\hat{h}^{t-1}_\mathbb{E}$ and $h^{t-1}_{base}$ denote the latents sampled from the same random noise using $\hat{\epsilon_\mathbb{E}}$ and $\epsilon_{base}$ at inference time with timestep $t$, respectively. Formally, we define
\begin{align}
    \label{eq:sampled_latents_1}
    &\hat{h}^{t-1}_\mathbb{E}=\psi(h^t_\mathbb{E},p,t,\hat{\epsilon_\mathbb{E}}) \\
    \label{eq:sampled_latents_2}
    &h^{t-1}_{base}=\psi(h^t_{base},p,t,\epsilon_{base})
\end{align}
where $\psi$ denotes a inference-time latent sampler and $p$ denotes an inference prompt. Then, we define LAN as follows:
\begin{align}
    \label{eq:latent_adaptive_norm}
    h_{\mathbb{E}}^{t-1}=\bigg(\frac{\hat{h}_{\mathbb{E}}^{t-1}-\mu(\hat{h}_{\mathbb{E}}^{t-1})}{\sigma(\hat{h}_{\mathbb{E}}^{t-1})}\bigg)\sigma(h_{base}^{t-1})+\mu(h_{base}^{t-1})
\end{align}
where $\mu$ and $\sigma$ calculate the channel-wise mean and standard deviation from the input, respectively. $h_{\mathbb{E}}^{t-1}$ is used in Eq.~\eqref{eq:sampled_latents_1} of the supplementary material at the next inference timestep. 

Table~\ref{table:sm_quantitative_ablation_easy_stage} reveals that $N>2$ ($\hat{\epsilon_\mathbb{E}}$) + LAN significantly reduces the hue distance compared to $N>2$ ($\hat{\epsilon_\mathbb{E}}$). Furthermore, $N>2$ ($\hat{\epsilon_\mathbb{E}}$) + LAN achieves comparable performance to $N>2$ ($\hat{\epsilon_\mathbb{E}}$) in human preference metrics (P-Score, HPS, I-Reward, and AES) and image-text alignment (CLIP). These findings validate LAN's effectiveness in addressing the color shift artifacts and support the hypothesis underlying the design of $\mathcal{L}_\textit{stat}$.

However, because LAN requires additional sampling from $\epsilon_{base}$ during inference, it incurs higher computational costs during inference compared to $N>2 + \mathcal{L}_\textit{stat}$. For this reason, we propose $\mathcal{L}_\textit{stat}$ as a more computationally efficient solution to mitigate the color shift artifacts.

\begin{figure*}
    \centering
    \includegraphics[width=0.98\linewidth]{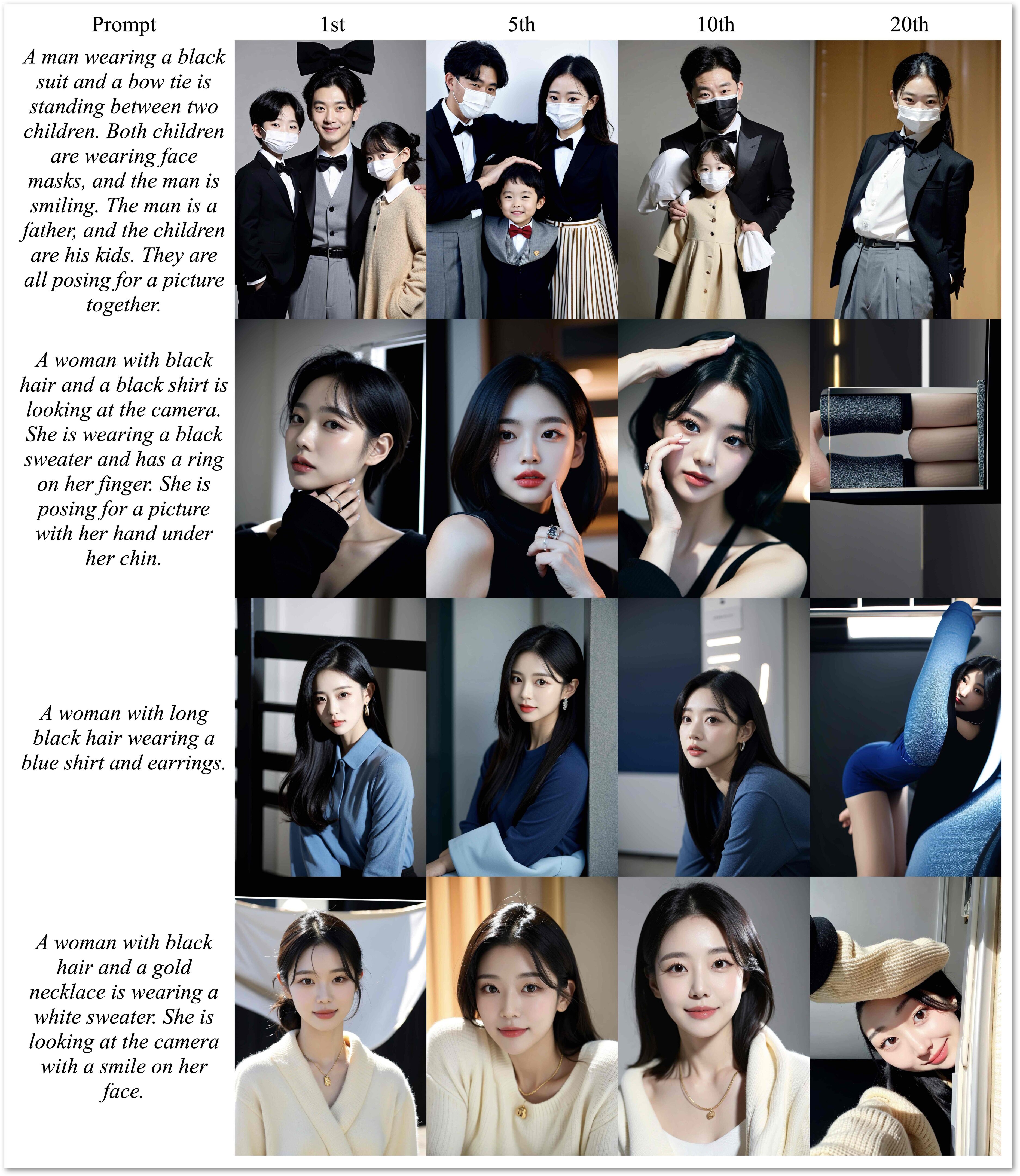}
    \caption{\textbf{Visualization of the image pool.} This figure shows the image pool with the size of 20 for the prompt in the leftmost column. The column labeled as 1st contains images with the highest PickScore, while the column labeled as 20th contains images with the 20th highest PickScore, i.e., the lowest PickScore, in the image pool. By selecting the image with the highest PickScore from this image pool as the winning image and the image with the 20th highest PickScore as the losing image, we magnify the semantic differences between the winning and losing images.}
    \label{fig:sm_easy_dataset}
\end{figure*}

\subsubsection{What causes the color shift artifacts?}
The color shift artifacts arise from the deviation of the channel-wise statistics of latents sampled using $\hat{\epsilon_\mathbb{E}}$ from those sampled using $\epsilon_{base}$, as demonstrated by the effectiveness of LAN in the previous paragraph. Here, to find the cause of this deviation, we further analyze the winning and losing images used in the easy stage. Specifically, we calculate the cosine distance of channel-wise statistics of encoded latents of winning and losing images. In Table~\ref{table:sm_cosine_distance}, the results reveal that the cosine distance between the latents' means for the winning and losing images is \textit{0.2035}, while the cosine distance for their standard deviations is \textit{0.005}. Since DPO trains the model to learn the differences between winning and losing images, it can be inferred that the differences in the channel-wise \textbf{mean} values of latents present in the dataset are also learned by the model. This can encourage the model to shift the mean of the sampled latents far from that of the losing image and close to that of the winning image.

\subsubsection{Why is it sufficient to match only the mean?}
$\mathcal{L}_\textit{stat}$ mitigates the color shift by preventing the aforementioned mean shift through the mean matching loss. Interestingly, as reported in the previous paragraph, we can observe that the cosine distance of standard deviation between the latents of winning and losing images is close to zero. We believe this is why matching only the mean in $\mathcal{L}_\textit{stat}$ is sufficient to prevent the color shift artifacts.

\subsubsection{Importance of the statistics matching loss}
\begin{figure}
    \centering
    \includegraphics[width=\linewidth]{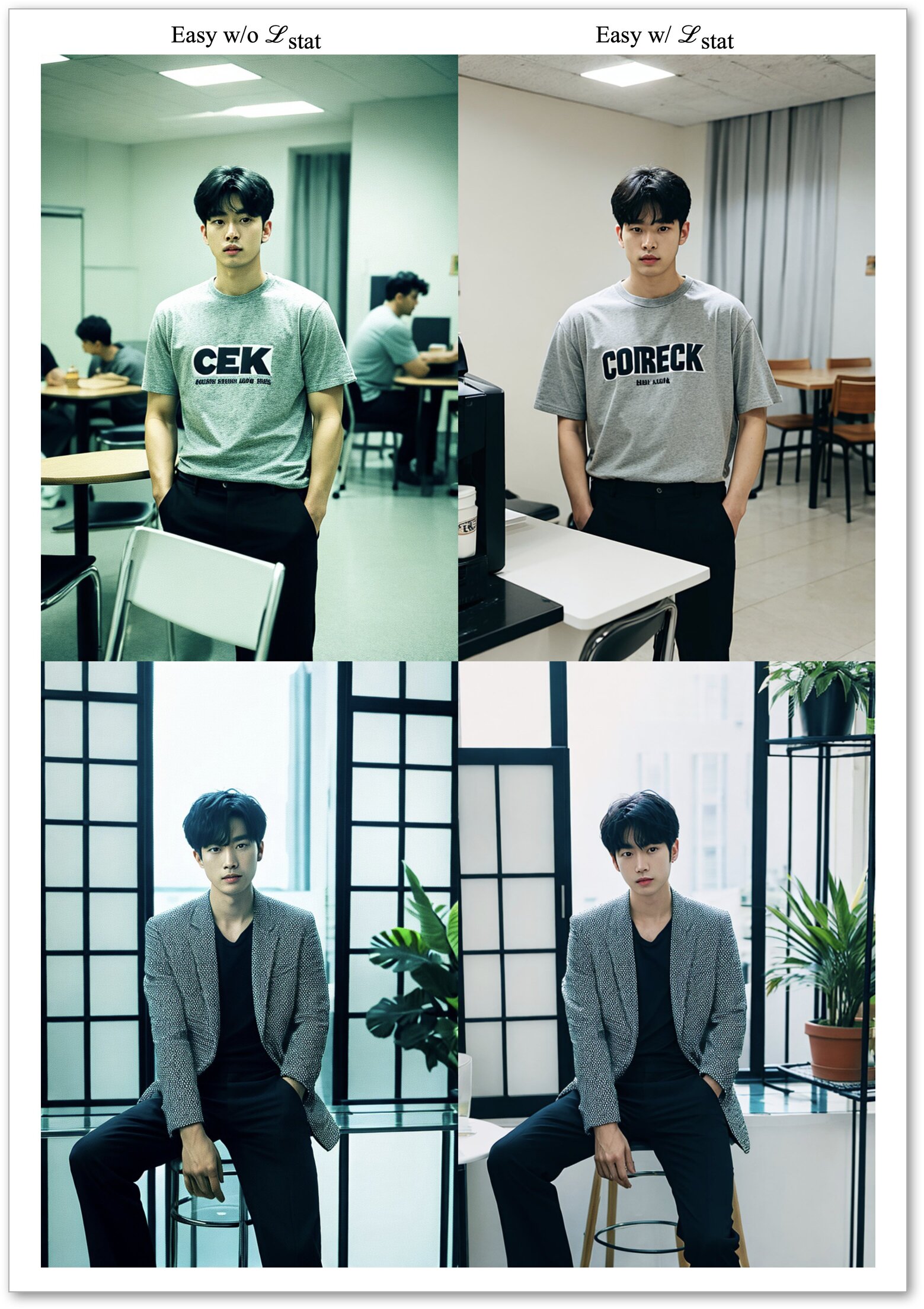}
    \caption{\textbf{Qualitative enhancements achieved with the statistics matching loss.} The statistics matching loss effectively removes the color shift artifacts, leading to the generation of significantly more natural images.}
    \label{fig:sm_wo_stat_vs_w_stat}
\end{figure}

As illustrated in Figure~\ref{fig:sm_wo_stat_vs_w_stat}, the absence of $\mathcal{L}_\textit{stat}$ results in generated images appearing unnatural due to the color shift artifacts. Incorporating $\mathcal{L}_\textit{stat}$ effectively eliminates these artifacts, producing noticeably more natural images.
\begin{table*}[t]
    \begin{center}
        \scalebox{0.96}{
        \small
            \begin{tabular}
                {lc@{~~~~}c@{~~~~}c@{~~~~}c@{~~~~}c@{~~~~}c@{~~~~}c@{~~~~}c@{~~~~}c@{~~~~}c}
                \toprule
                Model & P-Score ($\uparrow$) & HPS ($\uparrow$) & I-Reward ($\uparrow$) & AES ($\uparrow$) & CLIP ($\uparrow$) & FID ($\downarrow$) & CI-Q ($\uparrow$) & CI-S ($\uparrow$) & ATHEC ($\uparrow$) & Hue ($\downarrow$)\\
                \midrule
                Base ($\epsilon_{base}$) & 21.7364 & 0.2819 & -0.0665 & 6.1061 & 29.72 & 37.34 & \underline{0.9058} & \textbf{0.9573} & 18.73 & -\\
                \hdashline[3pt/1.5pt]
                $N = 2$ & 22.1939 & 0.2854 & 0.3610 & 6.1408 & 30.66 & \textbf{34.44} & 0.8887 & 0.9472 & 18.96 & \textbf{10.24}\\
                $N > 2$ ($\hat{\epsilon_\mathbb{E}}$) & \underline{22.5688} & 0.2872 & \textbf{0.7830} & \textbf{6.2544} & 31.50 & 37.29 & 0.8879 & 0.9471 & \textbf{27.20} & 98.54\\
                $N > 2$ ($\hat{\epsilon_\mathbb{E}}$) + $\beta \uparrow$ & 22.2506 & 0.2864 & 0.5435 & 6.1129 & 31.30 & \underline{36.00} & 0.8416 & 0.9141 & 19.17 & 23.77\\
                $N > 2$ ($\hat{\epsilon_\mathbb{E}}$) + LAN & \textbf{22.6474} & \textbf{0.2885} & \underline{0.7677} & \underline{6.1899} & \textbf{31.60} & 37.08 & \textbf{0.9086} & 0.9521 & 18.65 & \underline{16.13} \\
                \rowcolor{pastelblue!30}
                $N > 2$ + $\mathcal{L}_\textit{stat}$ ($\epsilon_\mathbb{E}$) & 22.5384 & \underline{0.2878} & 0.7146 & 6.1775 & \underline{31.56} & \underline{36.00} & 0.9057 & \underline{0.9547} & \underline{19.58} & 27.94\\
                \bottomrule
            \end{tabular}
        }
    \end{center}
    \vspace{-4.0mm}
    \caption{\textbf{Quantitative analysis of the easy stage.} For $\mathcal{D}_\mathbb{E}$, $N=2$ generates exactly two images per prompt, while $N>2$ builds an image pool. $N > 2 + \beta \uparrow$, $N > 2 +$ LAN, and $N > 2 + \mathcal{L}_\textit{stat}$ add regularization to address the color shift artifacts in $N>2$. Specifically, $N>2 + \beta \uparrow$ applies a higher $\beta$, which is a strength of the original regularization in $\mathcal{L}_\textit{D-DPO}$, $N > 2 +$ LAN applies latent adaptive normalization (Section~\ref{subsec:sm_statistics_matching_loss}), and $N > 2 + \mathcal{L}_\textit{stat}$ integrates $\mathcal{L}_\textit{stat}$. \textbf{Bold} text and \underline{underlined} text indicate the best and second-best results, respectively. The row corresponding to the proposed training configuration in the easy stage is highlighted in blue.
    }
    \label{table:sm_quantitative_ablation_easy_stage}
\end{table*}

\begin{table}
    \begin{center}
        \scalebox{1.0}{
        \small
            \begin{tabular}
                {lcc}
                \toprule
                & Mean & Standard deviation \\
                \midrule
                Cosine distance & 0.2035 & 0.0005 \\
                \bottomrule
            \end{tabular}
        }
    \end{center}
    \vspace{-4.0mm}
    \caption{\textbf{Difference of channel-wise statistics between winning and losing images.} Cosine distance of channel-wise statistics of encoded latents of winning and losing images. For the encoding, we use the encoder of VAE~\cite{kingma2013auto} used in HG-DPO.}
    \label{table:sm_cosine_distance}
\end{table}

\begin{figure*}[t]
    \centering
    \includegraphics[width=1.0\linewidth]{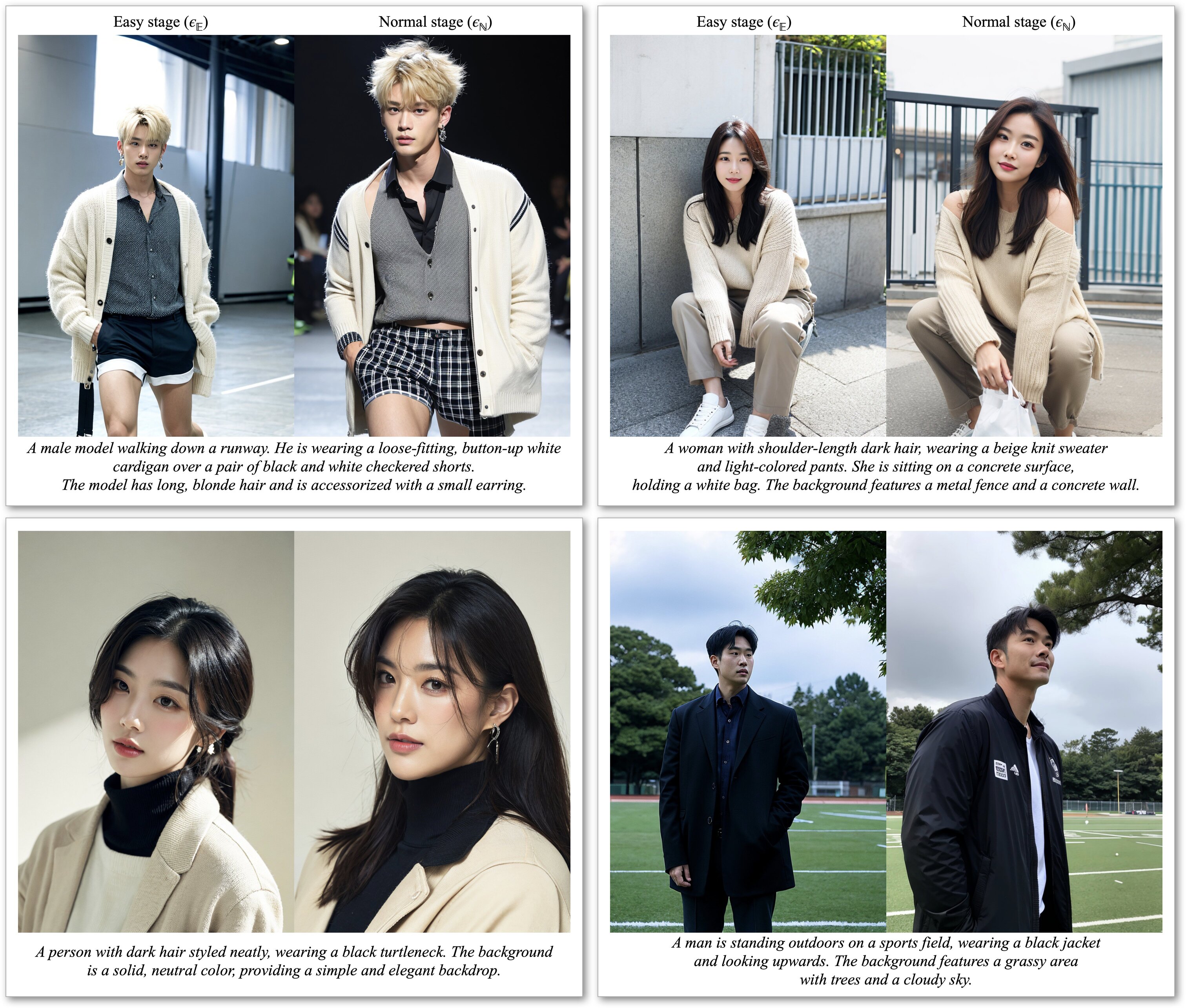}
    \vspace{-6.0mm}
    \caption{\textbf{Qualitative advancements achieved through the normal stage.} $\epsilon_\mathbb{N}$, derived by refining $\epsilon_\mathbb{E}$ through the normal stage, generates images with more \textbf{realistic compositions and poses} compared to $\epsilon_\mathbb{E}$.}
    \label{fig:sm_easy_vs_normal}
\end{figure*}

\begin{figure*}
    \centering
    \includegraphics[width=0.98\linewidth]{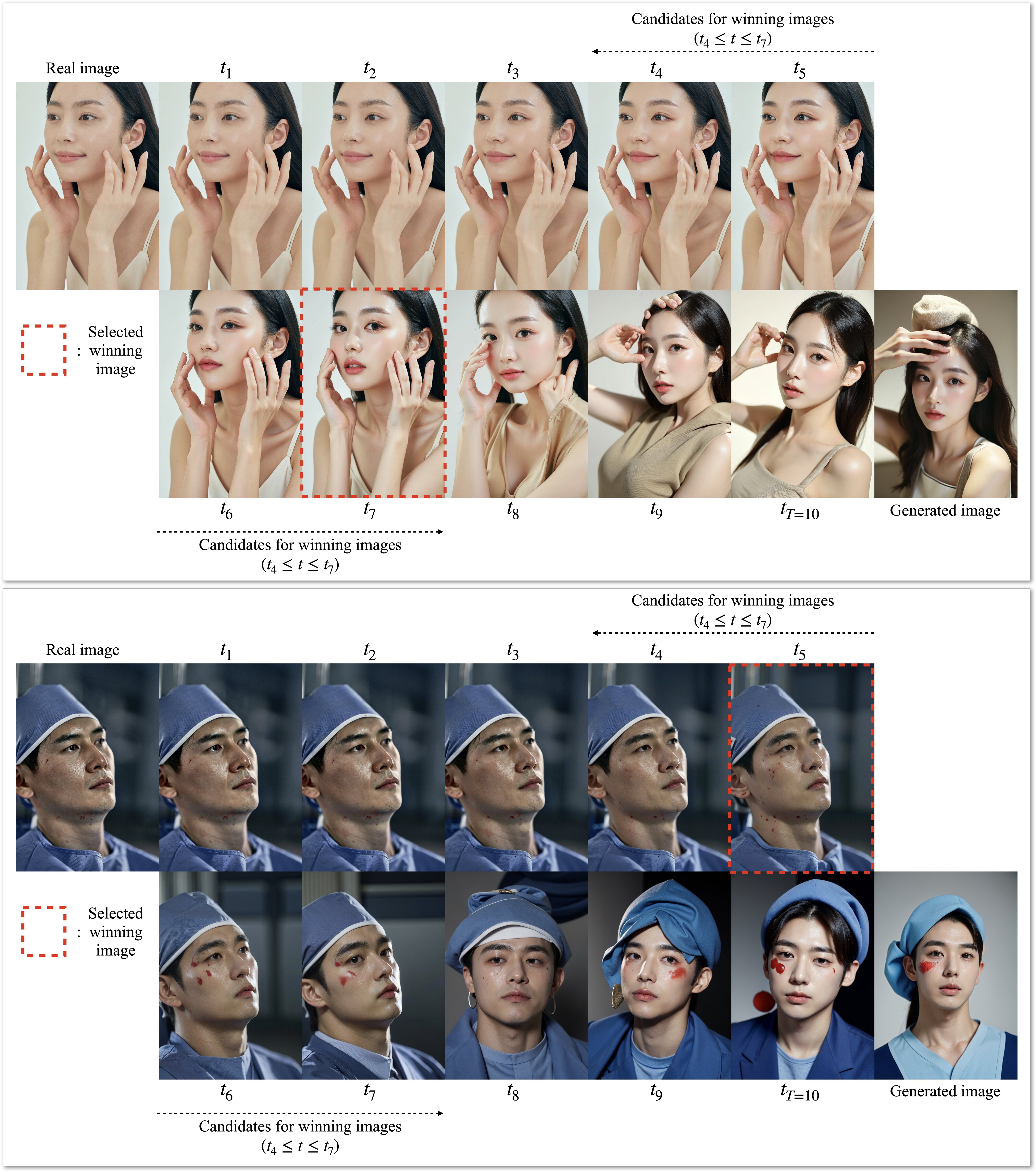}
    \caption{\textbf{Visualization of the intermediate domains.} The images labeled $t_1$ to $t_T$ are reconstructed from real images using the \textit{SDRecon} operation. The image labeled generated image is produced via text-to-image generation based on the caption of the real image. As the labels progress toward $t_T$, SDRecon applies increasingly stronger noise to the real image, causing it to lose more of its original characteristics and resemble the generated image more closely. For the normal stage, we select four intermediate domains, $t_4$ to $t_7$, as candidates for winning images, because they maintain the realistic pose of the real image while adopting the fine details typical of the generated image. The image with the highest PickScore among these candidates is chosen as the winning image.}
    \label{fig:sm_intermediate_domains}
\end{figure*}

\section{Additional Analysis on the Normal Stage}
\label{sec:sm_additional_analysis_on_the_normal_stage}
In this section, we present additional experimental results and further analysis of the normal stage.

\subsection{Effectiveness of the Normal Stage}
\label{subsec:sm_effectiveness_of_the_normal_stage}
We further explore the role of the normal stage, which refines $\epsilon_\mathbb{E}$, derived from the easy stage, to produce $\epsilon_\mathbb{N}$. While the easy stage enables $\epsilon_\mathbb{E}$ to generate images aligned with human preferences resulting in undistorted anatomical features and poses, they still fall short of achieving the realism found in real human portrait images. For example, as shown in Figure~\ref{fig:sm_easy_vs_normal}, although the poses are largely free from distortion, they still appear somewhat unnatural compared to those in real photographs. The normal stage enhances $\epsilon_\mathbb{E}$ by improving its ability to generate compositions and poses that are not only distortion-free but also realistic, closely mirroring those found in the real dataset. Figure~\ref{fig:sm_easy_vs_normal} illustrates that $\epsilon_\mathbb{N}$ achieves significantly more realistic compositions and poses, derived from real human portrait images, than $\epsilon_\mathbb{E}$.

\subsection{Intermediate Domains}
\label{subsec:sm_intermediate domains}
In the normal stage, we introduce intermediate domains for winning images. Figure~\ref{fig:sm_intermediate_domains} illustrates the outcomes of the \textit{SDRecon} operation used to create these intermediate domains, along with the winning images employed during the normal stage.

\subsubsection{Intermediate domains with SDRecon}
As shown in Figure~\ref{fig:sm_intermediate_domains}, we use 10 intermediate domains, labeled from $t_1$ to $t_T$. While $t_1$ is nearly identical to a real image, $t_T$ resembles a generated image, retaining little of the real image's original features. As the transition progresses from $t_1$ to $t_T$, the characteristics of the real image gradually fade, increasingly resembling those of a generated image. Specifically, fine-detail information is lost first, followed by the loss of pose information.

\subsubsection{Winning images from the intermediate domains}
As depicted in Figure~\ref{fig:sm_intermediate_domains}, we select four intermediate domains, $t_4$ through $t_7$, as candidates for the winning images in the normal stage. This is because the our qualitative analysis reveals that these domains generally retain the realistic pose of the real image while exhibiting fine details resembling those of generated images. Among these candidates, the image with the highest PickScore~\cite{kirstain2024pick} is chosen as the winning image.

\section{Additional Analysis on the Hard Stage}
\label{sec:sm_additional_analysis_on_the_hard_stage}
In this section, we present additional experimental results and further analysis of the hard stage.

\subsection{Effectiveness of the Hard Stage}
\label{subsec:sm_effectiveness_of_the_hard_stage}
\begin{figure*}
    \centering
    \includegraphics[width=\linewidth]{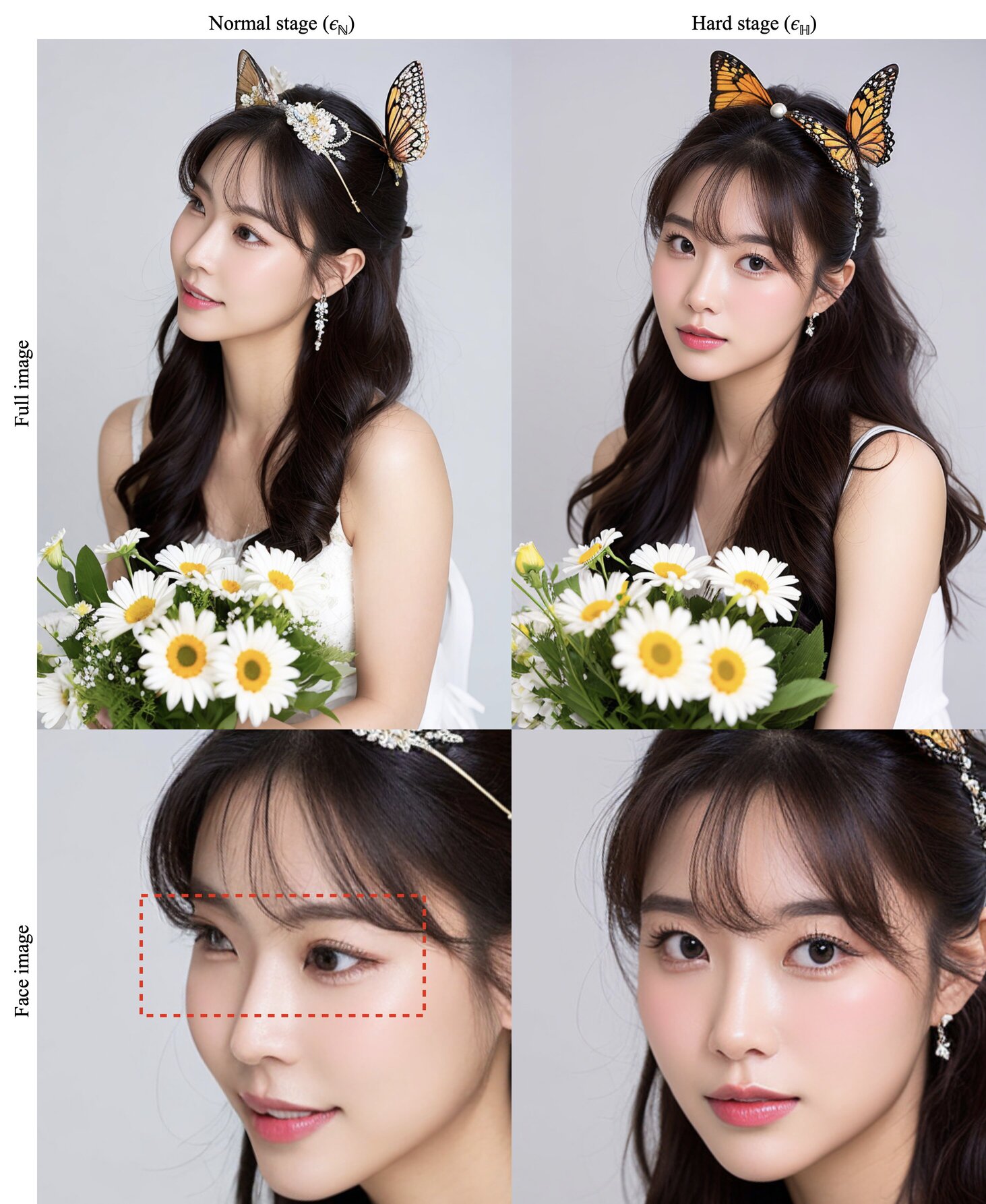}
    \caption{\textbf{Qualitative advancements achieved through the hard stage.} $\epsilon_\mathbb{H}$, derived by refining $\epsilon_\mathbb{N}$ through the hard stage, generates finer details, especially more realistic depictions of the \textbf{eyes}, compared to $\epsilon_\mathbb{N}$ as shown in the red box.}
    \label{fig:sm_normal_vs_hard_eyes_3}
\end{figure*}

\begin{figure*}
    \centering
    \includegraphics[width=\linewidth]{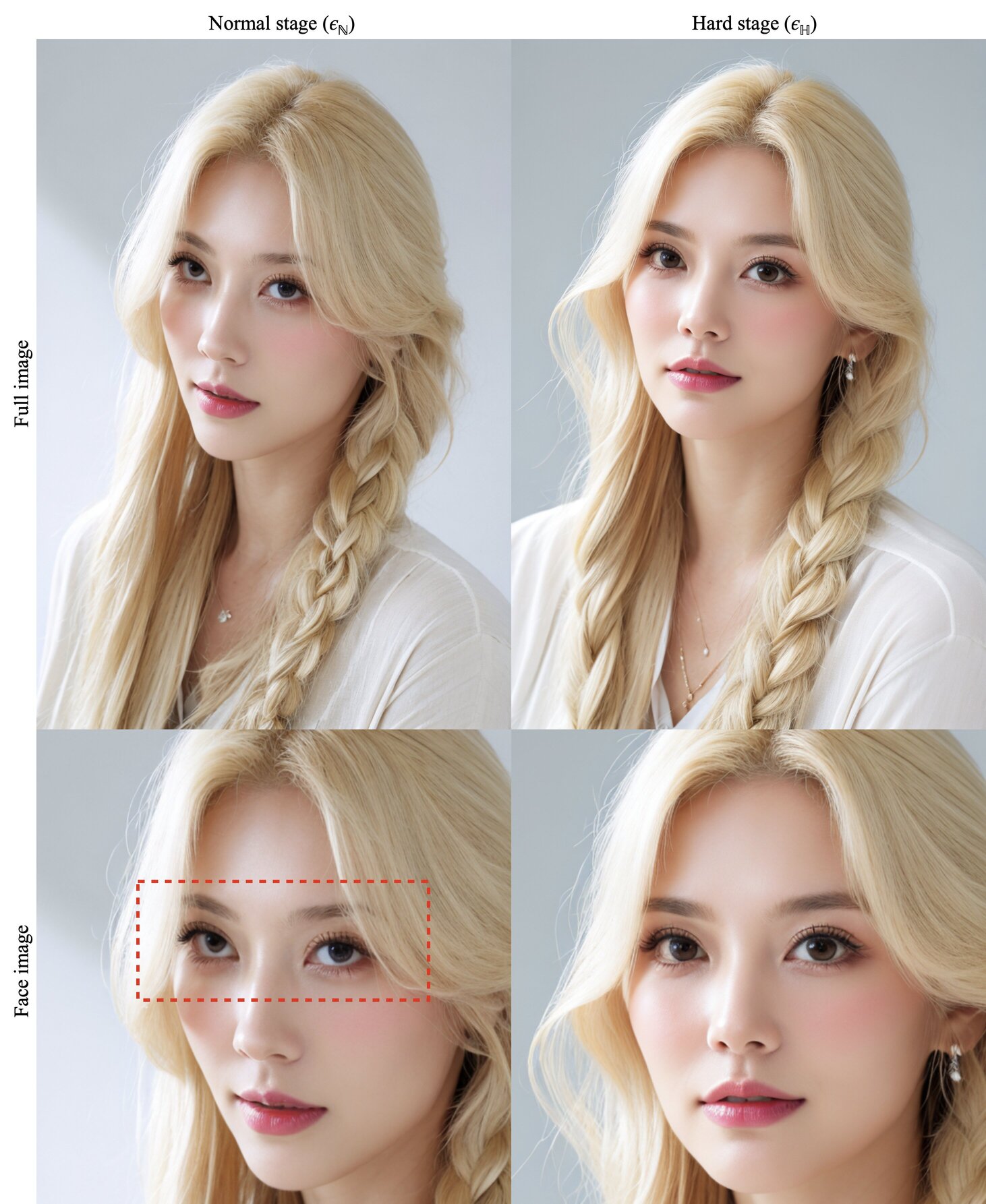}
    \caption{\textbf{Qualitative advancements achieved through the hard stage.} $\epsilon_\mathbb{H}$, derived by refining $\epsilon_\mathbb{N}$ through the hard stage, generates finer details, especially more realistic depictions of the \textbf{gaze}, compared to $\epsilon_\mathbb{N}$ as shown in the red box.}
    \label{fig:sm_normal_vs_hard_gaze_1}
\end{figure*}

\begin{figure*}
    \centering
    \includegraphics[width=\linewidth]{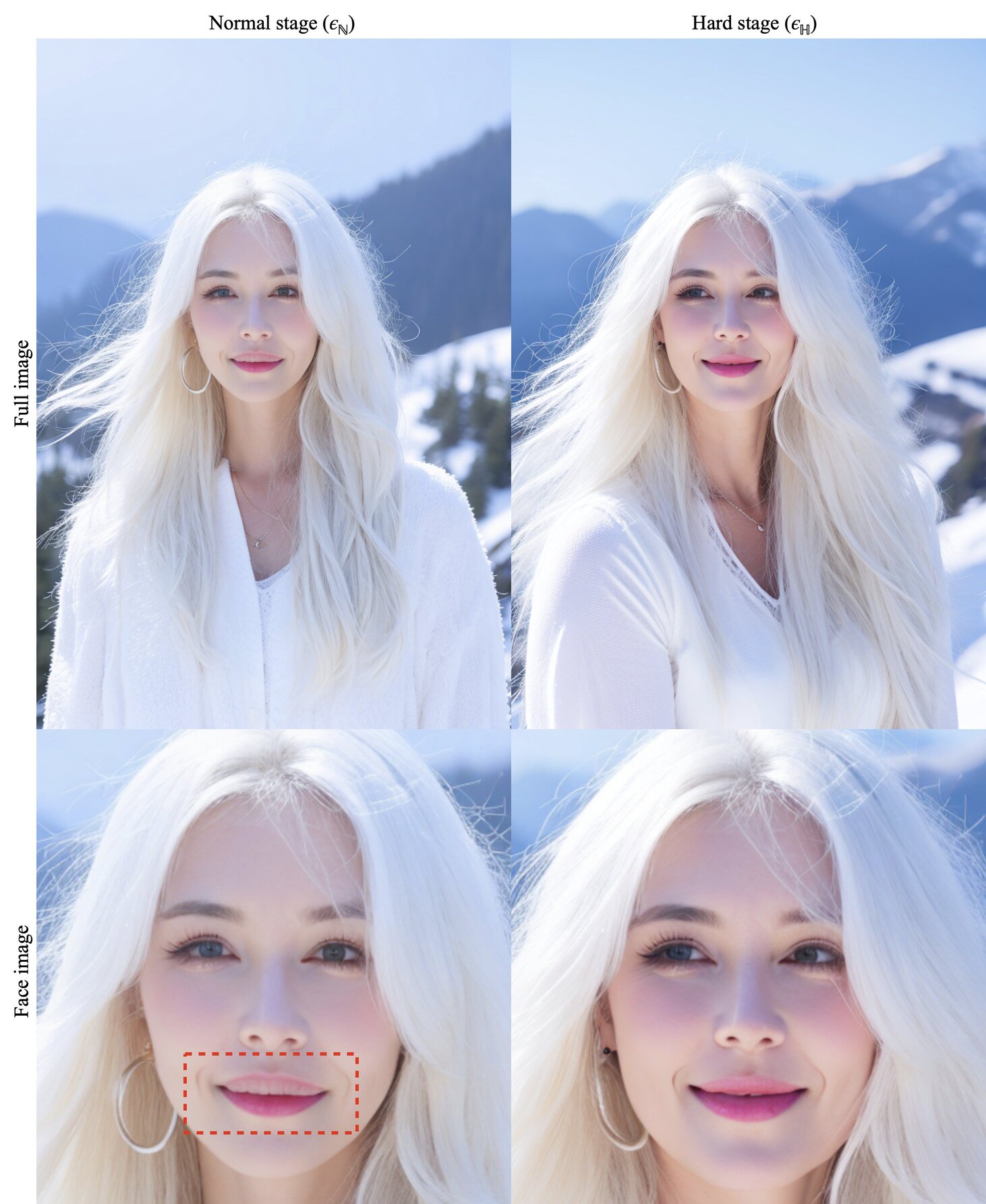}
    \caption{\textbf{Qualitative advancements achieved through the hard stage.} $\epsilon_\mathbb{H}$, derived by refining $\epsilon_\mathbb{N}$ through the hard stage, generates finer details, especially more realistic depictions of the \textbf{lips}, compared to $\epsilon_\mathbb{N}$ as shown in the red box.}
    \label{fig:sm_normal_vs_hard_tooth_1}
\end{figure*}

\begin{figure*}
    \centering
    \includegraphics[width=\linewidth]{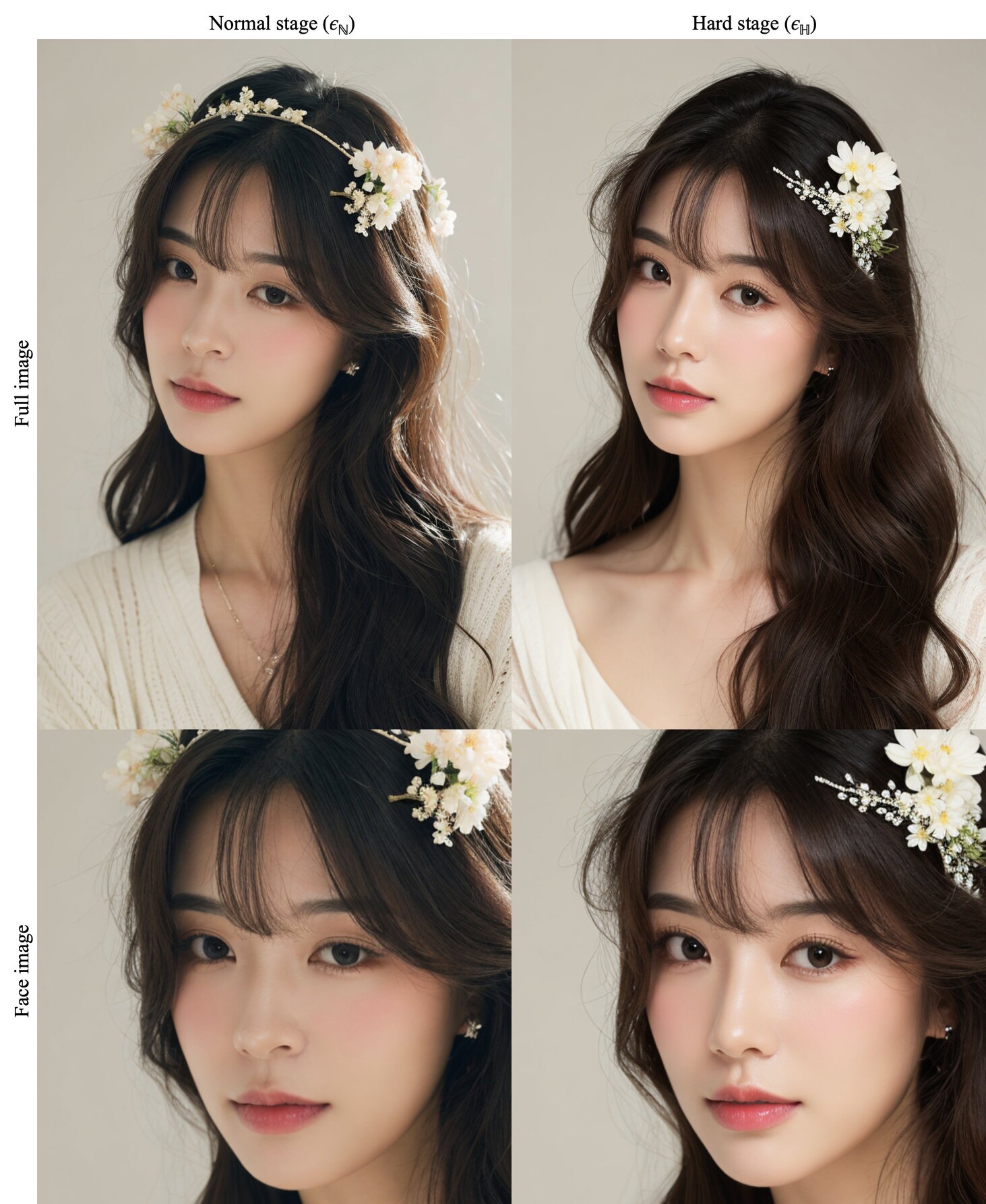}
    \caption{\textbf{Qualitative advancements achieved through the hard stage.} $\epsilon_\mathbb{H}$, derived by refining $\epsilon_\mathbb{N}$ through the hard stage, generates \textbf{sharper} images with improved fine details, particularly exhibiting more vivid and realistic \textbf{shading}, compared to $\epsilon_\mathbb{N}$.}
    \label{fig:sm_normal_vs_hard_sharpness_1}
\end{figure*}

We investigate the impact of the hard stage, which refines $\epsilon_\mathbb{N}$, obtained from the normal stage, to produce $\epsilon_\mathbb{H}$. While $\epsilon_\mathbb{N}$ achieves realistic composition and poses during the normal stage, it struggles to generate fine details. For instance, as shown in Figures~\ref{fig:sm_normal_vs_hard_eyes_3},~\ref{fig:sm_normal_vs_hard_gaze_1},~\ref{fig:sm_normal_vs_hard_tooth_1}, and \ref{fig:sm_normal_vs_hard_sharpness_1}, $\epsilon_\mathbb{N}$ 1) fails to accurately depict fine facial features such as eyes and lips, 2) requires better shading, and 3) suffers from image blurriness. Although these details may seem minor, they play a crucial role in achieving overall image realism. The hard stage addresses these limitations by enhancing $\epsilon_\mathbb{N}$, resulting in $\epsilon_\mathbb{H}$, which excels in generating realistic fine details. Figures~\ref{fig:sm_normal_vs_hard_eyes_3},~\ref{fig:sm_normal_vs_hard_gaze_1},~\ref{fig:sm_normal_vs_hard_tooth_1}, and \ref{fig:sm_normal_vs_hard_sharpness_1} illustrate that $\epsilon_\mathbb{H}$ effectively produces fine details that $\epsilon_\mathbb{N}$ cannot, significantly improving image realism. As shown in Figure~\ref{fig:sm_user_study_normal_vs_hard}, in a user study comparing $\epsilon_\mathbb{N}$ and $\epsilon_\mathbb{H}$, $\epsilon_\mathbb{H}$ is rated higher, further demonstrating its effectiveness.

\subsection{Winning Images of the Hard Stage}
In the hard stage, we employ images from the intermediate domain $t_1$ as winning images instead of real images. As illustrated in Figure~\ref{fig:sm_intermediate_domains}, images from the intermediate domain $t_1$ are visually nearly indistinguishable from real human portrait images, making this approach effectively comparable to using real images directly as winning images. This choice is motivated by the observation that, while real images and intermediate domain $t_1$ images appear almost identical to the human eye, utilizing intermediate domain images leads to slightly better quantitative performance. Specifically, as demonstrated in Table~\ref{table:sm_quantitative_hard_winning_images}, the model trained with intermediate domain $t_1$ images achieves results similar to those trained with real images, with a slight improvement in CI-Q scores.

\subsection{Effectiveness of the Enhanced Text Encoder}
\begin{figure}
    \centering
    \includegraphics[width=\linewidth]{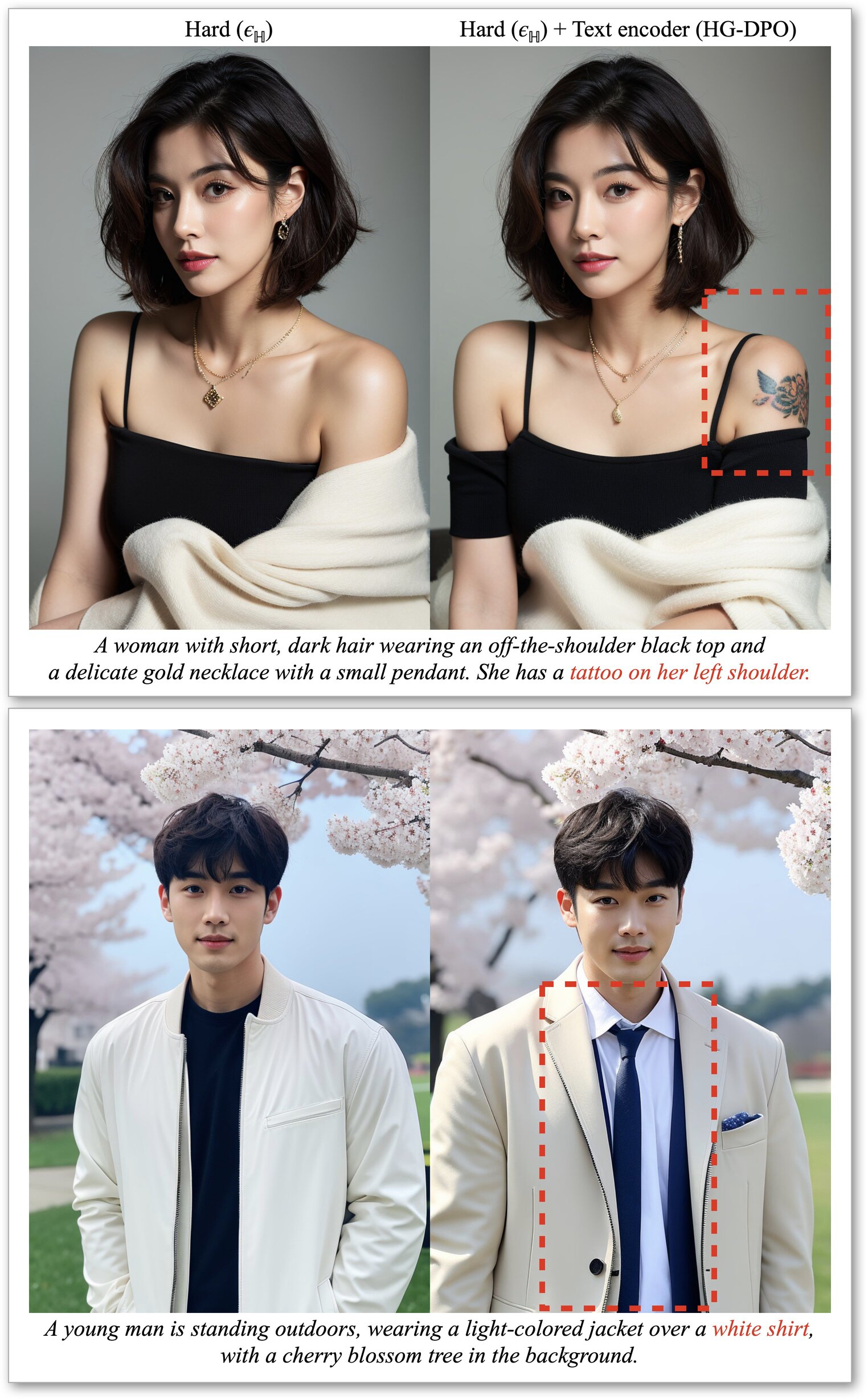}
    \caption{\textbf{Qualitative advancements achieved through the text encoder enhancement.} By training the text encoder through the easy stage and incorporating it with $\epsilon_\mathbb{H}$ during inference, we achieve improved image-text alignment compared to using $\epsilon_\mathbb{H}$ alone. Moreover, the use of the enhanced text encoder does not compromise the image quality produced by $\epsilon_\mathbb{H}$.}
    \label{fig:sm_hard_vs_hard_te}
\end{figure}

We train the text encoder during the easy stage to enhance image-text alignment and employ it alongside $\epsilon_\mathbb{H}$, derived from the hard stage, during inference. As shown in Figure~\ref{fig:sm_hard_vs_hard_te}, the enhanced text encoder effectively improves image-text alignment without compromising the image quality achieved by $\epsilon_\mathbb{H}$.

\section{Limitations}
\label{sec:sm_limitations_and_future_work}
\begin{figure}
    \centering
    \includegraphics[width=\linewidth]{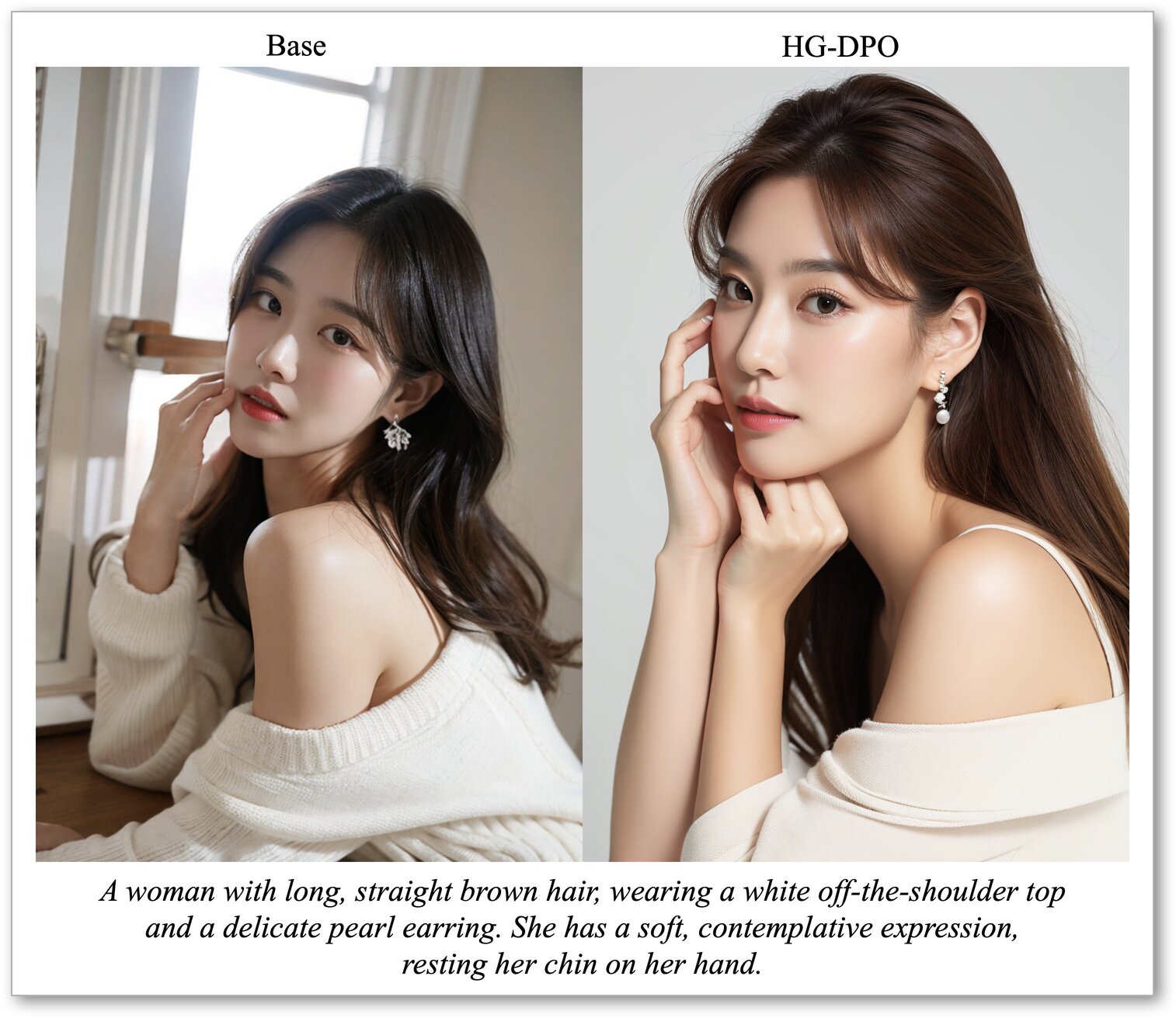}
    \caption{\textbf{Qualitative results illustrating the limitations of HG-DPO.} While HG-DPO significantly improves the base model in generating more realistic human images, it still struggles to accurately synthesize fingers.}
    \label{fig:sm_finger}
\end{figure}

Through a three-stage training pipeline, HG-DPO enhances the base model to generate not only realistic anatomical features and poses but also fine details with greater realism. Despite these improvements, HG-DPO does not address the generation of realistic fingers. As shown in Figure~\ref{fig:sm_finger}, HG-DPO produces an image with sharper and more realistic fine details compared to the base model. However, the generated fingers remain notably unrealistic. 
\begin{table*}
    \begin{center}
        \scalebox{0.96}{
        \small
            \begin{tabular}
                {lc@{~~~~}c@{~~~~}c@{~~~~}c@{~~~~}c@{~~~~}c@{~~~~}c@{~~~~}c@{~~~~}c@{~~~~}c}
                \toprule
                Model & P-Score ($\uparrow$) & HPS ($\uparrow$) & I-Reward ($\uparrow$) & AES ($\uparrow$) & CLIP ($\uparrow$) & FID ($\downarrow$) & CI-Q ($\uparrow$) & CI-S ($\uparrow$) & ATHEC ($\uparrow$) \\
                \midrule
                Real & \textbf{22.4773} & 0.2857 & 0.5388 & 6.1953 & 30.99 & \textbf{28.56} & 0.9298 & \textbf{0.9885} & 29.13 \\
                \rowcolor{pastelblue!30}
                Intermediate $t_1$ & 22.4698 & \textbf{0.2867} & \textbf{0.5791} & \textbf{6.1955} & \textbf{31.15} & 28.66 & \textbf{0.9365} & 0.9859 & \textbf{30.08} \\
                \bottomrule
            \end{tabular}
        }
    \end{center}
    \vspace{-4.0mm}
    \caption{\textbf{Quantitative results based on the type of images used as winning images in the hard stage.} The row labeled \textit{Real} displays the results for the model trained with real images as winning images, while the row labeled \textit{Intermediate $t_1$} shows the results for the model trained using images from the intermediate domain $t_1$ as winning images. \textbf{Bold} text indicates the best results. The row corresponding to the proposed training configuration in the hard stage is highlighted in blue.}
    \label{table:sm_quantitative_hard_winning_images}
\end{table*}

\begin{figure}
    \centering
    \includegraphics[width=0.6\linewidth]{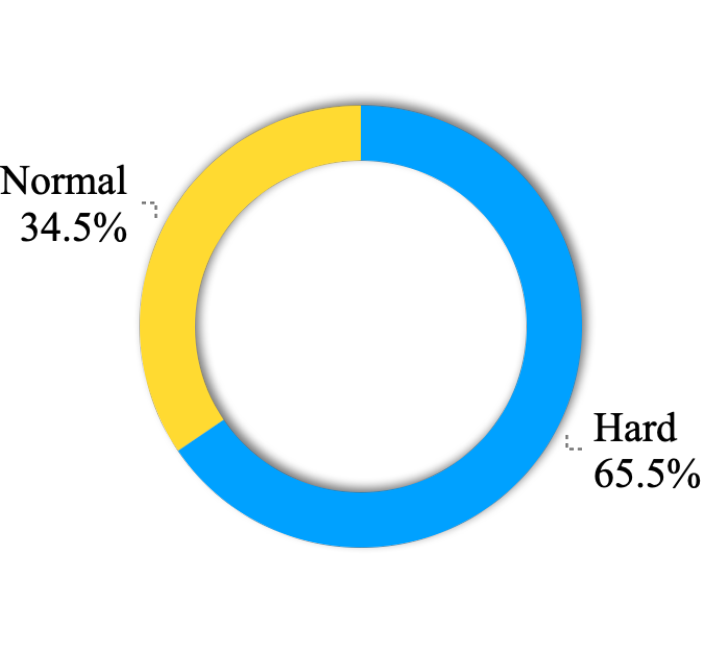}
    \caption{\textbf{User study comparing a model trained up to the normal stage ($\epsilon_\mathbb{N}$) with one trained through the hard stage ($\epsilon_\mathbb{H}$).} Participants were tasked with choosing the image that exhibited higher realism and better alignment with the given prompt from the outputs of the two models. The model trained through the hard stage achieves higher human evaluation scores due to its ability to generate finer details with greater realism compared to the model trained only up to the normal stage.}
    \label{fig:sm_user_study_normal_vs_hard}
\end{figure}

\section{Implementation Details}
\label{sec:implementation_details}
In this section, we provide implementation details on training and inference.

\subsection{Details on Supervised Fine-Tuning}
First, we introduce the method for obtaining $\epsilon_{base}$ through supervised fine-tuning. 

\paragraph{Text-to-image dataset.} We collected approximately 300k high-quality human images. Each image has a resolution of $704 \times 1024$. We use \texttt{LLaVa}~\cite{liu2023llava} to generate text prompts for all the collected images for training. This text-to-image dataset corresponds to $\mathcal{D}_\textrm{real}$ in our manuscript.

Furthermore, we use Qwen2-VL~\cite{wang2024qwen2} for visual question answering to analyze distribution of this dataset, which includes 40.7\% male and 59.3\% female, and 24.45\% child, 2.82\% teenager, 41.00\% youth, 31.61\% adult, and 0.12\% elderly. While the proportions of teenagers and elderly appear small, images in these groups may have been reasonably classified into adjacent categories (\eg, teenagers as child/youth, elderly as adult).

\paragraph{Architecture.} 
We employ Stable Diffusion 1.5 (SD1.5)~\cite{rombach2022high}, which is pre-trained with large text-to-image datasets, as our backbone model. More specifically, we use majicmix-v7~\cite{majicmix}, a fine-tuned model of SD1.5 specialized in human generation. We further fine-tune this backbone model with $\mathcal{D}_\textrm{real}$, to obtain our base model, $\epsilon_{base}$.

\paragraph{Loss function.} For fine-tuning, we use the noise prediction loss~\cite{ho2020denoising}. Also, we use DDPM noise scheduler~\cite{ho2020denoising} for the forward diffusion process during training.

\subsection{Details on HG-DPO Training}
In this section, we provide details on how to improve $\epsilon_{base}$ using HG-DPO. 

\subsubsection{Architecture}
\paragraph{U-Net.}
Instead of training the all parameters of $\epsilon_{base}$ through HG-DPO, we attach LoRA~\cite{hu2021lora} layers to the all linear layers in the attention modules and only train them. We set LoRA rank as 8.

\paragraph{Text encoder.}
When training the text encoder, we also attach LoRA~\cite{hu2021lora} layers to the all linear layers in the attention modules and only train them. For the text encoder, we set LoRA rank as 64.

\subsubsection{Loss function}
\paragraph{DPO loss.}
We adopt the objective function of Diffusion-DPO ($\mathcal{L}_\textit{D-DPO}$)~\cite{wallace2023diffusion} with $\beta=2500$. For $\mathcal{L}_\textit{D-DPO}$, we use DDPM noise scheduler~\cite{ho2020denoising} for the forward diffusion process. 

\paragraph{Statistics matching loss.}
For the statistics matching loss ($\mathcal{L}_\textit{stat}$), we set $\lambda_{stat}=10000$. Also, for the latent sampling in $\mathcal{L}_\textit{stat}$, we use DDPM sampler~\cite{ho2020denoising}. We tried DDIM sampler~\cite{song2020denoising}, but there was no significant difference. In addition, classifier-free guidance~\cite{ho2022classifier} is not used during the latent sampling in $\mathcal{L}_\textit{stat}$. 

\subsubsection{Optimization} 
For the optimization, we set the local batch size to four, which corresponds to the total batch size to 16 because we used four NVIDIA A100 GPUs. As an optimizer, we use the 8-bit Adam optimizer~\cite{dettmers20218} with $\beta_1$ and $\beta_2$ of the Adam optimizer to 0.9 and 0.999, respectively, and the learning rate to $1e-5$. Additionally, we utilize mixed precision for efficient training. For the easy, normal, and hard stages, we update the model for 300k, 20k, and 20k steps, respectively.

\subsubsection{Dataset}
\paragraph{Image pool.}
For the image pool generation, we simply use the prompt set from $\mathcal{D}_\textrm{real}$. Furthermore, as shown in Figure~\ref{fig:sm_easy_dataset}, we generate 20 images per prompt for the image pool, which corresponds to $N=20$ in our manuscript.

\paragraph{Intermediate domains.}
For the intermediate domains, we introduce 10 intermediate domains from $t_1$ to $t_{T=10}$ as shown in Figure~\ref{fig:sm_intermediate_domains}. These 10 domains are generated by evenly dividing the diffusion timesteps from 1 to 1000 into 10 intervals. Specifically, we set $t_1 = 100$, $t_2 = 200$, ..., $t_T = 1000$. Then, we set $t_r=t_4$ and $t_g=t_7$ for candidates of winning images as shown in Figure~\ref{fig:sm_intermediate_domains}.

\begin{figure}
    \centering
    \includegraphics[width=\linewidth]{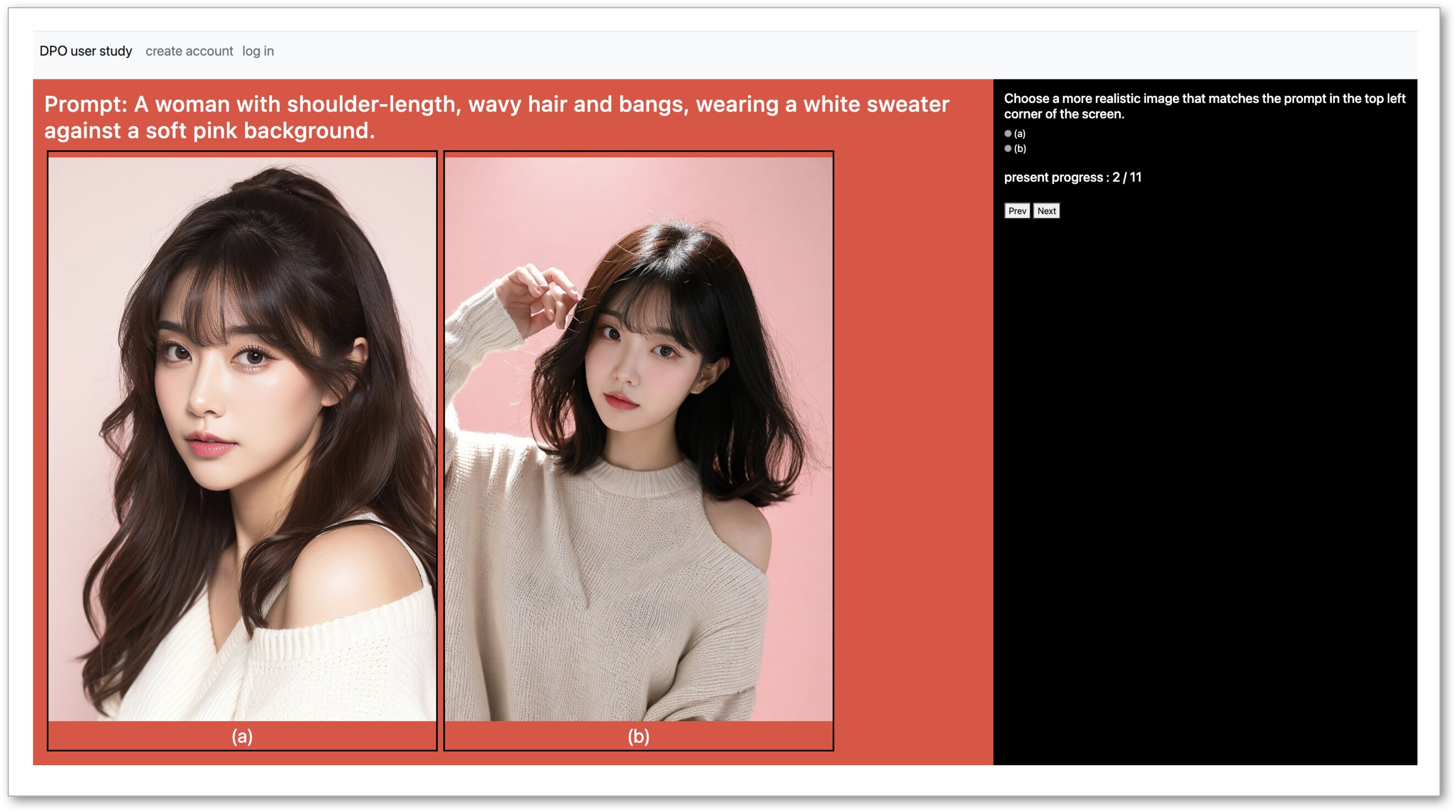}
    \caption{\textbf{User study interface.} We conduct the user study by providing a prompt and two images,
asking users to choose the one that appeared more realistic considering the given prompt.}
    \label{fig:sm_user_study_screenshot}
\end{figure}

\subsection{Adaptation to Personalized T2I model}
To adapt HG-DPO to the personalized T2I model, we firstly trained InstantBooth~\cite{shi2023instantbooth} using $\epsilon_{base}$ as the backbone. After training InstantBooth, we can seamlessly adapt the pre-trained HG-DPO LoRA layers to InstantBooth because they share the same backbone, $\epsilon_{base}$. 

\subsection{Details on Image Sampling} 
\paragraph{Sampling method.}
\textit{DPMSolverMultistepScheduler}~\cite{lu2022dpm} in \texttt{diffusers}~\cite{diffusers} is used with the step size of $50$ for sampling the images, using classifier-free guidance~\cite{ho2022classifier} with the guidance scale of $5.0$. 
\paragraph{LoRA configuration.}
In addition, the LoRA weight of 0.5 is applied to both the U-Net and the text encoder. The LoRA layers in the text encoder are specifically trained to improve image-text alignment rather than visual quality, so they are applied only to a subset of inference timesteps near the noise. Specifically, the text encoder's LoRA layers are activated during inference timesteps 900 to 1000. Additionally, as $\epsilon_\mathbb{H}$ focuses on enhancing visual fine details, its LoRA layers are applied solely to the upsampling blocks of the U-Net, while the remaining U-Net blocks are frozen. This approach is chosen because qualitative analysis suggested that applying $\epsilon_\mathbb{H}$'s LoRA layers to all U-Net blocks reduces image diversity. This method allows for improved image quality while preserving diversity as much as possible.

\subsection{Details on User Study} 
\label{subsec:user_study}
In Figures~\ref{fig:sm_user_study} and~\ref{fig:sm_user_study_normal_vs_hard}, we present the results of user studies. Each participant was tasked with selecting one of two images that best aligned with the given prompt and appeared more realistic. Here, these two images are generated by the models being compared. Evaluations were conducted using a web-based user interface, as illustrated in Figure~\ref{fig:sm_user_study_screenshot}.

\section{Broader Impacts}
\label{section:broader_impacts}
We recognize the potential negative societal impacts of our work. Since our method can generate high-quality human images, it could be misused to create malicious fake images, especially when combined with personalized T2I models. It can cause significant harm to specific individuals. However, our work can also have positive impacts on society when used beneficially, such as in the entertainment or film industries. For instance, users can create desired high-quality profile pictures using text input. It highlights the beneficial uses of our work.

\clearpage
{
    \small
    \bibliographystyle{ieeenat_fullname}
    \bibliography{main}

\begin{thebibliography}{93}
\providecommand{\natexlab}[1]{#1}
\providecommand{\url}[1]{\texttt{#1}}
\expandafter\ifx\csname urlstyle\endcsname\relax
  \providecommand{\doi}[1]{doi: #1}\else
  \providecommand{\doi}{doi: \begingroup \urlstyle{rm}\Url}\fi

\bibitem[maj()]{majicmix}
majicmix realistic.
\newblock \url{https://civitai.com/models/43331/majicmix-realistic}.

\bibitem[Bao et~al.(2020)Bao, He, Wang, Wu, Wang, Wu, Guo, Liu, and Xu]{bao2020plato}
Siqi Bao, Huang He, Fan Wang, Hua Wu, Haifeng Wang, Wenquan Wu, Zhen Guo, Zhibin Liu, and Xinchao Xu.
\newblock Plato-2: Towards building an open-domain chatbot via curriculum learning.
\newblock \emph{arXiv preprint arXiv:2006.16779}, 2020.

\bibitem[Bengio et~al.(2009)Bengio, Louradour, Collobert, and Weston]{bengio2009curriculum}
Yoshua Bengio, J{\'e}r{\^o}me Louradour, Ronan Collobert, and Jason Weston.
\newblock Curriculum learning.
\newblock In \emph{Proceedings of the 26th annual international conference on machine learning}, pages 41--48, 2009.

\bibitem[B{\"u}y{\"u}kta{\c{s}} et~al.(2021)B{\"u}y{\"u}kta{\c{s}}, Erdem, and Erdem]{buyuktacs2021curriculum}
Bar{\i}{\c{s}} B{\"u}y{\"u}kta{\c{s}}, {\c{C}}i{\u{g}}dem~Ero{\u{g}}lu Erdem, and Tanju Erdem.
\newblock Curriculum learning for face recognition.
\newblock In \emph{2020 28th European Signal Processing Conference (EUSIPCO)}, pages 650--654. IEEE, 2021.

\bibitem[Cao et~al.(2018)Cao, Shen, Xie, Parkhi, and Zisserman]{cao2018vggface2}
Qiong Cao, Li Shen, Weidi Xie, Omkar~M Parkhi, and Andrew Zisserman.
\newblock Vggface2: A dataset for recognising faces across pose and age.
\newblock In \emph{2018 13th IEEE international conference on automatic face \& gesture recognition (FG 2018)}, pages 67--74. IEEE, 2018.

\bibitem[Chen et~al.(2023{\natexlab{a}})Chen, Wang, Wu, Liao, Sun, Yan, and Lin]{chen2023enhancing}
Chaofeng Chen, Annan Wang, Haoning Wu, Liang Liao, Wenxiu Sun, Qiong Yan, and Weisi Lin.
\newblock Enhancing diffusion models with text-encoder reinforcement learning.
\newblock \emph{arXiv preprint arXiv:2311.15657}, 2023{\natexlab{a}}.

\bibitem[Chen et~al.(2023{\natexlab{b}})Chen, Zhang, Wu, Wang, Duan, Zhou, and Zhu]{chen2023disenbooth}
Hong Chen, Yipeng Zhang, Simin Wu, Xin Wang, Xuguang Duan, Yuwei Zhou, and Wenwu Zhu.
\newblock Disenbooth: Identity-preserving disentangled tuning for subject-driven text-to-image generation.
\newblock In \emph{The Twelfth International Conference on Learning Representations}, 2023{\natexlab{b}}.

\bibitem[Chen et~al.(2023{\natexlab{c}})Chen, Huang, Liu, Shen, Zhao, and Zhao]{chen2023anydoor}
Xi Chen, Lianghua Huang, Yu Liu, Yujun Shen, Deli Zhao, and Hengshuang Zhao.
\newblock Anydoor: Zero-shot object-level image customization.
\newblock \emph{arXiv preprint arXiv:2307.09481}, 2023{\natexlab{c}}.

\bibitem[Chen et~al.(2024)Chen, Deng, Yuan, Ji, and Gu]{chen2024self}
Zixiang Chen, Yihe Deng, Huizhuo Yuan, Kaixuan Ji, and Quanquan Gu.
\newblock Self-play fine-tuning converts weak language models to strong language models.
\newblock \emph{arXiv preprint arXiv:2401.01335}, 2024.

\bibitem[Cheng et~al.(2023)Cheng, Yang, Li, Dai, and Du]{cheng2023adversarial}
Pengyu Cheng, Yifan Yang, Jian Li, Yong Dai, and Nan Du.
\newblock Adversarial preference optimization.
\newblock \emph{arXiv preprint arXiv:2311.08045}, 2023.

\bibitem[Clark et~al.(2023)Clark, Vicol, Swersky, and Fleet]{clark2023directly}
Kevin Clark, Paul Vicol, Kevin Swersky, and David~J Fleet.
\newblock Directly fine-tuning diffusion models on differentiable rewards.
\newblock \emph{arXiv preprint arXiv:2309.17400}, 2023.

\bibitem[Croitoru et~al.(2024)Croitoru, Hondru, Ionescu, Sebe, and Shah]{croitoru2024curriculum}
Florinel-Alin Croitoru, Vlad Hondru, Radu~Tudor Ionescu, Nicu Sebe, and Mubarak Shah.
\newblock Curriculum direct preference optimization for diffusion and consistency models.
\newblock \emph{arXiv preprint arXiv:2405.13637}, 2024.

\bibitem[Dai et~al.(2023)Dai, Pan, Sun, Ji, Xu, Liu, Wang, and Yang]{dai2023safe}
Josef Dai, Xuehai Pan, Ruiyang Sun, Jiaming Ji, Xinbo Xu, Mickel Liu, Yizhou Wang, and Yaodong Yang.
\newblock Safe rlhf: Safe reinforcement learning from human feedback.
\newblock \emph{arXiv preprint arXiv:2310.12773}, 2023.

\bibitem[Deng et~al.(2019)Deng, Guo, Xue, and Zafeiriou]{deng2019arcface}
Jiankang Deng, Jia Guo, Niannan Xue, and Stefanos Zafeiriou.
\newblock Arcface: Additive angular margin loss for deep face recognition.
\newblock In \emph{Proceedings of the IEEE/CVF conference on computer vision and pattern recognition}, pages 4690--4699, 2019.

\bibitem[Dettmers et~al.(2021)Dettmers, Lewis, Shleifer, and Zettlemoyer]{dettmers20218}
Tim Dettmers, Mike Lewis, Sam Shleifer, and Luke Zettlemoyer.
\newblock 8-bit optimizers via block-wise quantization.
\newblock \emph{arXiv preprint arXiv:2110.02861}, 2021.

\bibitem[Dhariwal and Nichol(2021)]{dhariwal2021diffusion}
Prafulla Dhariwal and Alexander Nichol.
\newblock Diffusion models beat gans on image synthesis.
\newblock \emph{Advances in neural information processing systems}, 34:\penalty0 8780--8794, 2021.

\bibitem[Ethayarajh et~al.(2024)Ethayarajh, Xu, Muennighoff, Jurafsky, and Kiela]{ethayarajh2024kto}
Kawin Ethayarajh, Winnie Xu, Niklas Muennighoff, Dan Jurafsky, and Douwe Kiela.
\newblock Kto: Model alignment as prospect theoretic optimization.
\newblock \emph{arXiv preprint arXiv:2402.01306}, 2024.

\bibitem[Fan et~al.(2023)Fan, Watkins, Du, Liu, Ryu, Boutilier, Abbeel, Ghavamzadeh, Lee, and Lee]{fan2023dpok}
Ying Fan, Olivia Watkins, Yuqing Du, Hao Liu, Moonkyung Ryu, Craig Boutilier, Pieter Abbeel, Mohammad Ghavamzadeh, Kangwook Lee, and Kimin Lee.
\newblock Dpok: Reinforcement learning for fine-tuning text-to-image diffusion models.
\newblock \emph{Advances in Neural Information Processing Systems}, 36:\penalty0 79858--79885, 2023.

\bibitem[Fang et~al.(2019)Fang, Zhou, Du, Han, and Zhang]{fang2019curriculum}
Meng Fang, Tianyi Zhou, Yali Du, Lei Han, and Zhengyou Zhang.
\newblock Curriculum-guided hindsight experience replay.
\newblock \emph{Advances in neural information processing systems}, 32, 2019.

\bibitem[Florensa et~al.(2017)Florensa, Held, Wulfmeier, Zhang, and Abbeel]{florensa2017reverse}
Carlos Florensa, David Held, Markus Wulfmeier, Michael Zhang, and Pieter Abbeel.
\newblock Reverse curriculum generation for reinforcement learning.
\newblock In \emph{Conference on robot learning}, pages 482--495. PMLR, 2017.

\bibitem[Gambashidze et~al.(2024)Gambashidze, Kulikov, Sosnin, and Makarov]{gambashidze2024aligning}
Alexander Gambashidze, Anton Kulikov, Yuriy Sosnin, and Ilya Makarov.
\newblock Aligning diffusion models with noise-conditioned perception.
\newblock \emph{arXiv preprint arXiv:2406.17636}, 2024.

\bibitem[Goodfellow et~al.(2014)Goodfellow, Pouget-Abadie, Mirza, Xu, Warde-Farley, Ozair, Courville, and Bengio]{goodfellow2014generative}
Ian Goodfellow, Jean Pouget-Abadie, Mehdi Mirza, Bing Xu, David Warde-Farley, Sherjil Ozair, Aaron Courville, and Yoshua Bengio.
\newblock Generative adversarial nets.
\newblock \emph{Advances in neural information processing systems}, 27, 2014.

\bibitem[Gu et~al.(2022)Gu, Chen, Bao, Wen, Zhang, Chen, Yuan, and Guo]{gu2022vector}
Shuyang Gu, Dong Chen, Jianmin Bao, Fang Wen, Bo Zhang, Dongdong Chen, Lu Yuan, and Baining Guo.
\newblock Vector quantized diffusion model for text-to-image synthesis.
\newblock In \emph{Proceedings of the IEEE/CVF Conference on Computer Vision and Pattern Recognition}, pages 10696--10706, 2022.

\bibitem[Gu et~al.(2024)Gu, Wang, Yin, Xie, and Zhou]{gu2024diffusion}
Yi Gu, Zhendong Wang, Yueqin Yin, Yujia Xie, and Mingyuan Zhou.
\newblock Diffusion-rpo: Aligning diffusion models through relative preference optimization.
\newblock \emph{arXiv preprint arXiv:2406.06382}, 2024.

\bibitem[Heusel et~al.(2017)Heusel, Ramsauer, Unterthiner, Nessler, and Hochreiter]{heusel2017gans}
Martin Heusel, Hubert Ramsauer, Thomas Unterthiner, Bernhard Nessler, and Sepp Hochreiter.
\newblock Gans trained by a two time-scale update rule converge to a local nash equilibrium.
\newblock \emph{Advances in neural information processing systems}, 30, 2017.

\bibitem[Ho and Salimans(2022)]{ho2022classifier}
Jonathan Ho and Tim Salimans.
\newblock Classifier-free diffusion guidance.
\newblock \emph{arXiv preprint arXiv:2207.12598}, 2022.

\bibitem[Ho et~al.(2020)Ho, Jain, and Abbeel]{ho2020denoising}
Jonathan Ho, Ajay Jain, and Pieter Abbeel.
\newblock Denoising diffusion probabilistic models.
\newblock \emph{Advances in neural information processing systems}, 33:\penalty0 6840--6851, 2020.

\bibitem[Hong et~al.(2024)Hong, Paul, Lee, Rasul, Thorne, and Jeong]{hong2024margin}
Jiwoo Hong, Sayak Paul, Noah Lee, Kashif Rasul, James Thorne, and Jongheon Jeong.
\newblock Margin-aware preference optimization for aligning diffusion models without reference.
\newblock \emph{arXiv preprint arXiv:2406.06424}, 2024.

\bibitem[Hu et~al.(2021)Hu, Shen, Wallis, Allen-Zhu, Li, Wang, Wang, and Chen]{hu2021lora}
Edward~J Hu, Yelong Shen, Phillip Wallis, Zeyuan Allen-Zhu, Yuanzhi Li, Shean Wang, Lu Wang, and Weizhu Chen.
\newblock Lora: Low-rank adaptation of large language models.
\newblock \emph{arXiv preprint arXiv:2106.09685}, 2021.

\bibitem[Karras et~al.(2019)Karras, Laine, and Aila]{karras2019style}
Tero Karras, Samuli Laine, and Timo Aila.
\newblock A style-based generator architecture for generative adversarial networks.
\newblock In \emph{Proceedings of the IEEE/CVF conference on computer vision and pattern recognition}, pages 4401--4410, 2019.

\bibitem[Karras et~al.(2020)Karras, Laine, Aittala, Hellsten, Lehtinen, and Aila]{karras2020analyzing}
Tero Karras, Samuli Laine, Miika Aittala, Janne Hellsten, Jaakko Lehtinen, and Timo Aila.
\newblock Analyzing and improving the image quality of stylegan.
\newblock In \emph{Proceedings of the IEEE/CVF conference on computer vision and pattern recognition}, pages 8110--8119, 2020.

\bibitem[Karras et~al.(2021)Karras, Aittala, Laine, H{\"a}rk{\"o}nen, Hellsten, Lehtinen, and Aila]{karras2021alias}
Tero Karras, Miika Aittala, Samuli Laine, Erik H{\"a}rk{\"o}nen, Janne Hellsten, Jaakko Lehtinen, and Timo Aila.
\newblock Alias-free generative adversarial networks.
\newblock \emph{Advances in neural information processing systems}, 34:\penalty0 852--863, 2021.

\bibitem[Kingma(2013)]{kingma2013auto}
Diederik~P Kingma.
\newblock Auto-encoding variational bayes.
\newblock \emph{arXiv preprint arXiv:1312.6114}, 2013.

\bibitem[Kirstain et~al.(2024)Kirstain, Polyak, Singer, Matiana, Penna, and Levy]{kirstain2024pick}
Yuval Kirstain, Adam Polyak, Uriel Singer, Shahbuland Matiana, Joe Penna, and Omer Levy.
\newblock Pick-a-pic: An open dataset of user preferences for text-to-image generation.
\newblock \emph{Advances in Neural Information Processing Systems}, 36, 2024.

\bibitem[Kocmi and Bojar(2017)]{kocmi2017curriculum}
Tom Kocmi and Ondrej Bojar.
\newblock Curriculum learning and minibatch bucketing in neural machine translation.
\newblock \emph{arXiv preprint arXiv:1707.09533}, 2017.

\bibitem[Korbak et~al.(2023)Korbak, Shi, Chen, Bhalerao, Buckley, Phang, Bowman, and Perez]{korbak2023pretraining}
Tomasz Korbak, Kejian Shi, Angelica Chen, Rasika~Vinayak Bhalerao, Christopher Buckley, Jason Phang, Samuel~R Bowman, and Ethan Perez.
\newblock Pretraining language models with human preferences.
\newblock In \emph{International Conference on Machine Learning}, pages 17506--17533. PMLR, 2023.

\bibitem[Kumar et~al.(2011)Kumar, Turki, Preston, and Koller]{kumar2011learning}
M~Pawan Kumar, Haithem Turki, Dan Preston, and Daphne Koller.
\newblock Learning specific-class segmentation from diverse data.
\newblock In \emph{2011 International conference on computer vision}, pages 1800--1807. IEEE, 2011.

\bibitem[Li et~al.(2024)Li, Liu, Kag, Hu, Idelbayev, Sagar, Wang, Tulyakov, and Ren]{li2024textcraftor}
Yanyu Li, Xian Liu, Anil Kag, Ju Hu, Yerlan Idelbayev, Dhritiman Sagar, Yanzhi Wang, Sergey Tulyakov, and Jian Ren.
\newblock Textcraftor: Your text encoder can be image quality controller.
\newblock In \emph{Proceedings of the IEEE/CVF Conference on Computer Vision and Pattern Recognition}, pages 7985--7995, 2024.

\bibitem[Liang et~al.(2024)Liang, Yuan, Gu, Chen, Hang, Li, and Zheng]{liang2024step}
Zhanhao Liang, Yuhui Yuan, Shuyang Gu, Bohan Chen, Tiankai Hang, Ji Li, and Liang Zheng.
\newblock Step-aware preference optimization: Aligning preference with denoising performance at each step.
\newblock \emph{arXiv preprint arXiv:2406.04314}, 2024.

\bibitem[Liu et~al.(2018)Liu, He, Liu, Zhao, et~al.]{liu2018curriculum}
Cao Liu, Shizhu He, Kang Liu, Jun Zhao, et~al.
\newblock Curriculum learning for natural answer generation.
\newblock In \emph{IJCAI}, pages 4223--4229, 2018.

\bibitem[Liu et~al.(2023{\natexlab{a}})Liu, Li, Wu, and Lee]{liu2023llava}
Haotian Liu, Chunyuan Li, Qingyang Wu, and Yong~Jae Lee.
\newblock Visual instruction tuning.
\newblock In \emph{NeurIPS}, 2023{\natexlab{a}}.

\bibitem[Liu et~al.(2023{\natexlab{b}})Liu, Zhao, Joshi, Khalman, Saleh, Liu, and Liu]{liu2023statistical}
Tianqi Liu, Yao Zhao, Rishabh Joshi, Misha Khalman, Mohammad Saleh, Peter~J Liu, and Jialu Liu.
\newblock Statistical rejection sampling improves preference optimization.
\newblock \emph{arXiv preprint arXiv:2309.06657}, 2023{\natexlab{b}}.

\bibitem[Liu et~al.(2024)Liu, Qin, Wu, Shen, Khalman, Joshi, Zhao, Saleh, Baumgartner, Liu, et~al.]{liu2024lipo}
Tianqi Liu, Zhen Qin, Junru Wu, Jiaming Shen, Misha Khalman, Rishabh Joshi, Yao Zhao, Mohammad Saleh, Simon Baumgartner, Jialu Liu, et~al.
\newblock Lipo: Listwise preference optimization through learning-to-rank.
\newblock \emph{arXiv preprint arXiv:2402.01878}, 2024.

\bibitem[Liu et~al.(2023{\natexlab{c}})Liu, Wang, Wu, Li, Lv, Ling, Zhu, Zhang, Zheng, and Huang]{liu2023aligning}
Wenhao Liu, Xiaohua Wang, Muling Wu, Tianlong Li, Changze Lv, Zixuan Ling, Jianhao Zhu, Cenyuan Zhang, Xiaoqing Zheng, and Xuanjing Huang.
\newblock Aligning large language models with human preferences through representation engineering.
\newblock \emph{arXiv preprint arXiv:2312.15997}, 2023{\natexlab{c}}.

\bibitem[Liu et~al.(2020)Liu, Lai, Wong, and Chao]{liu2020norm}
Xuebo Liu, Houtim Lai, Derek~F Wong, and Lidia~S Chao.
\newblock Norm-based curriculum learning for neural machine translation.
\newblock \emph{arXiv preprint arXiv:2006.02014}, 2020.

\bibitem[Lu et~al.(2022)Lu, Zhou, Bao, Chen, Li, and Zhu]{lu2022dpm}
Cheng Lu, Yuhao Zhou, Fan Bao, Jianfei Chen, Chongxuan Li, and Jun Zhu.
\newblock Dpm-solver: A fast ode solver for diffusion probabilistic model sampling in around 10 steps.
\newblock \emph{arXiv preprint arXiv:2206.00927}, 2022.

\bibitem[Luo et~al.(2020)Luo, Kasaei, and Schomaker]{luo2020accelerating}
Sha Luo, Hamidreza Kasaei, and Lambert Schomaker.
\newblock Accelerating reinforcement learning for reaching using continuous curriculum learning.
\newblock In \emph{2020 International Joint Conference on Neural Networks (IJCNN)}, pages 1--8. IEEE, 2020.

\bibitem[Ma et~al.(2023)Ma, Liang, Chen, and Lu]{ma2023subject}
Jian Ma, Junhao Liang, Chen Chen, and Haonan Lu.
\newblock Subject-diffusion: Open domain personalized text-to-image generation without test-time fine-tuning.
\newblock \emph{arXiv preprint arXiv:2307.11410}, 2023.

\bibitem[Manela and Biess(2022)]{manela2022curriculum}
Binyamin Manela and Armin Biess.
\newblock Curriculum learning with hindsight experience replay for sequential object manipulation tasks.
\newblock \emph{Neural Networks}, 145:\penalty0 260--270, 2022.

\bibitem[Meng et~al.(2021)Meng, He, Song, Song, Wu, Zhu, and Ermon]{meng2021sdedit}
Chenlin Meng, Yutong He, Yang Song, Jiaming Song, Jiajun Wu, Jun-Yan Zhu, and Stefano Ermon.
\newblock Sdedit: Guided image synthesis and editing with stochastic differential equations.
\newblock \emph{arXiv preprint arXiv:2108.01073}, 2021.

\bibitem[Milano and Nolfi(2021)]{milano2021automated}
Nicola Milano and Stefano Nolfi.
\newblock Automated curriculum learning for embodied agents a neuroevolutionary approach.
\newblock \emph{Scientific reports}, 11\penalty0 (1):\penalty0 8985, 2021.

\bibitem[Murali et~al.(2018)Murali, Pinto, Gandhi, and Gupta]{murali2018cassl}
Adithyavairavan Murali, Lerrel Pinto, Dhiraj Gandhi, and Abhinav Gupta.
\newblock Cassl: Curriculum accelerated self-supervised learning.
\newblock In \emph{2018 IEEE International Conference on Robotics and Automation (ICRA)}, pages 6453--6460. IEEE, 2018.

\bibitem[Na(2022)]{na2022mfim}
Sanghyeon Na.
\newblock Mfim: Megapixel facial identity manipulation.
\newblock In \emph{European Conference on Computer Vision}, pages 143--159. Springer, 2022.

\bibitem[Narvekar et~al.(2016)Narvekar, Sinapov, Leonetti, and Stone]{narvekar2016source}
Sanmit Narvekar, Jivko Sinapov, Matteo Leonetti, and Peter Stone.
\newblock Source task creation for curriculum learning.
\newblock In \emph{Proceedings of the 2016 international conference on autonomous agents \& multiagent systems}, pages 566--574, 2016.

\bibitem[Nichol et~al.(2021)Nichol, Dhariwal, Ramesh, Shyam, Mishkin, McGrew, Sutskever, and Chen]{nichol2021glide}
Alex Nichol, Prafulla Dhariwal, Aditya Ramesh, Pranav Shyam, Pamela Mishkin, Bob McGrew, Ilya Sutskever, and Mark Chen.
\newblock Glide: Towards photorealistic image generation and editing with text-guided diffusion models.
\newblock \emph{arXiv preprint arXiv:2112.10741}, 2021.

\bibitem[Ouyang et~al.(2022)Ouyang, Wu, Jiang, Almeida, Wainwright, Mishkin, Zhang, Agarwal, Slama, Ray, et~al.]{ouyang2022training}
Long Ouyang, Jeffrey Wu, Xu Jiang, Diogo Almeida, Carroll Wainwright, Pamela Mishkin, Chong Zhang, Sandhini Agarwal, Katarina Slama, Alex Ray, et~al.
\newblock Training language models to follow instructions with human feedback.
\newblock \emph{Advances in neural information processing systems}, 35:\penalty0 27730--27744, 2022.

\bibitem[Peng(Forthcoming)]{peng2021athec}
Yilang Peng.
\newblock Athec: A python library for computational aesthetic analysis of visual media in social science research.
\newblock \emph{Computational Communication Research}, Forthcoming.

\bibitem[Podell et~al.(2023)Podell, English, Lacey, Blattmann, Dockhorn, M{\"u}ller, Penna, and Rombach]{podell2023sdxl}
Dustin Podell, Zion English, Kyle Lacey, Andreas Blattmann, Tim Dockhorn, Jonas M{\"u}ller, Joe Penna, and Robin Rombach.
\newblock Sdxl: Improving latent diffusion models for high-resolution image synthesis.
\newblock \emph{arXiv preprint arXiv:2307.01952}, 2023.

\bibitem[Prabhudesai et~al.(2023)Prabhudesai, Goyal, Pathak, and Fragkiadaki]{prabhudesai2023aligning}
Mihir Prabhudesai, Anirudh Goyal, Deepak Pathak, and Katerina Fragkiadaki.
\newblock Aligning text-to-image diffusion models with reward backpropagation.
\newblock \emph{arXiv preprint arXiv:2310.03739}, 2023.

\bibitem[Qin et~al.(2020)Qin, Hu, Liu, Fu, He, and Hong]{qin2020balanced}
Wei Qin, Zhenzhen Hu, Xueliang Liu, Weijie Fu, Jun He, and Richang Hong.
\newblock The balanced loss curriculum learning.
\newblock \emph{IEEE Access}, 8:\penalty0 25990--26001, 2020.

\bibitem[Radford et~al.(2021)Radford, Kim, Hallacy, Ramesh, Goh, Agarwal, Sastry, Askell, Mishkin, Clark, et~al.]{radford2021learning}
Alec Radford, Jong~Wook Kim, Chris Hallacy, Aditya Ramesh, Gabriel Goh, Sandhini Agarwal, Girish Sastry, Amanda Askell, Pamela Mishkin, Jack Clark, et~al.
\newblock Learning transferable visual models from natural language supervision.
\newblock In \emph{International conference on machine learning}, pages 8748--8763. PMLR, 2021.

\bibitem[Rafailov et~al.(2024)Rafailov, Sharma, Mitchell, Manning, Ermon, and Finn]{rafailov2024direct}
Rafael Rafailov, Archit Sharma, Eric Mitchell, Christopher~D Manning, Stefano Ermon, and Chelsea Finn.
\newblock Direct preference optimization: Your language model is secretly a reward model.
\newblock \emph{Advances in Neural Information Processing Systems}, 36, 2024.

\bibitem[Ramesh et~al.(2022)Ramesh, Dhariwal, Nichol, Chu, and Chen]{ramesh2022hierarchical}
Aditya Ramesh, Prafulla Dhariwal, Alex Nichol, Casey Chu, and Mark Chen.
\newblock Hierarchical text-conditional image generation with clip latents.
\newblock \emph{arXiv preprint arXiv:2204.06125}, 1\penalty0 (2):\penalty0 3, 2022.

\bibitem[Rombach et~al.(2022)Rombach, Blattmann, Lorenz, Esser, and Ommer]{rombach2022high}
Robin Rombach, Andreas Blattmann, Dominik Lorenz, Patrick Esser, and Bj{\"o}rn Ommer.
\newblock High-resolution image synthesis with latent diffusion models.
\newblock In \emph{Proceedings of the IEEE/CVF conference on computer vision and pattern recognition}, pages 10684--10695, 2022.

\bibitem[Ruiz et~al.(2023)Ruiz, Li, Jampani, Pritch, Rubinstein, and Aberman]{ruiz2023dreambooth}
Nataniel Ruiz, Yuanzhen Li, Varun Jampani, Yael Pritch, Michael Rubinstein, and Kfir Aberman.
\newblock Dreambooth: Fine tuning text-to-image diffusion models for subject-driven generation.
\newblock In \emph{Proceedings of the IEEE/CVF Conference on Computer Vision and Pattern Recognition}, pages 22500--22510, 2023.

\bibitem[Sachan and Xing(2016)]{sachan2016easy}
Mrinmaya Sachan and Eric Xing.
\newblock Easy questions first? a case study on curriculum learning for question answering.
\newblock In \emph{Proceedings of the 54th Annual Meeting of the Association for Computational Linguistics (Volume 1: Long Papers)}, pages 453--463, 2016.

\bibitem[Saharia et~al.(2022)Saharia, Chan, Saxena, Li, Whang, Denton, Ghasemipour, Gontijo~Lopes, Karagol~Ayan, Salimans, et~al.]{saharia2022photorealistic}
Chitwan Saharia, William Chan, Saurabh Saxena, Lala Li, Jay Whang, Emily~L Denton, Kamyar Ghasemipour, Raphael Gontijo~Lopes, Burcu Karagol~Ayan, Tim Salimans, et~al.
\newblock Photorealistic text-to-image diffusion models with deep language understanding.
\newblock \emph{Advances in neural information processing systems}, 35:\penalty0 36479--36494, 2022.

\bibitem[Schuhmann(2022)]{laionaes}
Christoph Schuhmann.
\newblock Laion-aesthetics.
\newblock \url{https://laion.ai/blog/laion-aesthetics/}, 2022.

\bibitem[Shi et~al.(2023)Shi, Xiong, Lin, and Jung]{shi2023instantbooth}
Jing Shi, Wei Xiong, Zhe Lin, and Hyun~Joon Jung.
\newblock Instantbooth: Personalized text-to-image generation without test-time finetuning.
\newblock \emph{arXiv preprint arXiv:2304.03411}, 2023.

\bibitem[Shi and Ferrari(2016)]{shi2016weakly}
Miaojing Shi and Vittorio Ferrari.
\newblock Weakly supervised object localization using size estimates.
\newblock In \emph{Computer Vision--ECCV 2016: 14th European Conference, Amsterdam, The Netherlands, October 11-14, 2016, Proceedings, Part V 14}, pages 105--121. Springer, 2016.

\bibitem[Song et~al.(2024)Song, Yu, Li, Yu, Huang, Li, and Wang]{song2024preference}
Feifan Song, Bowen Yu, Minghao Li, Haiyang Yu, Fei Huang, Yongbin Li, and Houfeng Wang.
\newblock Preference ranking optimization for human alignment.
\newblock In \emph{Proceedings of the AAAI Conference on Artificial Intelligence}, pages 18990--18998, 2024.

\bibitem[Song et~al.(2020{\natexlab{a}})Song, Meng, and Ermon]{song2020denoising}
Jiaming Song, Chenlin Meng, and Stefano Ermon.
\newblock Denoising diffusion implicit models.
\newblock \emph{arXiv preprint arXiv:2010.02502}, 2020{\natexlab{a}}.

\bibitem[Song et~al.(2020{\natexlab{b}})Song, Sohl-Dickstein, Kingma, Kumar, Ermon, and Poole]{song2020score}
Yang Song, Jascha Sohl-Dickstein, Diederik~P Kingma, Abhishek Kumar, Stefano Ermon, and Ben Poole.
\newblock Score-based generative modeling through stochastic differential equations.
\newblock \emph{arXiv preprint arXiv:2011.13456}, 2020{\natexlab{b}}.

\bibitem[Soviany et~al.(2020)Soviany, Ardei, Ionescu, and Leordeanu]{soviany2020image}
Petru Soviany, Claudiu Ardei, Radu~Tudor Ionescu, and Marius Leordeanu.
\newblock Image difficulty curriculum for generative adversarial networks (cugan).
\newblock In \emph{Proceedings of the IEEE/CVF winter conference on applications of computer vision}, pages 3463--3472, 2020.

\bibitem[Soviany et~al.(2021)Soviany, Ionescu, Rota, and Sebe]{soviany2021curriculum}
Petru Soviany, Radu~Tudor Ionescu, Paolo Rota, and Nicu Sebe.
\newblock Curriculum self-paced learning for cross-domain object detection.
\newblock \emph{Computer Vision and Image Understanding}, 204:\penalty0 103166, 2021.

\bibitem[Tang et~al.(2012)Tang, Yang, and Gao]{tang2012self}
Ye Tang, Yu-Bin Yang, and Yang Gao.
\newblock Self-paced dictionary learning for image classification.
\newblock In \emph{Proceedings of the 20th ACM international conference on Multimedia}, pages 833--836, 2012.

\bibitem[von Platen et~al.(2022)von Platen, Patil, Lozhkov, Cuenca, Lambert, Rasul, Davaadorj, Nair, Paul, Berman, Xu, Liu, and Wolf]{diffusers}
Patrick von Platen, Suraj Patil, Anton Lozhkov, Pedro Cuenca, Nathan Lambert, Kashif Rasul, Mishig Davaadorj, Dhruv Nair, Sayak Paul, William Berman, Yiyi Xu, Steven Liu, and Thomas Wolf.
\newblock Diffusers: State-of-the-art diffusion models.
\newblock \url{https://github.com/huggingface/diffusers}, 2022.

\bibitem[Wallace et~al.(2023)Wallace, Dang, Rafailov, Zhou, Lou, Purushwalkam, Ermon, Xiong, Joty, and Naik]{wallace2023diffusion}
Bram Wallace, Meihua Dang, Rafael Rafailov, Linqi Zhou, Aaron Lou, Senthil Purushwalkam, Stefano Ermon, Caiming Xiong, Shafiq Joty, and Nikhil Naik.
\newblock Diffusion model alignment using direct preference optimization.
\newblock \emph{arXiv preprint arXiv:2311.12908}, 2023.

\bibitem[Wang et~al.(2023)Wang, Chan, and Loy]{wang2022exploring}
Jianyi Wang, Kelvin~CK Chan, and Chen~Change Loy.
\newblock Exploring clip for assessing the look and feel of images.
\newblock In \emph{AAAI}, 2023.

\bibitem[Wang et~al.(2024{\natexlab{a}})Wang, Bai, Tan, Wang, Fan, Bai, Chen, Liu, Wang, Ge, Fan, Dang, Du, Ren, Men, Liu, Zhou, Zhou, and Lin]{Qwen2VL}
Peng Wang, Shuai Bai, Sinan Tan, Shijie Wang, Zhihao Fan, Jinze Bai, Keqin Chen, Xuejing Liu, Jialin Wang, Wenbin Ge, Yang Fan, Kai Dang, Mengfei Du, Xuancheng Ren, Rui Men, Dayiheng Liu, Chang Zhou, Jingren Zhou, and Junyang Lin.
\newblock Qwen2-vl: Enhancing vision-language model's perception of the world at any resolution.
\newblock \emph{arXiv preprint arXiv:2409.12191}, 2024{\natexlab{a}}.

\bibitem[Wang et~al.(2024{\natexlab{b}})Wang, Bai, Tan, Wang, Fan, Bai, Chen, Liu, Wang, Ge, et~al.]{wang2024qwen2}
Peng Wang, Shuai Bai, Sinan Tan, Shijie Wang, Zhihao Fan, Jinze Bai, Keqin Chen, Xuejing Liu, Jialin Wang, Wenbin Ge, et~al.
\newblock Qwen2-vl: Enhancing vision-language model's perception of the world at any resolution.
\newblock \emph{arXiv preprint arXiv:2409.12191}, 2024{\natexlab{b}}.

\bibitem[Wu et~al.(2023)Wu, Hao, Sun, Chen, Zhu, Zhao, and Li]{wu2023human}
Xiaoshi Wu, Yiming Hao, Keqiang Sun, Yixiong Chen, Feng Zhu, Rui Zhao, and Hongsheng Li.
\newblock Human preference score v2: A solid benchmark for evaluating human preferences of text-to-image synthesis.
\newblock \emph{arXiv preprint arXiv:2306.09341}, 2023.

\bibitem[Wu et~al.(2024)Wu, Hu, Shi, Dziri, Suhr, Ammanabrolu, Smith, Ostendorf, and Hajishirzi]{wu2024fine}
Zeqiu Wu, Yushi Hu, Weijia Shi, Nouha Dziri, Alane Suhr, Prithviraj Ammanabrolu, Noah~A Smith, Mari Ostendorf, and Hannaneh Hajishirzi.
\newblock Fine-grained human feedback gives better rewards for language model training.
\newblock \emph{Advances in Neural Information Processing Systems}, 36, 2024.

\bibitem[Xiao et~al.(2023)Xiao, Yin, Freeman, Durand, and Han]{xiao2023fastcomposer}
Guangxuan Xiao, Tianwei Yin, William~T Freeman, Fr{\'e}do Durand, and Song Han.
\newblock Fastcomposer: Tuning-free multi-subject image generation with localized attention.
\newblock \emph{arXiv preprint arXiv:2305.10431}, 2023.

\bibitem[Xu et~al.(2024)Xu, Liu, Wu, Tong, Li, Ding, Tang, and Dong]{xu2024imagereward}
Jiazheng Xu, Xiao Liu, Yuchen Wu, Yuxuan Tong, Qinkai Li, Ming Ding, Jie Tang, and Yuxiao Dong.
\newblock Imagereward: Learning and evaluating human preferences for text-to-image generation.
\newblock \emph{Advances in Neural Information Processing Systems}, 36, 2024.

\bibitem[Yang et~al.(2023)Yang, Tao, Lyu, Ge, Chen, Li, Shen, Zhu, and Li]{yang2023using}
Kai Yang, Jian Tao, Jiafei Lyu, Chunjiang Ge, Jiaxin Chen, Qimai Li, Weihan Shen, Xiaolong Zhu, and Xiu Li.
\newblock Using human feedback to fine-tune diffusion models without any reward model.
\newblock \emph{arXiv preprint arXiv:2311.13231}, 2023.

\bibitem[Yang et~al.(2024)Yang, Tao, Lyu, Ge, Chen, Shen, Zhu, and Li]{yang2024using}
Kai Yang, Jian Tao, Jiafei Lyu, Chunjiang Ge, Jiaxin Chen, Weihan Shen, Xiaolong Zhu, and Xiu Li.
\newblock Using human feedback to fine-tune diffusion models without any reward model.
\newblock In \emph{Proceedings of the IEEE/CVF Conference on Computer Vision and Pattern Recognition}, pages 8941--8951, 2024.

\bibitem[Yuan et~al.(2024)Yuan, Chen, Ji, and Gu]{yuan2024self}
Huizhuo Yuan, Zixiang Chen, Kaixuan Ji, and Quanquan Gu.
\newblock Self-play fine-tuning of diffusion models for text-to-image generation.
\newblock \emph{arXiv preprint arXiv:2402.10210}, 2024.

\bibitem[Zhan et~al.(2021)Zhan, Liu, Wong, and Chao]{zhan2021meta}
Runzhe Zhan, Xuebo Liu, Derek~F Wong, and Lidia~S Chao.
\newblock Meta-curriculum learning for domain adaptation in neural machine translation.
\newblock In \emph{Proceedings of the AAAI Conference on Artificial Intelligence}, pages 14310--14318, 2021.

\bibitem[Zhang et~al.(2021)Zhang, Wang, Hou, Wu, Wang, Okumura, and Shinozaki]{zhang2021flexmatch}
Bowen Zhang, Yidong Wang, Wenxin Hou, Hao Wu, Jindong Wang, Manabu Okumura, and Takahiro Shinozaki.
\newblock Flexmatch: Boosting semi-supervised learning with curriculum pseudo labeling.
\newblock \emph{Advances in Neural Information Processing Systems}, 34:\penalty0 18408--18419, 2021.

\bibitem[Zhao et~al.(2020)Zhao, Wu, Niu, and Wang]{zhao2020reinforced}
Mingjun Zhao, Haijiang Wu, Di Niu, and Xiaoli Wang.
\newblock Reinforced curriculum learning on pre-trained neural machine translation models.
\newblock In \emph{Proceedings of the AAAI Conference on Artificial Intelligence}, pages 9652--9659, 2020.

\bibitem[Zhao et~al.(2023)Zhao, Joshi, Liu, Khalman, Saleh, and Liu]{zhao2023slic}
Yao Zhao, Rishabh Joshi, Tianqi Liu, Misha Khalman, Mohammad Saleh, and Peter~J Liu.
\newblock Slic-hf: Sequence likelihood calibration with human feedback.
\newblock \emph{arXiv preprint arXiv:2305.10425}, 2023.

\bibitem[Zhou et~al.(2020)Zhou, Yang, Wong, Wan, and Chao]{zhou2020uncertainty}
Yikai Zhou, Baosong Yang, Derek~F Wong, Yu Wan, and Lidia~S Chao.
\newblock Uncertainty-aware curriculum learning for neural machine translation.
\newblock In \emph{Proceedings of the 58th Annual Meeting of the association for computational linguistics}, pages 6934--6944, 2020.

\end{thebibliography}
}

\end{document}